\documentclass[10pt,twocolumn,letterpaper]{article}

\usepackage{cvpr}
\usepackage{times}
\usepackage{epsfig}
\usepackage{graphicx}
\usepackage{amsmath}
\usepackage{amssymb}
\usepackage{cite}

\newcommand{\PAR}[1]{\vskip4pt \noindent{\bf #1~}}

\usepackage{acronym}
\usepackage{subcaption}
\usepackage{array,tabularx,booktabs}
\usepackage[group-separator={,}]{siunitx}
\usepackage{floatrow}
\usepackage[dvipsnames]{xcolor}
\usepackage{tikz}
\usepackage[utf8]{inputenc}
\usepackage{gensymb}
\usepackage{nicefrac}
\DeclareMathOperator*{\argmax}{arg\,max}
\DeclareMathOperator*{\argmin}{arg\,min}
\usepackage[pagebackref=true,breaklinks=true,letterpaper=true,colorlinks,bookmarks=false]{hyperref}

\cvprfinalcopy % *** Uncomment this line for the final submission

\def\cvprPaperID{3623} % *** Enter the CVPR Paper ID here

\ifcvprfinal\fi
\begin{document}

%%%%%%%%% TITLE
\title{D2-Net: A Trainable CNN for \emph{Joint Description and Detection} of Local Features}

\author{Mihai Dusmanu$^{1,2,3}$\\
\and
Ignacio Rocco$^{1,2}$\\
\and
Tomas Pajdla$^{4}$\\
\and
Marc Pollefeys$^{3,5}$\\
\and
Josef Sivic$^{1,2,4}$\\
\and
Akihiko Torii$^{6}$\\
\and
Torsten Sattler$^7$\\[0.5mm]
\and
\normalfont
\normalsize\textsuperscript{$1$}DI, ENS\quad
\normalsize\textsuperscript{$2$}Inria\quad
\normalsize\textsuperscript{$3$}Department of Computer Science, ETH Zurich\quad 
\normalsize\textsuperscript{$4$}CIIRC, CTU in Prague
\quad
\and
\normalfont
\normalsize\textsuperscript{$5$}Microsoft\quad
\normalsize\textsuperscript{$6$}Tokyo Institute of Technology\quad
\normalsize\textsuperscript{$7$}Chalmers University of Technology
}
\maketitle

% Institution footnotes
\footnotetext[1]{Département d’informatique de l’ENS, École normale supérieure, CNRS, PSL Research University, 75005}
\footnotetext[4]{Czech Institute of Informatics, Robotics, and Cybernetics, Czech Technical University in Prague}

\thispagestyle{empty}

%%%%%%%%% Acronyms
\newacro{DoG}[DoG]{Difference-of-Gaussians}
\newacro{LoG}[LoG]{Laplacian-of-Gaussian}
\newacro{LIFT}[LIFT]{Learned Invariant Feature Transform}
\newacro{ReLU}[ReLU]{Rectified Linear Unit}
\newacro{SIFT}[SIFT]{Scale Invariant Feature Transform}
\newacro{SfM}[SfM]{Structure-from-Motion}
\newacro{CNN}[CNN]{Convolution Neural Network}
\newacro{RANSAC}[RANSAC]{RANdom SAmple Consensus}
\newacro{P3P}[P3P]{Perspective-Three-Point}
\newacro{DUC}[DUC]{Danforth University Center}
\newacro{CSE}[CSE]{Computer Science \& Engineering}
\newacro{PE}[PE]{Pose Estimation}
\newacro{LO}[LO]{Locally Optimized}
\newacro{PV}[PV]{Pose Verification}

%%%%%%%%% ABSTRACT
\begin{abstract}
    In this work we address the problem of finding reliable pixel-level correspondences under difficult imaging conditions. We propose an approach where a single convolutional neural network plays a dual role: It is simultaneously a dense feature descriptor and a feature detector. 
    By postponing the detection to a later stage, the obtained keypoints are more stable than their traditional counterparts based on early detection of low-level structures.
    We show that this model can be trained using pixel correspondences extracted from readily available large-scale SfM reconstructions, without any further annotations.
    The proposed method obtains state-of-the-art performance on both the difficult Aachen Day-Night localization dataset and the InLoc indoor localization benchmark, as well as competitive performance on other benchmarks for image matching and 3D reconstruction.
\end{abstract}

%%%%%%%%% BODY TEXT
\section{Introduction}
Establishing pixel-level correspondences between images is one of the fundamental computer vision problems, with applications in 3D computer vision, video compression, tracking, image retrieval, and visual localization.

\begin{figure}[t]
    \centering
    \includegraphics[width=0.8\columnwidth]{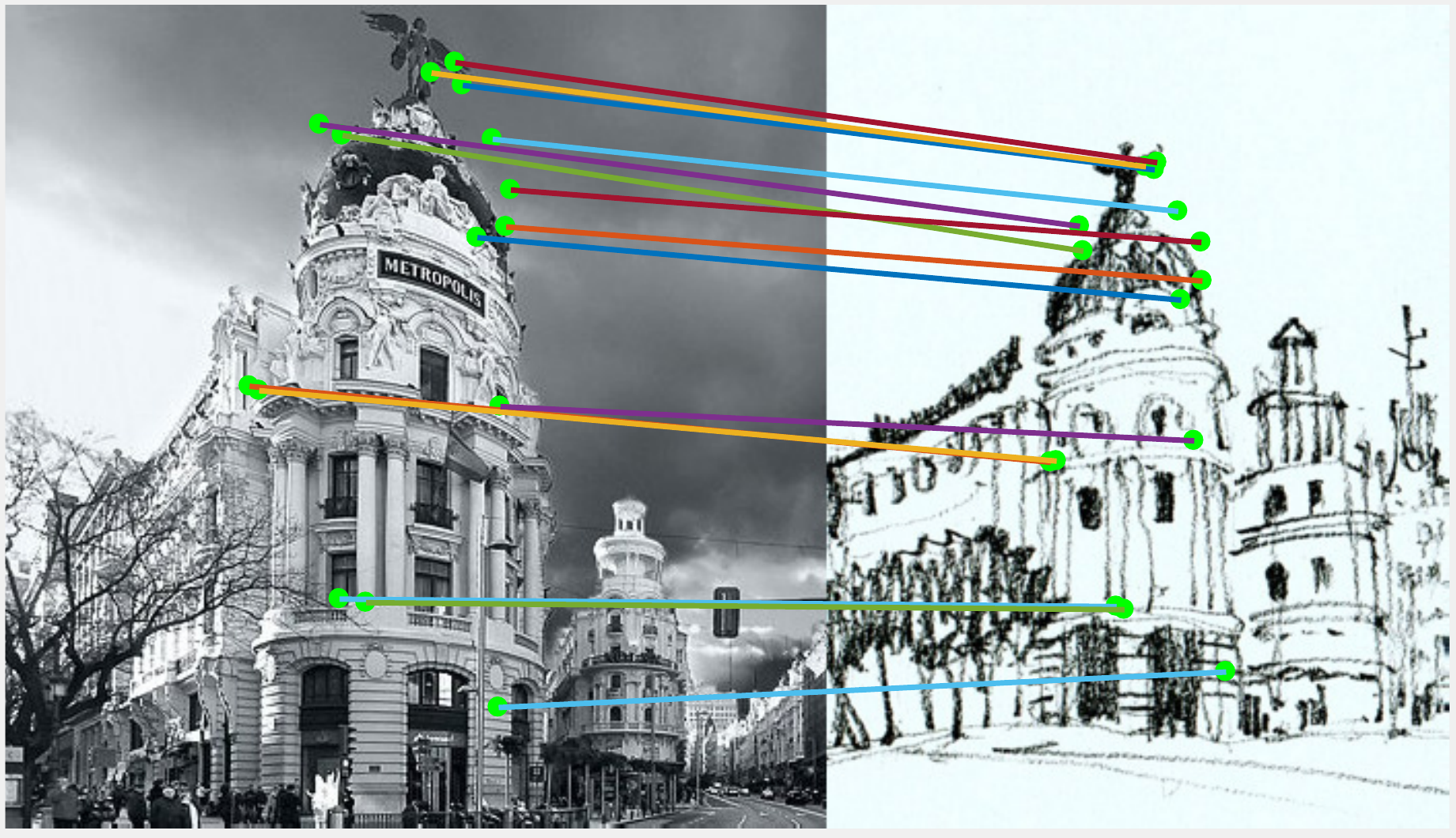}
    \includegraphics[width=0.8\columnwidth]{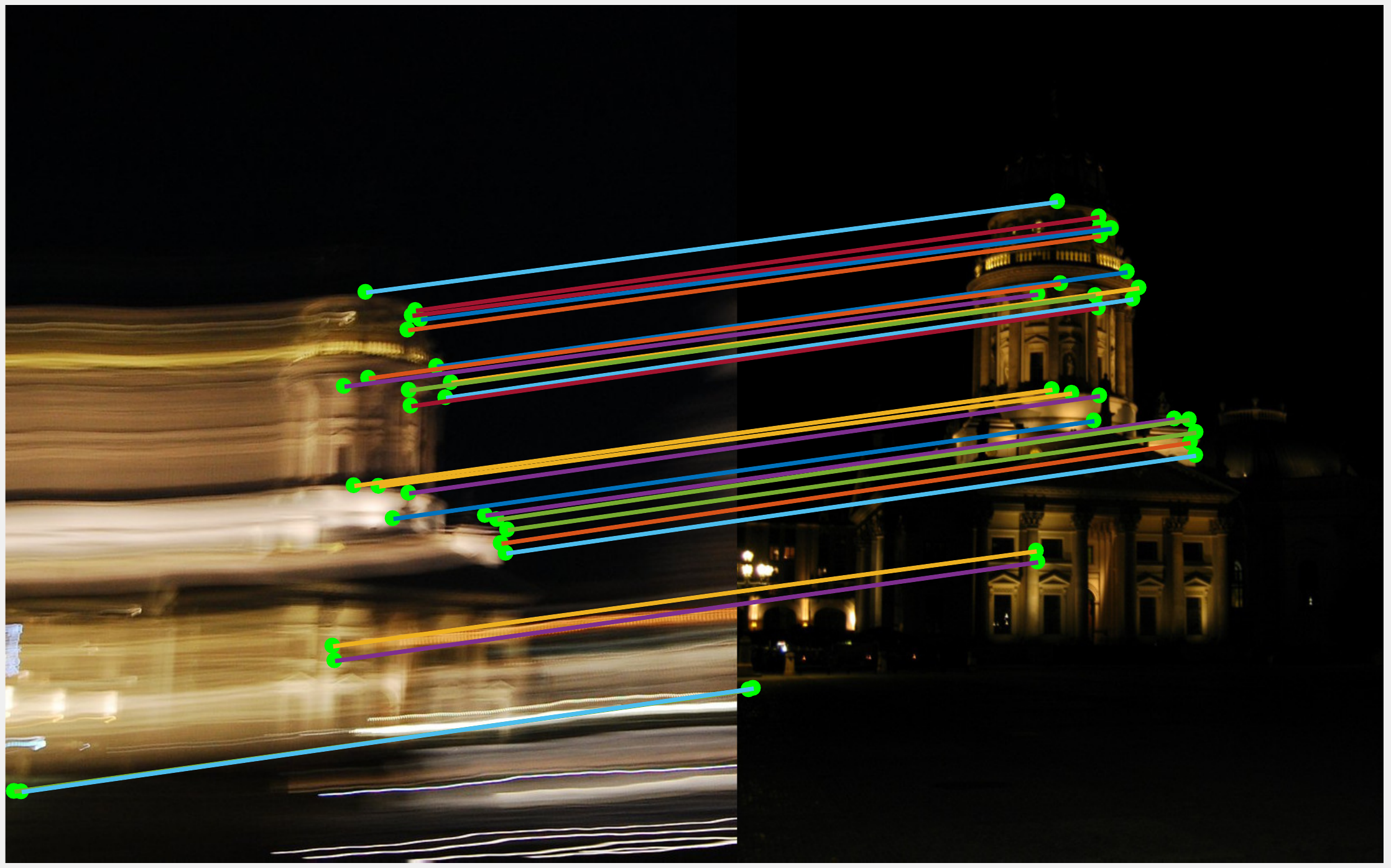}
    \includegraphics[width=0.8\columnwidth]{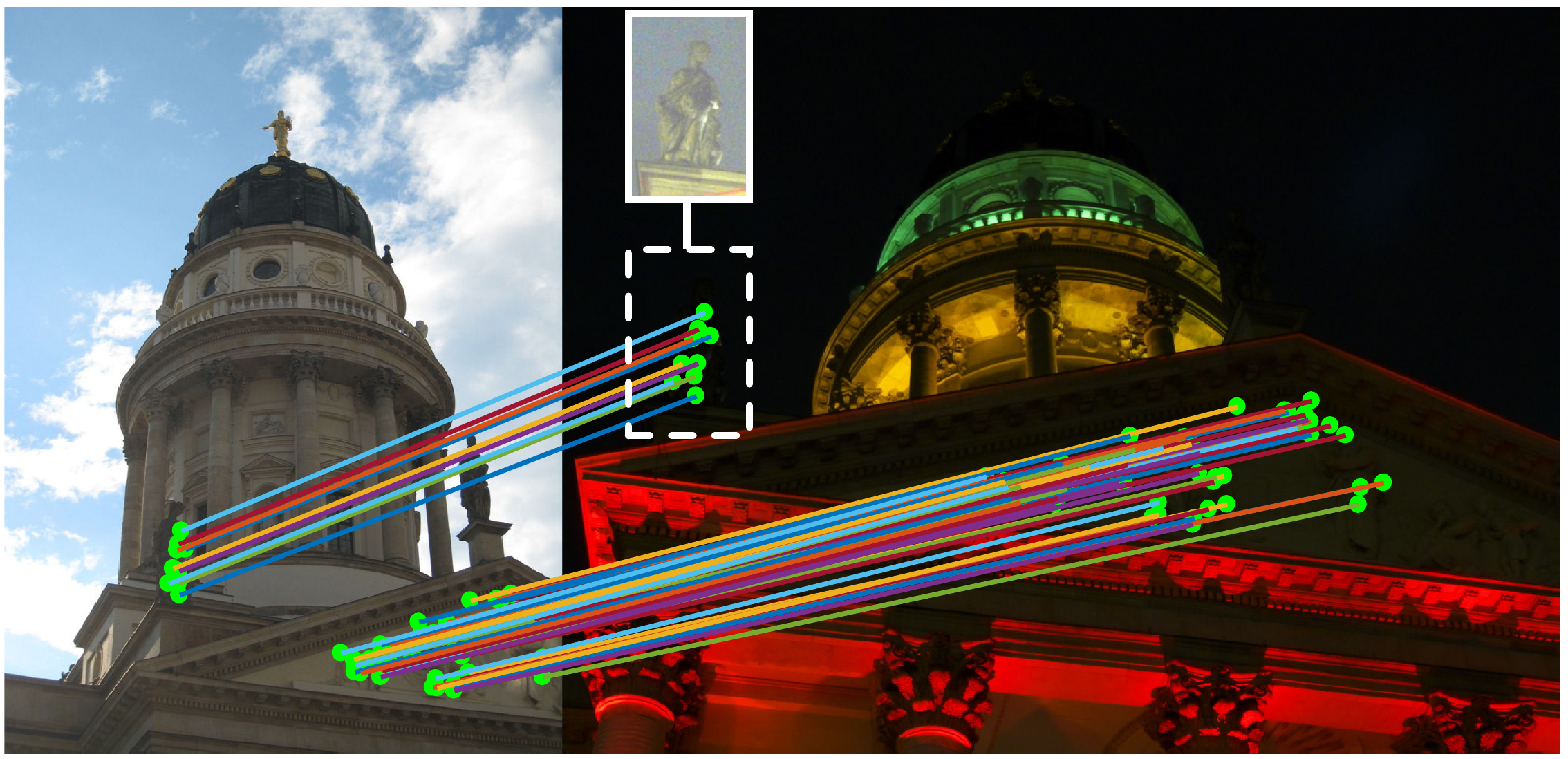}
    \caption{\small {\bf Examples of matches obtained by the D2-Net method.} The proposed method can find image correspondences even under significant appearance differences caused by strong changes in illumination such as day-to-night, changes in depiction style or under image degradation caused by motion blur.}
    \label{fig:teaser}
\end{figure}

Sparse local features~\cite{Mikolajczyk2005IJCV,Harris1988Combined,Bay2006ECCV,Mikolajczyk2004Scale,Savinov2017CVPR,Detone2018CVPRW,Yi2016LIFT,Mikolajczyk2005PAMI,Lowe2004Distinctive,Balntas2016Learning,Brown2011PAMI,Dong2015CVPR,SimoSerra2015ICCV,Tola2010,Simonyan2014PAMI} are a popular approach to correspondence estimation. 
These methods follow a \emph{detect-then-describe} approach that first applies a feature detector~\cite{Mikolajczyk2005IJCV,Harris1988Combined,Lowe2004Distinctive,Bay2006ECCV,Mikolajczyk2004Scale,Savinov2017CVPR,Detone2018CVPRW,Yi2016LIFT} to identify a set of keypoints or interest points. 
The detector then provides image patches extracted around the keypoints to the following feature description stage~\cite{Mikolajczyk2005PAMI,Bay2006ECCV,Lowe2004Distinctive,Balntas2016Learning,Yi2016LIFT,Brown2011PAMI,Dong2015CVPR,SimoSerra2015ICCV,Tola2010,Simonyan2014PAMI}. 
The output of this stage is a compact representation for each patch. 
Sparse local features offer a set of advantages: 
Correspondences can be \emph{matched efficiently} via (approximate) nearest neighbor search~\cite{Muja2014PAMI} and the Euclidean distance. 
Sparse features offer a \emph{memory efficient} representation and thus enable approaches such as Structure-from-Motion (SfM)~\cite{Schoenberger2018CVPR,Heinly2015CVPR} or visual localization ~\cite{Sattler2017CVPR,Li2012ECCV,Svarm2017PAMI} to scale. 
The keypoint detector typically considers low-level image information such as corners~\cite{Harris1988Combined} or blob-like structures ~\cite{Lowe2004Distinctive,Mikolajczyk2004Scale}. 
As such, local features can often be \emph{accurately localized} in an image, which is an important property for 3D reconstruction~\cite{Furukawa2010PAMI,Schoenberger2018CVPR}. 

Sparse local features have been applied successfully under a wide range of imaging conditions. 
However, they typically perform poorly under extreme appearance changes, \eg, between day and night~\cite{Zhou2016ECCVW} or seasons~\cite{Sattler2017Benchmarking}, or in weakly textured scenes~\cite{Taira2018InLoc}. 
Recent results indicate that a major reason for this observed drop in performance is the lack of repeatability in the keypoint detector:
While local descriptors consider larger patches and potentially encode higher-level structures, the keypoint detector only considers small image regions. 
As a result, the detections are unstable under strong appearance changes.
This is due to the fact that the low-level information used by the detectors is often significantly more affected by changes in low-level image statistics such as pixel intensities. 
Nevertheless, it has been observed that local descriptors can still be matched successfully even if keypoints cannot be detected reliably~\cite{Zhou2016ECCVW,Sattler2017Benchmarking,Taira2018InLoc,Torii2015CVPR}. 
Thus, approaches that forego the detection stage and instead densely extract descriptors perform much better in challenging conditions.
Yet, this gain in robustness comes at the price of higher matching times and memory consumption. 

In this paper, we aim at obtaining the best of both worlds, \ie, a sparse set of features that are robust under challenging conditions and  efficient to match and to store. 
To this end, we propose a \emph{describe-and-detect} approach to sparse local feature detection and description: 
Rather than performing feature detection early on based on low-level information, we propose to postpone the detection stage. 
We first compute a set of feature maps via a Deep Convolutional Neural Network (CNN). 
These feature maps are then used to compute the descriptors (as slices through all maps at a specific pixel position) \emph{and} to detect keypoints (as local maxima of the feature maps). 
As a result, the feature detector is tightly coupled with the feature descriptor. 
Detections thereby correspond to pixels with locally distinct descriptors that should be well-suited for matching.
At the same time, using feature maps from deeper layers of a CNN enables us to base both feature detection and description on higher-level information~\cite{Zeiler2014Visualizing}. 
Experiments show that our approach requires significantly less memory than dense methods. 
At the same time, it performs comparably well or even better under challenging conditions (\cf Fig.~\ref{fig:teaser}) such as day-night illumination changes~\cite{Sattler2017Benchmarking} and weakly textured scenes~\cite{Taira2018InLoc}. 
Our approach already achieves state-of-the-art performance without any training. 
It can be improved further by fine-tuning on a large dataset of landmark scenes~\cite{Li2018MegaDepth}.

Naturally, our approach has some drawbacks too: 
Compared to classical sparse features, our approach is less efficient due to the need to densely extract descriptors. 
Still, this stage can be done at a reasonable efficiency via a single forward pass through a CNN. 
Detection based on higher-level information inherently leads to more robust but less accurate keypoints -- yet, we show that our approach is still accurate enough for visual localization and SfM.

\begin{figure}
    \centering
    \includegraphics[width=0.9\textwidth]{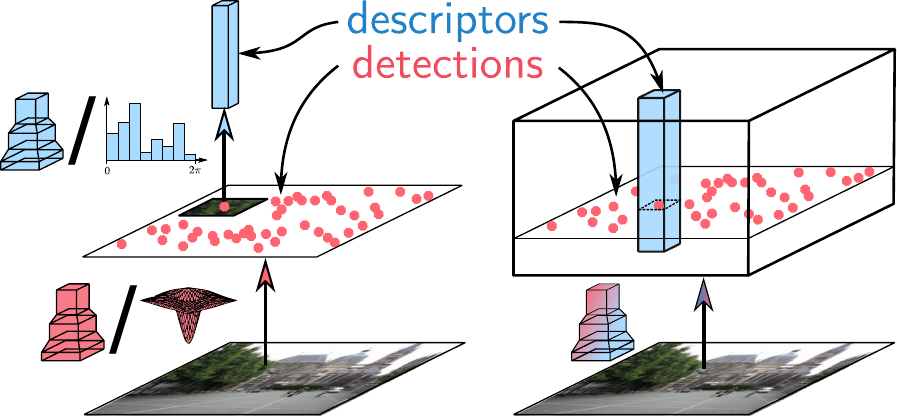}
    \begin{subfigure}[b]{0.445\textwidth}
        \caption{\centering{detect-then-describe}}
        \label{fig:approaches_a}
    \end{subfigure}
    \begin{subfigure}[b]{0.445\textwidth}
        \caption{\centering{detect-and-describe}}
        \label{fig:approaches_b}
    \end{subfigure}
    \caption{\small {\bf Comparison between different approaches for feature detection and description.} Pipeline (a) corresponds to different variants of the two-stage detect-then-describe approach. In contrast, our proposed pipeline (b) uses a single CNN which extracts dense features that serve as both descriptors and detectors. \label{fig:approaches}}
\end{figure}

%%%%%%%%%%%%%%%%%%%%%%%%%%%%%%%%%%%%%%%%%%%%%%%%%
\section{Related Work\label{sec:related_work}}
%%%%%%%%%%%%%%%%%%%%%%%%%%%%%%%%%%%%%%%%%%%%%%%%%

\PAR{Local features.} The most common approach to sparse feature extraction -- the detect-then-describe approach -- first performs feature detection~\cite{Mikolajczyk2004Scale,Mikolajczyk2005IJCV,Harris1988Combined,Lowe2004Distinctive,Bay2006ECCV} and then extracts a feature descriptor~\cite{Lowe2004Distinctive,Bay2006ECCV,rublee2011orb,calonder2010brief,leutenegger2011brisk} from a patch centered around each keypoint. 
The keypoint detector is typically responsible for providing robustness or invariance against effects such as scale, rotation, or viewpoint changes by normalizing the patch accordingly. 
However, some of these responsibilities might also be delegated to the descriptor~\cite{Zamir2016ECCV}. 
Fig.~\ref{fig:approaches_a} illustrates the common variations of this pipeline, from using hand-crafted detectors
~\cite{Mikolajczyk2004Scale,Mikolajczyk2005IJCV,Harris1988Combined,Lowe2004Distinctive,Bay2006ECCV}
and descriptors
~\cite{Lowe2004Distinctive,Bay2006ECCV,rublee2011orb,calonder2010brief,leutenegger2011brisk}, replacing either the descriptor
~\cite{Balntas2016Learning,Simonyan2014PAMI,SimoSerra2015ICCV}
or detector~\cite{Zhang2018CVPR,Savinov2017CVPR} with a learned alternative, or learning both the detector and descriptor~\cite{Yi2016LIFT, Ono2019LFNet}.
For efficiency, the feature detector often considers only small image regions~\cite{Yi2016LIFT} and typically focuses on low-level structures such as corners~\cite{Harris1988Combined} or blobs~\cite{Lowe2004Distinctive}.
The descriptor then captures higher level information in a larger patch around the keypoint. 
In contrast, this paper proposes a \emph{single branch describe-and-detect} approach to sparse feature extraction, as shown in Fig.~\ref{fig:approaches_b}. 
As a result, our approach is able to detect keypoints belonging to higher-level structures and locally unique descriptors. 
The work closest to our approach is SuperPoint~\cite{Detone2018CVPRW} as it also shares a deep representation between detection and description. However, they rely on different decoder branches which are trained independently with specific losses.
On the contrary, our method shares all parameters between detection and description and uses a joint formulation that simultaneously optimizes for both tasks.
Our experiments demonstrate that our describe-and-detect strategy performs significantly better under challenging conditions, \eg, when matching day-time and night-time images, than the previous approaches.

\PAR{Dense descriptor extraction and matching.}
An alternative to the detect-then-describe approach is to forego the detection stage and perform the description stage densely across the whole image~\cite{Schoenberger2018CVPR,Choy2016NIPS,Fathy2018ECCV,Savinov2017NIPS}. 
In practice, this approach has shown to lead to better matching results than sparse feature matching~\cite{Sattler2017Benchmarking,Taira2018InLoc,Zhou2016ECCVW}, particularly under strong variations in illumination~\cite{Zhou2016ECCVW}. 
This identifies the detection stage as a significant weakness in  detect-then-describe methods, which has motivated our approach.

\PAR{Image retrieval.}
The task of image retrieval~\cite{Arandjelovic2016NetVLAD,Noh2017Largescale,Tolias2016ICLR,Radenovic2018PAMI,Gordo2017IJCV,Torii2015CVPR} also deals with finding correspondences between images in challenging situations with strong illumination or viewpoint changes.
Several of these methods start by dense descriptor extraction~\cite{Arandjelovic2016NetVLAD,Noh2017Largescale,Tolias2016ICLR,Torii2015CVPR} and later aggregate these descriptors into a compact image-level descriptor for retrieval. 
Works most related to our approach are~\cite{Noh2017Largescale,Tolias2016ICLR}: 
\cite{Noh2017Largescale} develops an approach similar to ours, where an attention module is added on top of the dense description stage to perform keypoint selection. 
However, their method is designed to produce only a few reliable keypoints as to reduce the false positive matching rate during retrieval. 
Our experiments demonstrate that our approach 
performs significantly better for matching and camera localization;
\cite{Tolias2016ICLR} implicitly detects a set of keypoints as the global maxima of all feature maps, before pooling this information into a global image descriptor. 
\cite{Tolias2016ICLR} has inspired us to detect features as local maxima of feature maps.

\PAR{Object detection.}
The proposed describe-and-detect approach is also conceptually similar to modern approaches used in object detection~\cite{fasterrcnn,yolov1,liu2016ssd}. These methods also start by a dense feature extraction step, which is followed by the scoring of a set of region proposals. A non-maximal-suppression stage is then performed to select only the most locally-salient proposals with respect to a classification score.
Although these methods share conceptual similarities,  they target a very different task and cannot be applied directly to obtain pixel-wise image correspondences. 

This work builds on these previous ideas and proposes a method to perform joint detection and descriptions of keypoints, presented next.

\section{Joint Detection and Description Pipeline~\label{sec:method}}

\begin{figure*}
    \centering
    \includegraphics[width=0.9\textwidth]{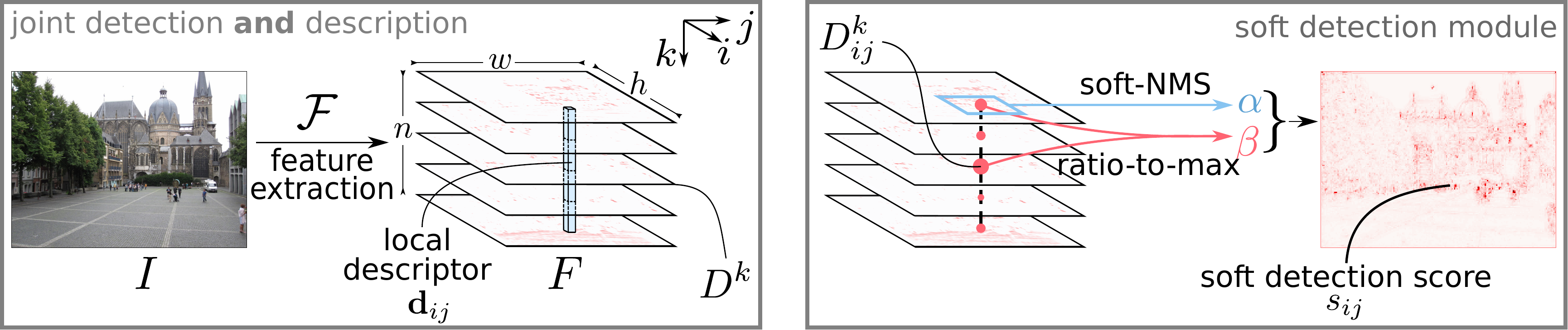}
    \caption{\small \textbf{Proposed detect-and-describe (D2) network.} A feature extraction \ac{CNN} $\mathcal{F}$ is used to extract feature maps that play a dual role: (i) local descriptors $\mathbf{d}_{ij}$ are simply obtained by traversing all the $n$ feature maps $D^k$ at a spatial position $(i,j)$; (ii) detections are obtained by performing a non-local-maximum suppression on a feature map followed by a non-maximum suppression across each descriptor - during training, keypoint detection scores $s_{ij}$ are computed from a soft local-maximum score $\alpha$ and a ratio-to-maximum score per descriptor $\beta$.\label{fig:model}}
\end{figure*}

Contrary to the classical detect-then-describe approaches, which use a two-stage pipeline, we propose to perform dense feature extraction to obtain a representation that is simultaneously a detector and a descriptor. Because both detector \emph{and} descriptor share the underlying representation, we refer to our approach as  D2. Our approach is illustrated in Fig.~\ref{fig:model}.

The first step of the method is to apply a \ac{CNN}  $\mathcal{F}$ on the input image $I$ to obtain a 3D tensor $F=\mathcal{F}(I), F \in \mathbb{R}^{h\times w\times n}$, where $h\times w$ is the spatial resolution of the feature maps and $n$ the number of channels. 

\subsection{Feature Description}

As in other previous work \cite{Taira2018InLoc, Noh2017Largescale, Rocco2017Convolutional}, the most straightforward interpretation of the 3D tensor $F$ is as a dense set of descriptor vectors $\mathbf{d}$: 
\begin{equation}
\mathbf{d}_{ij}=F_{ij:}, \mathbf{d}\in\mathbb{R}^n \enspace , 
\end{equation}
\noindent with $i=1,\dots,h$ and $j=1,\dots,w$.  These descriptor vectors can be readily compared between images to establish correspondences using the Euclidean distance. During the training stage, these descriptors will be adjusted such that the same points in the scene produce similar descriptors, even when the images contain strong appearance changes. In practice, we apply an L2 normalization on the descriptors prior to comparing them: $\hat{\mathbf{d}}_{ij}=\mathbf{d}_{ij}/\lVert\mathbf{d}_{ij}\rVert_2$.

\subsection{Feature Detection} A different interpretation of the 3D tensor $F$ is as a collection of 2D responses $D$~\cite{Tolias2016ICLR}:
\begin{equation}
D^k = F_{::k}, D^k\in\mathbb{R}^{h\times w} \enspace ,
\end{equation}
\noindent where $k=1,\dots,n$. In this interpretation, the feature extraction function $\mathcal{F}$ can be thought of as $n$ different feature detector functions $\mathcal{D}^k$, each producing a 2D response map $D^k$. These detection response maps are analogous to the \ac{DoG} response maps obtained in \ac{SIFT}~\cite{Lowe2004Distinctive} or to the cornerness score maps obtained in Harris' corner detector \cite{Harris1988Combined}. In our work, these raw scores are post-processed to select only a subset of locations as the output keypoints. This process is described next.

\PAR{Hard feature detection.} In traditional feature detectors such as \ac{DoG}, the detection map would be sparsified by performing a spatial non-local-maximum suppression. However, in our approach, contrary to traditional feature detectors, there exist multiple detection maps $D^k$ ($k=1,\dots,n$), and a detection can take place on any of them.
Therefore, for a point $(i,j)$ to be detected, we require:
\begin{align}
\begin{aligned}
(i,j)\text{ is a detection } \iff &
 D^k_{ij} \text{ is a local max. in } D^k \enspace,\\
& \text{with }k = \argmax_t D^t_{ij} \enspace .
\end{aligned}
\end{align}
\noindent Intuitively, for each pixel $(i, j)$, this corresponds to selecting the most preeminent detector $\mathcal{D}^k$ (channel selection), and then verifying whether there is a local-maximum  at position $(i,j)$ on that particular detector's response map $D^k$.

\PAR{Soft feature detection.} During training, the hard detection procedure described above is softened to be amenable for back-propagation. First, we define a soft local-max. score 
\begin{equation}
    \alpha^k_{i j} = \frac{\exp \left( D^k_{i j} \right)}{\sum_{(i^\prime, j^\prime) \in \mathcal{N}(i, j)} \exp \left( D^k_{i^\prime j^\prime} \right)} \enspace ,
\end{equation}
\noindent where $\mathcal{N}(i, j)$ is the set of $9$ neighbours of the pixel $(i, j)$ (including itself). Then, we define the soft channel selection, which computes a ratio-to-max. per descriptor that emulates channel-wise non-maximum suppression:
\begin{equation}
    \beta^k_{i j} = {D^k_{i j}}\big/{\max_{t}D^t_{i j}} \enspace .
\end{equation}
\noindent Next, in order to take both criteria into account, we maximize the product  of both scores across all feature maps $k$ to obtain a single score map:
\begin{equation}
    \gamma_{i j} = \max_k \left( \alpha^k_{i j} \beta^k_{i j} \right) \enspace .
\end{equation}
\noindent Finally, the soft detection score $s_{ij}$ at a pixel $(i, j)$ is obtained by performing an image-level normalization:
\begin{equation}
    s_{i j} = {\gamma_{i j}}\bigg/
    {\sum_{(i^\prime, j^\prime)}\gamma_{i^\prime j^\prime}} \enspace .
    \label{eq:softscore}
\end{equation}

\PAR{Multiscale Detection.} Although \ac{CNN} descriptors have a certain degree of scale invariance due to pre-training with data augmentations, they \emph{are not} inherently invariant to scale changes and the matching tends to fail in cases with a significant difference in viewpoint. 

In order to obtain features that are more robust to scale changes, we propose to use an image pyramid~\cite{adelson1984pyramid}, as typically done in hand-crafted local feature detectors~\cite{lindeberg1994scale,Lowe2004Distinctive,Mikolajczyk2004Scale} or even for some object detectors~\cite{felzenszwalb2010object}. This is only performed during \textit{test time}. 

Given the input image $I$, an image pyramid $I^\rho$ containing three different resolutions $\rho=0.5, 1, 2$ (corresponding to half resolution, input resolution, and double resolution) is constructed and used to extract feature maps $F^\rho$ at each resolution. Then, the larger image structures are propagated from the lower resolution feature maps to the higher resolution ones, in the following way:
\begin{equation}
    \tilde{F}^\rho = F^\rho + \sum_{\gamma<\rho} F^\gamma \enspace .
    \label{eq:mult_scale}
\end{equation}
Note that the feature maps $F^\rho$ have different resolutions. To enable the summation in~\eqref{eq:mult_scale}, feature maps $F^\gamma$ are resized to the resolution of $F^\rho$ using bilinear interpolation. 

Detections are obtained by applying the post-processing described above to the fused feature maps $\tilde{F}^\rho$. In order to prevent re-detecting features, we use the following response gating mechanism: Starting at the coarsest scale, we mark the detected positions; these masks are upsampled (nearest neighbor) to the resolutions of the next scales; detections falling into marked regions are then ignored.

\section{Jointly optimizing detection and description\label{sec:jointly_optimizing}}
This section describes the loss, the dataset used for training, and provides implementation details.

\subsection{Training loss\label{section:loss}}
In order to train the proposed model, which uses a single \ac{CNN} $\mathcal{F}$ for both detection and description, we require an appropriate loss $\mathcal{L}$ that jointly optimizes the detection and description objectives. In the case of detection, we want  keypoints to be \emph{repeatable} under changes in viewpoint or illumination. In the case of description, we want descriptors to be \emph{distinctive}, so that they are not mismatched. To this end, we propose an extension to the triplet margin ranking loss, which has been successfully used for descriptor learning~\cite{Balntas2016Learning, Mishchuk2017Working}, to also account for the detection stage. We will first review the triplet margin ranking loss, and then present our extended version for joint detection and description.

Given a pair of images $(I_1,I_2)$ and a correspondence $c: A \leftrightarrow B$ between them (where $A\in I_1$, $B\in I_2$), our version of the triplet margin ranking loss seeks to minimize the distance of the corresponding descriptors $\hat{\mathbf{d}}^{(1)}_A$ and $\hat{\mathbf{d}}^{(2)}_B$, while maximizing the distance to other confounding descriptors $\hat{\mathbf{d}}^{(1)}_{N_1}$ or $\hat{\mathbf{d}}^{(2)}_{N_2}$ in either image, which might exist due to similarly looking image structures. To this end, we define the \emph{positive} descriptor distance $p(c)$ between the corresponding descriptors as:
\begin{equation}
    p(c) = \lVert \hat{\mathbf{d}}^{(1)}_A - \hat{\mathbf{d}}^{(2)}_B \rVert_2  \enspace ,
\end{equation}
The \emph{negative} distance $n(c)$, which accounts for the most confounding descriptor for either $\hat{\mathbf{d}}^{(1)}_A$ or $\hat{\mathbf{d}}^{(2)}_B$, is defined as:
\begin{equation}
    n(c) = \min \left(
   \lVert \hat{\mathbf{d}}^{(1)}_A - \hat{\mathbf{d}}^{(2)}_{N_2} \rVert_2,
   \lVert \hat{\mathbf{d}}^{(1)}_{N_1} - \hat{\mathbf{d}}^{(2)}_B \rVert_2
   \right) \enspace ,
\end{equation}
\noindent where the negative samples $\mathbf{d}^{(1)}_{N_1}$ and $\mathbf{d}^{(2)}_{N_2}$ are the hardest negatives that lie outside of a square local neighbourhood of the correct correspondence:
\begin{equation}
    N_1 = \argmin_{P\in I_1} \lVert \hat{\mathbf{d}}^{(1)}_P - \hat{\mathbf{d}}^{(2)}_B \rVert_2 \text{ s.t. } \lVert P-A \rVert_\infty > K \enspace ,
\end{equation}
\noindent and similarly for $N_2$.
The triplet margin ranking loss for a margin $M$ can be then defined as:
\begin{equation}
    m(c) = \max \left( 0, M + p(c)^2 - n(c)^2 \right) \enspace.
    \label{eq:triplet_margin}
\end{equation}
Intuitively, this triplet margin ranking loss seeks to enforce the \emph{distinctiveness} of descriptors by penalizing any confounding descriptor that would lead to a wrong match assignment.
In order to additionally seek for the \emph{repeatability} of detections, an detection term is added to the triplet margin ranking loss in the following way:
\begin{equation}
    \mathcal{L}(I_1, I_2) = \sum_{c\in\mathcal{C}} \frac{s^{(1)}_c s^{(2)}_c}{\sum_{q\in\mathcal{C}} s^{(1)}_{q} s^{(2)}_{q}} m(p(c), n(c)) \ ,
    \label{eq:loss}
\end{equation}
where $s^{(1)}_c$ and $s^{(2)}_c$ are the soft detection scores~\eqref{eq:softscore} at points $A$ and $B$ in $I_1$ and $I_2$, respectively, and $\mathcal{C}$ is the set of all correspondences between $I_1$ and $I_2$. 

The proposed loss produces a weighted average of the margin terms $m$ over all matches based on their detection scores. Thus, in order for the loss to be minimized, the most distinctive correspondences (with a lower margin term) will get higher relative scores and vice-versa - correspondences with higher relative scores are encouraged to have a similar descriptors distinctive from the rest.

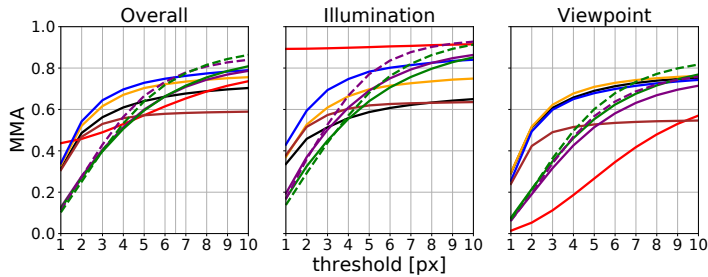
\begin{figure*}[t]
    \begin{subfigure}[t]{0.60\textwidth}
    \centering
    \raisebox{-0.5\height}{\includegraphics[width=0.9\textwidth]{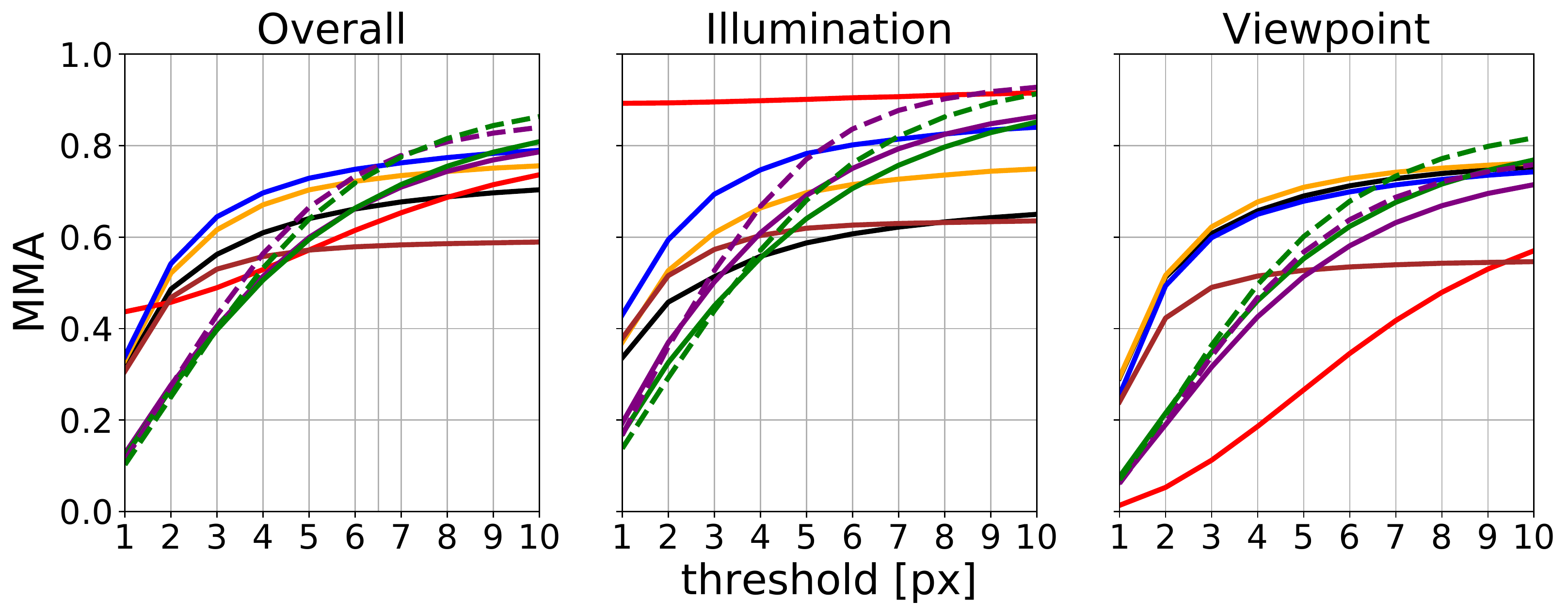}}
    \end{subfigure}
    ~
    \begin{subfigure}[t]{0.35\textwidth}
    \footnotesize
    \centering
    \begin{tabular}{@{}c@{\hspace{0.5mm}}l@{\hspace{1mm}}c@{\hspace{1mm}}c@{}}
    \toprule
    \multicolumn{2}{@{}l}{\textbf{Method}} & \textbf{\# Features} & \textbf{\# Matches} \\ 
    \midrule
    \tikz[baseline]{\draw[color=Black,line width=0.7mm] (0,.5ex)--++(.45,0) ;} & Hes. det. + RootSIFT & $6.7$K & $2.8$K \\ 
    \tikz[baseline]{\draw[color=Dandelion,line width=0.7mm] (0,.5ex)--++(.45,0) ;} & HAN + HN++~\cite{Mishkin2018ECCV, Mishchuk2017Working} & $3.9$K & $2.0$K \\
    \tikz[baseline]{\draw[color=Brown,line width=0.7mm] (0,.5ex)--++(.45,0) ;} & LF-Net~\cite{Ono2019LFNet} & $0.5$K & $0.2$K \\
    \tikz[baseline]{\draw[color=Blue,line width=0.7mm] (0,.5ex)--++(.45,0) ;} & SuperPoint~\cite{Detone2018CVPRW} & $1.7$K & $0.9$K \\ 
    \tikz[baseline]{\draw[color=Red,line width=0.7mm] (0,.5ex)--++(.45,0) ;} & DELF~\cite{Noh2017Largescale} & $4.6$K & $1.9$K \\
    \tikz[baseline]{\draw[color=Purple,line width=0.7mm] (0,.5ex)--++(.45,0) ;} & D2 SS (ours) & $3.0$K & $1.2$K \\
    \tikz[baseline]{\draw[color=Green,line width=0.7mm] (0,.5ex)--++(.45,0) ;} & D2 MS (ours) & $4.9$K & $1.7$K \\
    \tikz[baseline]{\draw[color=Purple,line width=0.7mm,densely dashed] (0,.5ex)--++(.45,0) ;} & D2 SS Trained (ours) & $6.0$K & $2.5$K \\
    \tikz[baseline]{\draw[color=Green,line width=0.7mm,densely dashed] (0,.5ex)--++(.45,0) ;} & D2 MS Trained (ours) & $8.3$K & $2.8$K \\
    \bottomrule 
    \end{tabular}
    \end{subfigure}

    \vspace*{-3mm}
    \caption{\small \textbf{Evaluation on HPatches~\cite{HPATCHES} image pairs.} For each method, the mean matching accuracy (MMA) as a function of the matching threshold (in pixels) is shown. We also report the mean number of detected features and the mean number of mutual nearest neighbor matches. Our approach achieves the best overall performance after a threshold of $6.5$px, both using a single (SS) and multiple (MS) scales. 
    }
    \label{fig:hsequences-results}
\end{figure*}

\subsection{Training Data\label{section:training-data}}

To generate training data on the level of pixel-wise correspondences, we used the MegaDepth dataset \cite{Li2018MegaDepth} consisting of $196$ different scenes reconstructed from \num{1070468} internet photos using COLMAP \cite{schoenberger2016sfm, schoenberger2016mvs}. The authors provide camera intrinsics / extrinsics and depth maps from Multi-View Stereo for \num{102681} images.

In order to extract the correspondences, we first considered all pairs of images with at least $50\%$ overlap in the sparse SfM point cloud. For each pair, all points of the second image with depth information were projected into the first image. A depth-check with respect to the depth map of the first image was run to remove occluded pixels. In the end, we obtained \num{327036} image pairs. 
This dataset was split in a validation dataset with \num{18149} image pairs (from $78$ scenes, each with less than $500$ image pairs) and a training dataset from the remaining $118$ scenes. 

\subsection{Implementation details}

The VGG16 architecture \cite{Simonyan2014Very}, pretrained on ImageNet~\cite{Deng2009ImageNet} and truncated after the \texttt{conv4\_3} layer, was used to initialize the feature extraction network $\mathcal{F}$.

\PAR{Training.} The last layer of the dense feature extractor (\texttt{conv4\_3}) was fine-tuned for $50$ epochs using Adam~\cite{kingma2014adam} with an initial learning rate of $10^{-3}$, which was further divided by $2$ every $10$ epochs. A fixed number ($100$) of random image pairs from each scene are used for training at every epoch in order to compensate the scene imbalance present in the dataset. For each pair, we selected a random $256 \times 256$ crop centered around one correspondence. We use a batch size of $1$ and make sure that the training pairs present at least $128$ correspondences in order to obtain meaningful gradients.

\PAR{Testing.} At test time, in order to increase the resolution of the feature maps, the last pooling layer (\texttt{pool3}) from $\mathcal{F}$ with a stride of $2$ is replaced by an average pooling layer with a stride of $1$. Then, the subsequent convolutional layers (\texttt{conv4\_1} to \texttt{conv4\_3}) are replaced with dilated convolutions \cite{holschneider1990real} with a rate of $2$, so that their receptive field remains unchanged.
With these modifications, the obtained feature maps have a resolution of one fourth of the input resolution, which allows for more tentative keypoints and a better localization. 
The position of the detected keypoints is improved using a local refinement at feature map level following the approach used in SIFT~\cite{Lowe2004Distinctive}. The descriptors are then bilinearly interpolated at the refined positions.

\vskip4pt Our implementation will be available at \url{https://github.com/mihaidusmanu/d2-net}.

\section{Experimental Evaluation\label{sec:experimental_eval}}
The main motivation behind our work was to develop a local features approach that is able to better handle challenging conditions.
Firstly, we evaluate our method on a standard image matching task based on sequences with illumination or viewpoint changes. Then, we present the results of our method in two more complex computer vision pipelines: 3D reconstruction and visual localization. In particular, the visual localization task is evaluated under extremely challenging conditions such as registering night-time images against 3D models generated from day-time imagery~\cite{Sattler2017Benchmarking,Sattler2012Image} and localizing images in challenging indoor scenes~\cite{Taira2018InLoc} dominated by weakly textured surfaces and repetitive structures. Qualitative examples of the results of our method are presented in Fig.~\ref{fig:teaser}. %{\bf 
{Please see the supplementary material for additional qualitative examples.}

\subsection{Image Matching}
\label{sec:experiments:matching}

In a first experiment, we consider a standard image matching scenario where given two images we would like to extract and match features between them. 
For this experiment, we use the sequences of full images provided by the HPatches dataset~\cite{HPATCHES}. 
Out of the 116 available sequences collected from various datasets~\cite{ASIFT,PHOS,HAN,DTU,AMOS,Mikolajczyk2005PAMI,HPATCHES}, we selected 108.\footnote{We left out sequences with an image resolution beyond $1200 \times 1600$ pixels as not all methods were able to handle this resolution.}
Each sequence consists of 6 images of progressively larger illumination (52 sequences without viewpoint changes) or viewpoint changes (56 sequences without illumination changes). 
For each sequence, we match the first against all other images, resulting in 540 pairs. 

\PAR{Evaluation protocol.}
For each image pair, we match the features extracted by each method using nearest neighbor search, accepting only mutual nearest neighbors.
A match is considered correct if its reprojection error, estimated using the homographies provided by the dataset, is below a given matching threshold. % (in pixels). 
We vary the threshold and record the mean matching accuracy (MMA) \cite{Mikolajczyk2005PAMI} over all pairs, \ie, the average percentage of correct matches per image pair.

As baselines for the classical detect-then-describe strategy, we use RootSIFT~\cite{Arandjelovic2012Three,Lowe2004Distinctive} with the Hessian Affine keypoint detector~\cite{Mikolajczyk2004Scale}, a variant using a learned shape estimator (HesAffNet~\cite{Mishkin2018ECCV} - HAN) and descriptor (HardNet++~\cite{Mishchuk2017Working} - HN++\footnote{HardNet++ was trained on the HPatches dataset~\cite{HPATCHES}.}), and an end-to-end trainable variant (LF-Net~\cite{Ono2019LFNet}).
% In addition, w
We also compare against SuperPoint~\cite{Detone2018CVPRW} and DELF~\cite{Noh2017Largescale}, which are conceptually more similar to our approach.

\PAR{Results.}
Fig.~\ref{fig:hsequences-results} shows results for illumination and viewpoint changes, as well as mean accuracy over both conditions. 
For each method, we also report the mean number of detected features and the mean number of mutual nearest neighbor matches per image.
As can be seen, our method achieves the best overall performance for matching thresholds of $6.5$ pixels or more. 

DELF does not refine its keypoint positions - thus, detecting the same pixel positions at feature map level yields perfect accuracy for strict thresholds. Even though powerful for the illumination sequences, the downsides of their method when used as a local feature extractor can be seen in the viewpoint sequences. 
For LF-Net, increasing the number of keypoints to more than the default value ($500$) worsened the results. 
However, \cite{Ono2019LFNet} does not enforce that matches are mutual nearest neighbors and we suspect that their method is not suited for this type of matching.

As can be expected, our method performs worse than detect-then-describe approaches for stricter matching thresholds: 
The latter use detectors firing at low-level blob-like structures, which are inherently better localized than the higher-level features used by our approach. 
At the same time, our features are also detected at the lower resolution of the CNN features.

We suspect that the inferior performance for the sequences with viewpoint changes is due to a major bias in our training dataset - roughly $90\%$ of image pairs have a change in viewpoint lower than $20\degree$ (measured as the angle between the principal axes of the two cameras).

The proposed pipeline for multiscale detection improves the viewpoint robustness of our descriptors, but it also adds more confounding descriptors that negatively affect the robustness to illumination changes.

\subsection{3D Reconstruction}
\label{sec:experiments:3d}

In a second experiment, we evaluate the performance of our proposed describe-and-detect approach in the context of 3D reconstruction. 
This task requires well-localized features and might thus be challenging for our method.

For evaluation, we use three medium-scale internet-collected datasets with a significant number of different cameras and conditions (Madrid Metropolis, Gendarmenmarkt and Tower of London \cite{wilson2014robust}) from a recent local feature evaluation benchmark~\cite{Schonberger2017Comparative}. 
All three datasets are small enough to allow exhaustive image matching, thus avoiding the need for using image retrieval.

\PAR{Evaluation protocol.} We follow the protocol defined by~\cite{Schonberger2017Comparative} and first run SfM~\cite{schoenberger2016sfm}, followed by Multi-View Stereo (MVS)~\cite{schoenberger2016mvs}. 
For the SfM models, we report the number of images and 3D points, the mean track lengths of the 3D points, and the mean reprojection error. For the MVS models, we report the number of dense points. Except for the reprojection error, larger numbers are better. 
We use RootSIFT~\cite{Arandjelovic2012Three, Lowe2004Distinctive} (the best perfoming method according to the benchmark's website) and GeoDesc~\cite{Luo2018GeoDesc}, a state-of-the-art trained descriptor\footnote{In contrast to~\cite{Luo2018GeoDesc}, we use the ratio test for matching with the threshold suggested by the authors - $0.89$.} as baselines. Both follow the detect-then-describe approach to local features.

\PAR{Results.}
Tab.~\ref{tab:lfe} shows the results of our experiment. Overall, the results show that our approach performs on par with state-of-the-art local features on this task. 
This shows that, even though our features are less accurately localized compared to detect-then-describe approaches, they are sufficiently accurate for the task of SfM as our approach is still able to register a comparable number of images.

Our method reconstructs fewer 3D points due to the strong ratio test filtering~\cite{Lowe2004Distinctive} of the matches that is performed in the 3D reconstruction pipeline. While this filtering is extremely important to remove incorrect matches and prevent incorrect registrations, we noticed that for our method it also removes an important number of correct matches ($20$\%--$25$\%)\footnote{Please see the supplementary material for additional details.}, as the loss used for training our method does not take this type of filtering into account.

\begin{figure*}[t]
    \begin{subfigure}[t]{0.45\textwidth}
    \centering
    \raisebox{-0.5\height}{\includegraphics[width=\textwidth]{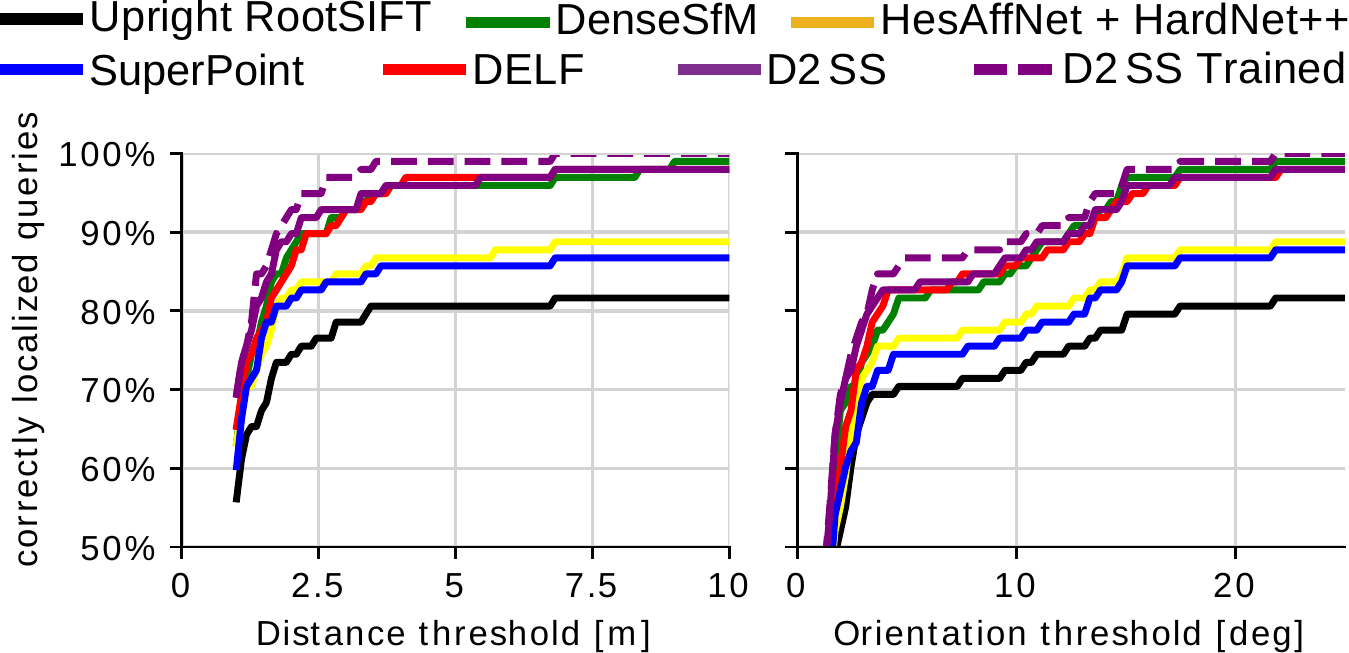}}
    \end{subfigure}
    ~
    \begin{subfigure}[t]{0.50\textwidth}
    \footnotesize{
    \centering
    \begin{tabular}{@{}l@{\hspace{1mm}}c@{\hspace{2mm}} c@{\hspace{0.75mm}}c@{\hspace{0.75mm}}c@{\hspace{0.75mm}}c@{}}
    \toprule
    & & \multicolumn{4}{c}{\textbf{Correctly localized queries (\%)}} \\
    \textbf{Method} & \textbf{\# Features} & $0.5$m, $2^\circ$ & $1.0$m, $5^\circ$ & $5.0$m, $10^\circ$ & $10$m, $25^\circ$ \\
    \midrule
    Upright RootSIFT~\cite{Lowe2004Distinctive} & $11.3$K & $36.7$ & $54.1$ & $72.5$ & $81.6$ \\
    DenseSfM~\cite{Sattler2017Benchmarking} & $7.5$K / $30$K & $39.8$ & $60.2$ & $84.7$ & $99.0$ \\
    HAN\,+\,HN++~\cite{Mishchuk2017Working, Mishkin2018ECCV} & $11.5$K & $39.8$ & $61.2$ & $77.6$ & $88.8$\\
    SuperPoint~\cite{Detone2018CVPRW} & $6.6$K & $42.8$ & $57.1$ & $75.5$ & $86.7$\\
    DELF~\cite{Noh2017Largescale} & $11$K & $38.8$ & $62.2$ & $85.7$ & $98.0$ \\ \hline
    D2 SS (ours) & $7$K & $41.8$ & $66.3$ & $85.7$ & $98.0$ \\
    D2 MS (ours) & $11.4$K & $43.9$ & $\mathbf{67.3}$ & $87.8$ & $99.0$ \\
    D2 SS Trained (ours) & $14.5$K & $\mathbf{44.9}$ & $66.3$ & $\mathbf{88.8}$ & $\mathbf{100}$ \\
    D2 MS Trained (ours) & $19.3$K & $\mathbf{44.9}$ & $64.3$ & $\mathbf{88.8}$ & $\mathbf{100}$ \\
    \bottomrule 
    \end{tabular}
    }
    \end{subfigure}
    \vspace*{-2mm}
    \caption{\small \textbf{Evaluation on the Aachen Day-Night dataset~\cite{Sattler2012Image,Sattler2017Benchmarking}.} We report the percentage of images registered within given error thresholds. Our approach improves upon state-of-the art methods by a significant margin under strict pose thresholds.}
    \label{fig:aachen}
\end{figure*}

\subsection{Localization under Challenging Conditions}
\label{sec:experiments:challenging}
The previous experiments showed that our approach performs comparable with the state-of-the-art in standard applications. 
In a final experiment, we show that our approach sets the state-of-the-art for sparse features under two very challenging conditions: Localizing images under severe illumination changes and in complex indoor scenes.

\begin{table}
    \scriptsize{
    \begin{tabular}{@{}l@{\hspace{2mm}}l@{\hspace{1.5mm}}c@{\hspace{1.2mm}}c@{\hspace{1.2mm}}c@{\hspace{1.2mm}}c@{\hspace{1.2mm}}c@{}}
    \toprule
     & & \textbf{\#Reg. } & \textbf{\# Sparse.} & \textbf{Track} & \textbf{Reproj.} & \textbf{\# Dense}\\
    \textbf{Dataset} & \textbf{Method} & \textbf{Images} & \textbf{Points} & \textbf{Length} & \textbf{Error} & \textbf{Points} \\ 
    \midrule
	\textbf{Madrid} & RootSIFT \cite{Arandjelovic2012Three, Lowe2004Distinctive} & $500$ & $116$K & $6.32$ & $\mathbf{0.60}$\textbf{px} & $\mathbf{1.82}$\textbf{M} \\
	\textbf{Metropolis} & GeoDesc \cite{Luo2018GeoDesc} & $495$ & $\textbf{144}$\textbf{K} & $5.97$ & $0.65$px & $1.56$M \\
	$1344$ images & D2 MS (ours) & $\mathbf{501}$ & $84$K & $6.33$ & $1.28$px & $1.46$M \\ 
	& D2 MS trained (ours) & $495$ & $\textbf{144}$\textbf{K} & $\mathbf{6.39}$ & $1.35$px & $1.46$M \\
	\midrule
	\textbf{Gendarmen-} & RootSIFT~\cite{Arandjelovic2012Three, Lowe2004Distinctive} & $1035$ & $338$K & $5.52$ & $\mathbf{0.69}$\textbf{px} & $\mathbf{4.23}$\textbf{M} \\
	\textbf{markt} & GeoDesc~\cite{Luo2018GeoDesc} & $1004$ & $\mathbf{441}$\textbf{K} & $5.14$ & $0.73$px & $3.88$M\\
	$1463$ images & D2 MS (ours) & $\mathbf{1053}$ & $250$K & $5.08$ & $1.19$px & $3.49$M \\ 
	& D2 MS trained (ours) & $965$ & $310$K & $\mathbf{5.55}$ & $1.28$px & $3.15$M \\
	\midrule
	\textbf{Tower of} & RootSIFT~\cite{Arandjelovic2012Three, Lowe2004Distinctive} & $\mathbf{804}$ & $239$K & $\mathbf{7.76}$ & $\mathbf{0.61}$\textbf{px} & $\mathbf{3.05}$\textbf{M} \\
	\textbf{London} & GeoDesc~\cite{Luo2018GeoDesc} & $776$ & $\mathbf{341}$\textbf{K} & $6.71$ & $0.63$px & $2.73$M \\
	$1576$ images & D2 MS (ours) & $785$ & $180$K & $5.32$ & $1.24$px & $2.73$M \\ 
	& D2 MS trained (ours) & $708$ & $287$K & $5.20$ & $1.34$px & $2.86$M \\
	\bottomrule
    \end{tabular}
    }
    \caption{\small \textbf{Evaluation on the Local Feature Evaluation Benchmark~\cite{Schonberger2017Comparative}.} 
    Each method is used for the 3D reconstruction of each scene and different statistics are reported. Overall, our method obtains a comparable performance with respect to SIFT and its trainable counterparts, despite using less well-localized keypoints.
    }
    \label{tab:lfe}
\end{table}

\PAR{Day-Night Visual Localization.} 
We evaluate our approach on the {Aachen Day-Night} dataset \cite{Sattler2017Benchmarking,Sattler2012Image} in a local reconstruction task~\cite{Sattler2017Benchmarking}: 
For each of the 98 night-time images contained in the dataset, up to 20 relevant day-time images with known camera poses are given. 
After exhaustive feature matching between the day-time images in each set, their known poses are used to triangulate the 3D structure of the scenes. 
Finally, these resulting 3D models are used to localize the night-time query images. 
This task was proposed in~\cite{Sattler2017Benchmarking} to evaluate the perfomance of local features in the context of long-term localization without the need for a specific localization pipeline. 

We use the code and evaluation protocol from~\cite{Sattler2017Benchmarking} and report the percentage of night-time queries localized within a given error bound on the estimated camera position and orientation. 
We compare against upright RootSIFT descriptors extracted from DoG keypoints~\cite{Lowe2004Distinctive}, HardNet++ descriptors with HesAffNet features~\cite{Mishchuk2017Working, Mishkin2018ECCV}, DELF~\cite{Noh2017Largescale}, SuperPoint~\cite{Detone2018CVPRW} and DenseSfM~\cite{Sattler2017Benchmarking}. DenseSfM densely extracts CNN features using VGG16, followed by dense hierarchical matching (\texttt{conv4} then \texttt{conv3}). 

For all methods with a threshold controlling the number of detected features (i.e. HAN + HN++, DELF, and SuperPoint), we employed the following tuning methodology: Starting from the default value, we increased and decreased the threshold gradually stopping as soon as the results started declining. Stricter localization thresholds were considered more important than looser ones. We reported the best results each method was able to achieve.

As can be seen from Fig.~\ref{fig:aachen}, our approach performs better than all baselines, especially for strict accuracy thresholds for the estimated pose. 
Our sparse feature approach even outperforms DenseSfM, even though the later is using significantly more features (and thus time and memory). The results clearly validate our describe-and-detect approach as it significantly outperforms %methods based on the classical 
detect-then-describe methods in this highly challenging scenario. 
The results also show that the lower keypoint accuracy of our approach does not prevent it from being used for applications aiming at estimating accurate camera poses.

\begin{table}
    \footnotesize
    \centering
    \begin{tabular}{@{}l@{\hspace{1mm}}c@{\hskip 1.5mm}c@{\hskip 1mm}c}
    \toprule
    & \multicolumn{3}{@{\hspace{-1mm}}l@{}}{\textbf{Localized queries (\%)}} \\
    \textbf{Method} & $0.25$m & $0.5$m & $1.0$m \\
    \midrule
    \textbf{Direct PE} - Aff. RootSIFT \cite{Mikolajczyk2004Scale,Lowe2004Distinctive,Arandjelovic2012Three} & $18.5$ & $26.4$ & $30.4$ \\
    \textbf{Direct PE} - D2 MS (ours) & $\mathbf{27.7}$ & $\mathbf{40.4}$ & $\mathbf{48.6}$ \\ \midrule
    \textbf{Sparse PE} - Aff. RootSIFT -- $5$MB & $21.3$ & $32.2$ & $44.1$ \\
    \textbf{Sparse PE} - D2 MS (ours) -- $15$MB & $\mathbf{35.0}$ & $\mathbf{48.6}$ & $\mathbf{62.6}$ \\
    \textbf{Dense PE} \cite{Taira2018InLoc} -- $44$MB & $\mathbf{35.0}$ & $46.2$ & $58.1$ \\ \midrule
    \textbf{Sparse PE} - Aff. RootSIFT + \textbf{Dense PV} & $29.5$ & $42.6$ & $54.5$ \\
    \textbf{Sparse PE} - D2 MS + \textbf{Dense PV} (ours) & $38.0$ & $54.1$ & $65.4$ \\
    \textbf{Dense PE} + \textbf{Dense PV} (= \textbf{InLoc}) \cite{Taira2018InLoc} & $\mathbf{41.0}$ & $\mathbf{56.5}$ & $\mathbf{69.9}$ \\ \midrule
    \textbf{InLoc} + D2 MS (ours) & $\mathbf{43.2}$ & $\mathbf{61.1}$ & $\mathbf{74.2}$ \\
    \bottomrule 
    \end{tabular}
    \caption{\small \textbf{Evaluation on the InLoc dataset~\cite{Taira2018InLoc}.} 
    Our method outperforms SIFT by a large margin in both \textbf{Direct PE} and \textbf{Sparse PE} setups. It also outperforms the dense matching \textbf{Dense PE} method when used alone, while requiring less memory during pose estimation. By a combined approach of D2 and \textbf{InLoc} we obtained a new state-of-the art on this dataset.
    }
    \label{tab:inloc-results}
\end{table}

\PAR{Indoor Visual Localization.}
We also evaluate our approach on the {InLoc} dataset~\cite{Taira2018InLoc}, a recently proposed benchmark dataset for large-scale indoor localization. 
The dataset is challenging due to its sheer size ($\sim$10k database images covering two buildings), strong differences in viewpoint and / or illumination between the database and query images, and changes in the scene over time.

For this experiment, we integrated our features into two variants of the pipeline proposed in~\cite{Taira2018InLoc}, using the code released by the authors. 
The first variant, \textbf{Direct \ac{PE}}, matches features between the query image and the top-ranked database image found by image retrieval~\cite{Arandjelovic2016NetVLAD} and uses these matches for pose estimation. 
In the second variant, \textbf{Sparse \ac{PE}}, the query is matched against the top-100 retrieved images, and a spatial verification~\cite{philbin2007object} step is used to reject outliers matches. The query camera pose is then estimated using the database image with the largest number of verified matches. 

Tab.~\ref{tab:inloc-results} compares our approach with baselines from~\cite{Taira2018InLoc}: 
The original \textbf{Direct / Sparse \ac{PE}} pipelines are based on affine covariant features with RootSIFT descriptors \cite{Mikolajczyk2004Scale,Lowe2004Distinctive,Arandjelovic2012Three}. 
\textbf{Dense \ac{PE}} matches densely extracted CNN descriptors between the images (using guided matching from the \texttt{conv5} to the \texttt{conv3} layer in a VGG16 network). 
As in~\cite{Taira2018InLoc}, we report the percentage of query images localized within varying thresholds on their position error, considering only images with an orientation error of $10^\circ$ or less. 
We also report the average memory usage of features per image.
As can be seen, our approach outperforms both baselines. 

In addition to \textbf{Dense \ac{PE}}, the \textbf{InLoc} method proposed in~\cite{Taira2018InLoc} also verifies its estimated poses using dense information (\textbf{Dense \ac{PV}}): A synthetic image is rendered from the estimated pose and then compared to the query image using densely extracted SIFT descriptors. 
A similarity score is computed based on this comparison and used to re-rank the top-10 images after \textbf{Dense \ac{PE}}. 
Only this baseline outperforms our sparse feature approach, albeit at a higher computational cost. 
Combining our approach with \textbf{Dense \ac{PV}} also improves performance, but not to the level of \textbf{InLoc}. 
This is not surprising, given that \textbf{InLoc} is able to leverage dense data. 
Still, our results show that sparse methods can perform close to this strong baseline. 

Finally, by combining our method and \textbf{InLoc}, we were able to achieve a new state of the art --- we employed a pose selection algorithm using the \textbf{Dense \ac{PV}} scores for the top $10$ images of each method. In the end, $182$ \textbf{Dense \ac{PE}} poses and $174$ \textbf{Sparse \ac{PE}} (using D2 MS) poses were selected.

\section{Conclusions\label{sec:conclusion}}
We have proposed a novel approach to local feature extraction using a \emph{describe-and-detect} methodology. The detection is not done on low-level image structures but postponed until more reliable information is available, and done jointly with the image description. 
We have shown that our method surpasses the state-of-the-art in camera localization under challenging conditions such as day-night changes and indoor scenes. Moreover, even though our features are less well-localized compared to classical feature detectors, they are also suitable for 3D reconstruction.

An obvious direction for future work is to increase the accuracy at which our keypoints are detected. 
This could for example be done by increasing the spatial resolution of the CNN feature maps or by regressing more accurate pixel positions. 
Integrating a ratio test-like objective into our loss could help to improve the performance of our approach in applications such as SfM.

{\small
\PAR{Acknowledgements}
This work was partially supported by JSPS KAKENHI Grant Numbers 15H05313, 16KK0002, EU-H2020 project LADIO No. 731970, ERC grant LEAP No. 336845, CIFAR Learning in Machines \& Brains program. The ERDF supported J. Sivic under project IMPACT (CZ.02.1.01/0.0/0.0/15 003/0000468) and T. Pajdla under project Robotics for Industry 4.0 (CZ.02.1.01/0.0/0.0/15 003/0000470).
}

\appendix

\section*{Supplementary material}

This supplementary material provides the following additional information:
Section~\ref{sec:ratio} details how we chose the threshold for Lowe's ratio test~\cite{Lowe2004Distinctive} used for the 3D reconstructions in Section 5.2 in the paper.
As mentioned in Section 4.3 in the paper, Section~\ref{sec:architecture} provides implementation details on the architecture. In addition, the section also evaluates another backbone architecture (ResNet~\cite{He2015Deep}). 
Section~\ref{sec:loss} provides additional details on the loss function used to train our method. 
Section~\ref{sec:qualitative} shows qualitative examples for the matches found with our approach on the InLoc~\cite{Taira2018InLoc} and Aachen Day-Night~\cite{Sattler2017Benchmarking,Sattler2012Image} datasets.

\section{Impact of the ratio test on D2 features}
\label{sec:ratio}
Throughout our experiments on the local feature evaluation benchmark~\cite{Schonberger2017Comparative}, we noticed that Lowe's ratio test~\cite{Lowe2004Distinctive} plays an important role because it significantly reduces the number of wrong registrations due to repetitive structures and semantically similar scenes.

In order to find an adequate ratio threshold for our features, we employ Lowe's methodology~\cite{Lowe2004Distinctive}: 
we compute the probability density functions (PDFs) of correct and incorrect matches with respect to the ratio test threshold. However, contrary to Lowe's evaluation, we considered only mutual nearest neighbors during the matching process.

Our evaluation was done on the entire HPatches~\cite{HPATCHES} image pairs dataset consisting of $580$ pairs from $116$ sequences ($57$ with illumination changes and $59$ with viewpoint changes). 
A match is considered correct if its projection error, estimated using the homographies provided by the dataset, is below $4$ pixels - the default threshold in COLMAP~\cite{schoenberger2016sfm,schoenberger2016mvs} during geometric verification and bundle adjustment. To take into account the possible errors in annotations and to have a clear separation between correct and incorrect matches, the threshold for incorrect matches is set to $20$ pixels. Matches with projection errors between 4 and 20 pixels are therefore discarded during this evaluation.

\begin{figure}[t]
    \begin{subfigure}[t]{\columnwidth}
    \centering
    \includegraphics[width=\textwidth]{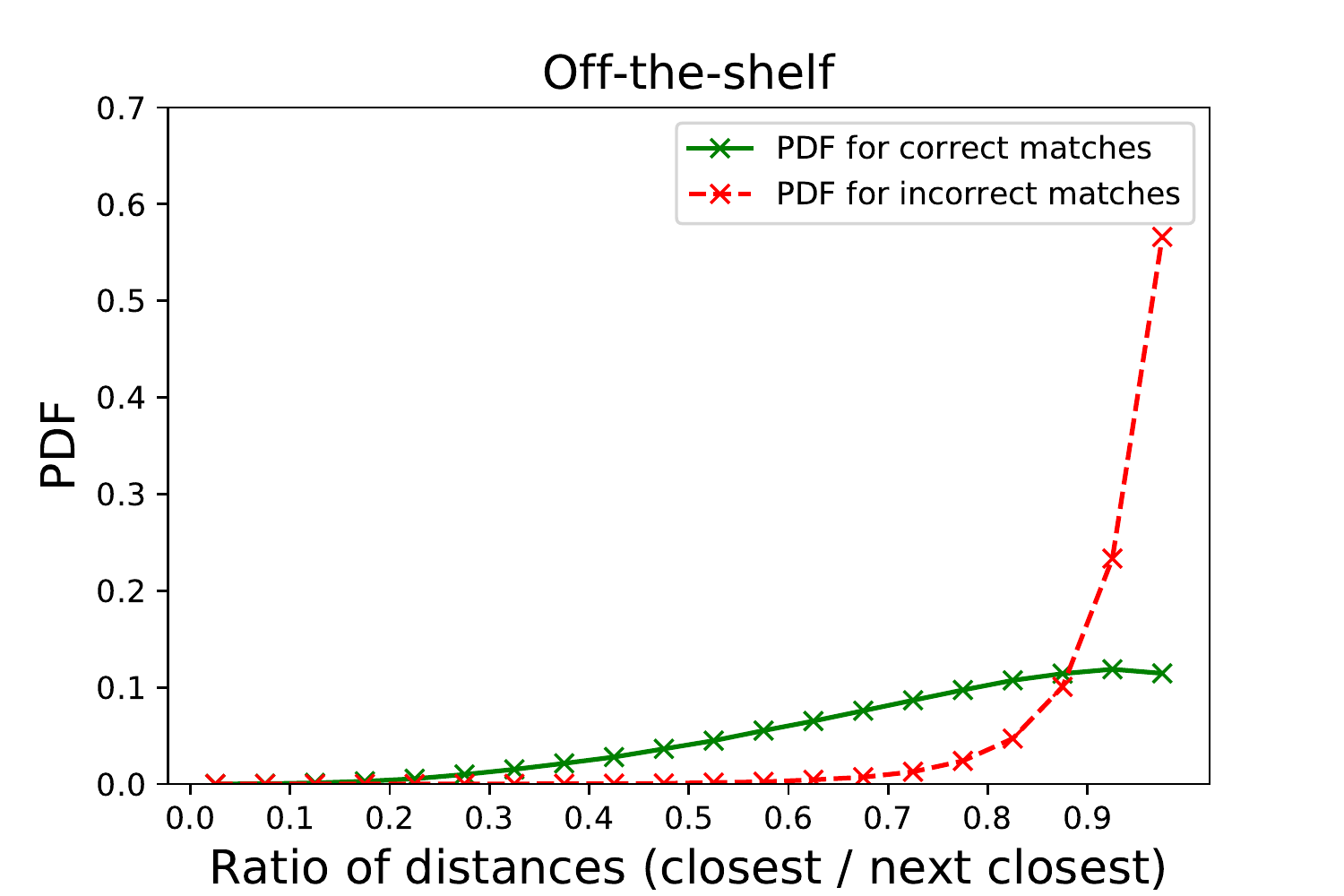}
    \end{subfigure}
    
    \begin{subfigure}[t]{\columnwidth}
    \centering
    \includegraphics[width=\textwidth]{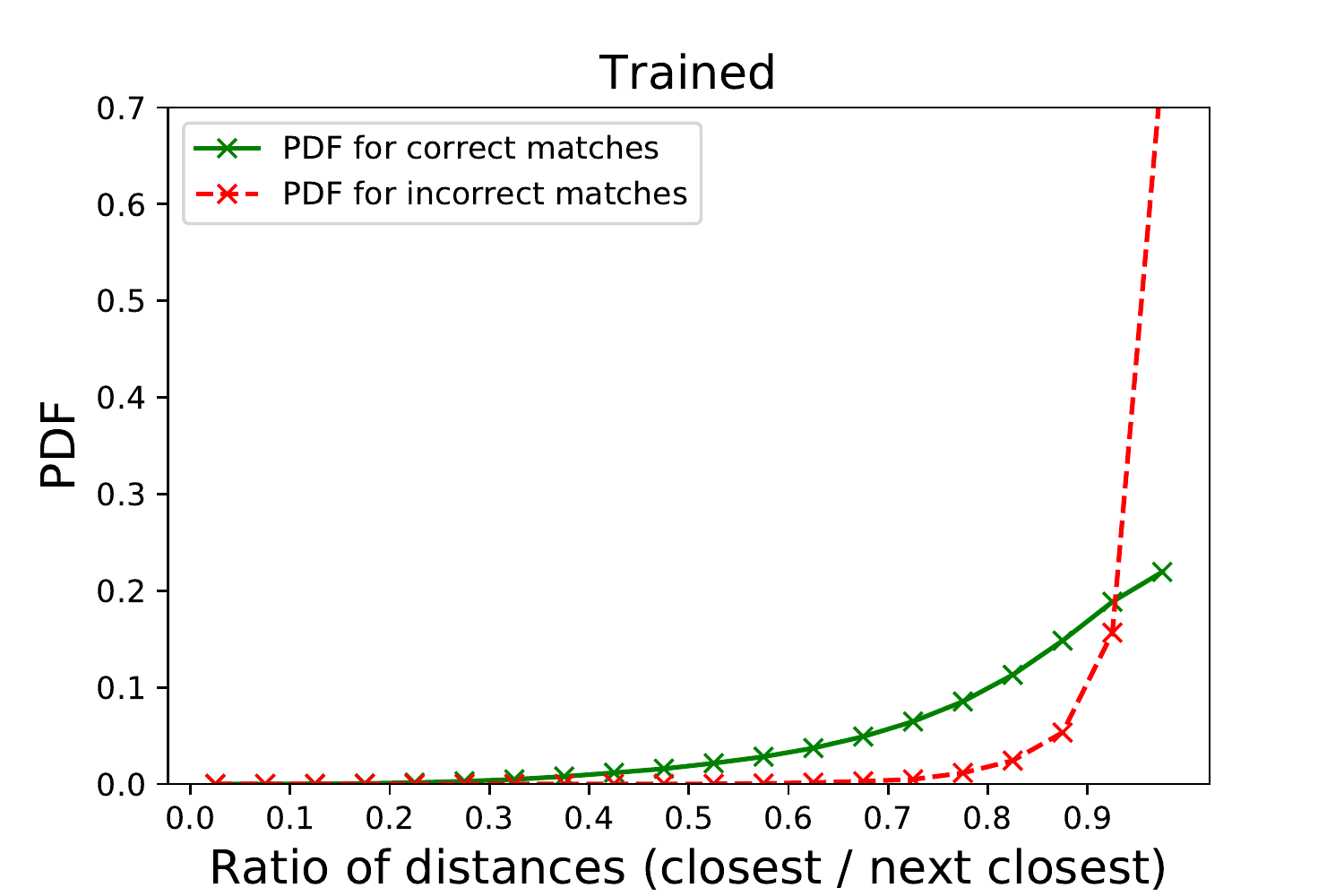}
    \end{subfigure}
    \caption{\small \textbf{Ratio PDFs for D2 multi-scale features.} PDF in terms of ratio on the full HPatches~\cite{HPATCHES} image pairs dataset for the D2 off-the-shelf and fine-tuned features. There is no clear separation between the mean ratios of correct and incorrect matches as in the case of SIFT~\cite{Lowe2004Distinctive}.}
    \label{fig:ratio-test}
\end{figure}

Figure~\ref{fig:ratio-test} shows the two PDFs. 
As can be seen, the D2 features do not work too well with ratio filtering because the mean ratio of correct matches is close to the one of incorrect matches. 
Still, we used thresholds of $0.90$ for the off-the-shelf descriptors and $0.95$ for the fine-tuned ones, which filter out $79.9\%$ and $74.4\%$ of incorrect matches, respectively. Unfortunately, these thresholds also discard a significant amount of correct matches ($23.3\%$ and $21.9\%$, respectively) which can have a negative impact on the number of registered images and sparse points.

In practice, we suggest not using the ratio test for camera localization under difficult conditions (e.g. day-night, indoors). For 3D reconstruction, using the threshold suggested above and / or increasing the minimum number of inlier matches required for an image pair to be considered during \ac{SfM} should be sufficient to avoid most wrong registrations. Please note that, in the second case, the geometric verification can be significantly slower as RANSAC needs to handle a larger outlier ratio.

\section{Details of the backbone architecture}
\label{sec:architecture}

For the feature extraction network $\mathcal{F}$, we used a VGG16 network pretrained on the ImageNet dataset~\cite{Deng2009ImageNet}, truncated after the \texttt{conv4\_3} layer, as detailed in Section 4.3 of the paper. In addition, as also detailed in Section 4.3, we use a different image and feature resolution during training compared to testing. In particular, during testing, we take advantage of dilated convolutions~\cite{Yu2016ICLR,holschneider1990real} to increase the resolution of the feature maps - this is not done in training due to memory limitations. More detailed descriptions of the network architectures during the training and testing phases are provided in Tables~\ref{tab:training} and~\ref{tab:testing}, respectively.

We additionally assess the choice of the network used for feature extraction, by performing a comparison between the chosen VGG16~\cite{Simonyan2014Very} architecture and ResNet50~\cite{He2015Deep} (which is the state of the art backbone architecture used in various other works). We evaluate them on the HPatches image pairs dataset using the same evaluation protocol that is described in Section 5.1 of the main paper. 

For both architectures, we used weights trained on ImageNet~\cite{Deng2009ImageNet}. In the case of ResNet50, the network was truncated after \texttt{conv4\_6} (following the approach in DELF~\cite{Noh2017Largescale}). At this point in the architecture, the resolution is $\nicefrac{1}{16}^{\text{th}}$ of the input resolution and the descriptors are $1024$-dimensional. However, in the case of the original VGG16, the output after \texttt{conv4\_3} has $\nicefrac{1}{8}^{\text{th}}$ of the input resolution and $512$ channels. In order to account for this difference in resolution, we use dilated convolutions (also sometimes referred to as ``atrous convolutions") to increase the resolution for the ResNet50 network. In addition, dilated convolutions are applied to both networks to further increase the feature resolution to $\nicefrac{1}{4}^{\text{th}}$ of the input resolution. For simplicity, only single-scale features are considered in this comparison.

\begin{figure*}[t]
    \begin{subfigure}[t]{0.60\textwidth}
    \centering
    \raisebox{-0.5\height}{\includegraphics[width=0.9\textwidth]{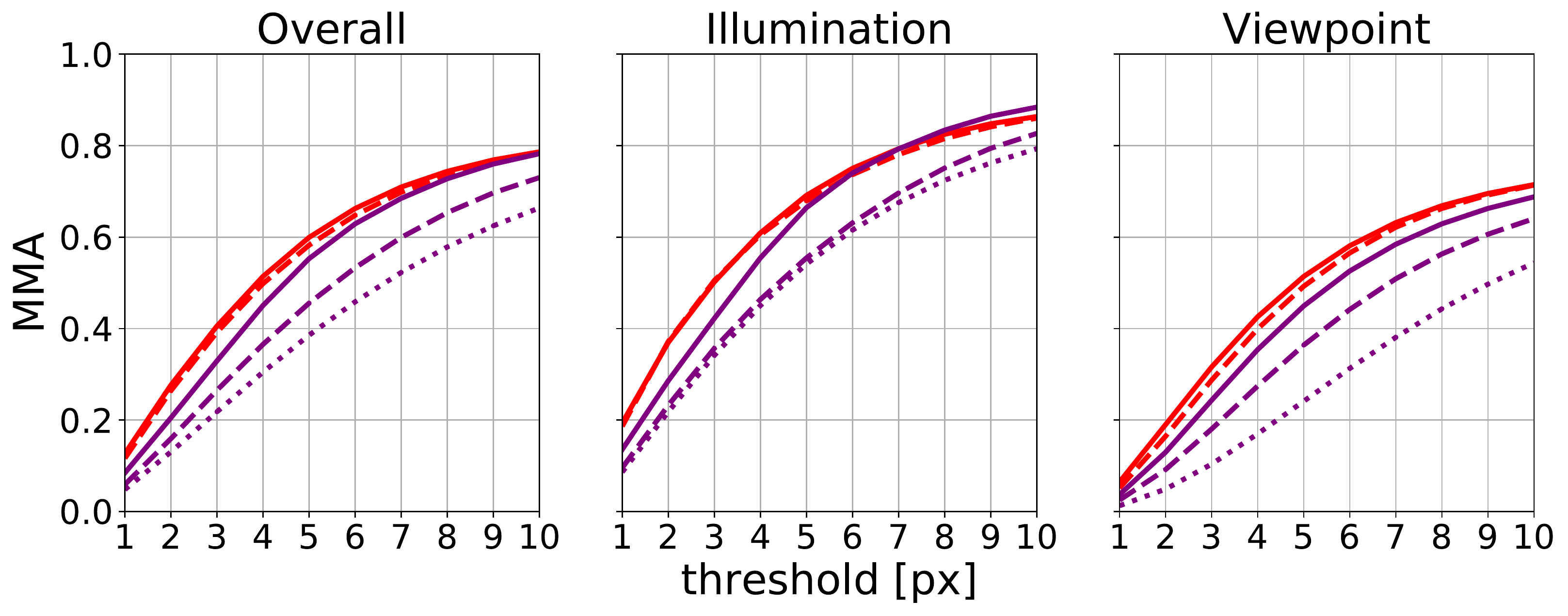}}
    \end{subfigure}
    ~
    \begin{subfigure}[t]{0.35\textwidth}
    \centering
    \scriptsize {
    \begin{tabular}{@{}c@{\hspace{0.5mm}}l@{\hspace{1mm}}c@{\hspace{1mm}}c@{\hspace{1mm}}c@{}}
    \toprule
    \multicolumn{2}{@{}l}{\textbf{Method}} & \textbf{Feature map res.} & \textbf{\# Features} & \textbf{\# Matches} \\ 
    \midrule
    \tikz[baseline]{\draw[color=Red,line width=0.7mm,densely dashed] (0,.5ex)--++(.45,0) ;} & \textbf{VGG16} & $\times \nicefrac{1}{8}$ & $2.7$K & $1.1$K \\
    \tikz[baseline]{\draw[color=Red,line width=0.7mm] (0,.5ex)--++(.45,0) ;} & VGG16 & $\times \nicefrac{1}{4}$ & $3.0$K & $1.2$K \\
    \tikz[baseline]{\draw[color=Purple,line width=0.7mm,dotted] (0,.5ex)--++(.45,0) ;} & \textbf{ResNet50} & $\times \nicefrac{1}{16}$ & $1.5$K & $0.6$K \\
    \tikz[baseline]{\draw[color=Purple,line width=0.7mm,dashed] (0,.5ex)--++(.45,0) ;} & ResNet50 & $\times \nicefrac{1}{8}$ & $3.1$K & $1.1$K \\
    \tikz[baseline]{\draw[color=Purple,line width=0.7mm] (0,.5ex)--++(.45,0) ;} & ResNet50 & $\times \nicefrac{1}{4}$ & $8.5$K & $2.5$K \\
    \bottomrule 
    \end{tabular}
    }
    \end{subfigure}
    \caption{\small \textbf{Evaluation of different backbone architectures on the HPatches image pairs.} The original networks are in \textbf{bold} - the others were obtained by removing the stride of the deepest layers and adding dilations to the subsequent ones. Dilated convolutions offer more keypoints and better performance in viewpoint sequences. VGG16 outperforms ResNet50 by a significant margin even at a similar feature map resolution.
    }
    \label{fig:hsequences-architecture}
\end{figure*}
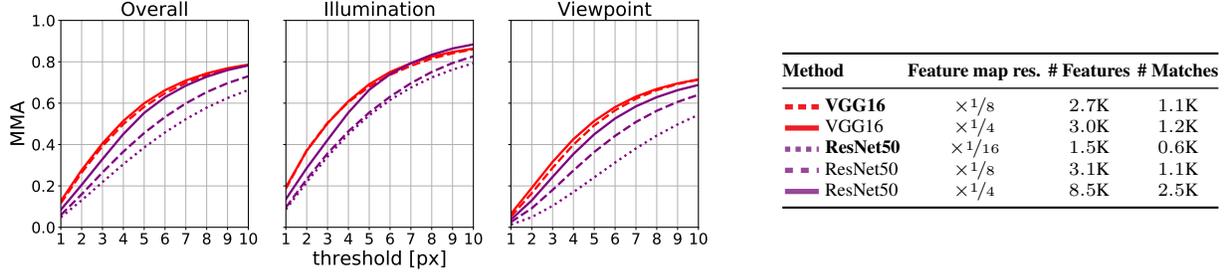

The results can be seen in Figure~\ref{fig:hsequences-architecture}. Dilated convolutions~\cite{Yu2016ICLR,holschneider1990real} increase the number of detections and the performance of D2 features especially in the case of viewpoint changes. The ResNet50 features also benefit from dilated convolutions and the increase in the resolution. 
However, although ResNet50 features seem slightly more robust to illumination changes and are able to outperform VGG16 features for thresholds larger than $6.5$ pixels, they are less robust to viewpoint changes. Overall, ResNet50 features obtain worse results in this evaluation which motivated our decision to use VGG16.

\begin{table}[t]
    \centering
    \scriptsize{
    \begin{tabular}{c@{\hspace{1mm}}c@{\hspace{1mm}}c@{\hspace{1mm}}c@{\hspace{1mm}}c}
        \toprule
        Layer & Stride & Dilation & ReLU & Resolution \\ \midrule
        \texttt{input} ($256 \times 256$) - $3$ ch. & & & & $\times 1$ \\ \midrule
        \texttt{conv1\_1} - $3 \times 3$, $64$ ch. & $1$ & $1$ & \checkmark & $\times 1$\\
        \texttt{conv1\_2} - $3 \times 3$, $64$ ch. & $1$ & $1$ & \checkmark & $\times 1$\\
        \texttt{pool1} - $2 \times 2$, max. & $2$ & $1$ & & $\times \nicefrac{1}{2}$ \\ \midrule
        \texttt{conv2\_1} - $3 \times 3$, $128$ ch. & $1$ & $1$ & \checkmark & $\times \nicefrac{1}{2}$ \\
        \texttt{conv2\_2} - $3 \times 3$, $128$ ch. & $1$ & $1$ & \checkmark & $\times \nicefrac{1}{2}$ \\
        \texttt{pool2} - $2 \times 2$, max. & $2$ & $1$ & & $\times \nicefrac{1}{4}$ \\ \midrule
        \texttt{conv3\_1} - $3 \times 3$, $256$ ch. & $1$ & $1$ & \checkmark & $\times \nicefrac{1}{4}$ \\
        \texttt{conv3\_2} - $3 \times 3$, $256$ ch. & $1$ & $1$ & \checkmark & $\times \nicefrac{1}{4}$ \\
        \texttt{conv3\_3} - $3 \times 3$, $256$ ch.& $1$ & $1$ & \checkmark & $\times \nicefrac{1}{4}$ \\
        \texttt{pool3} - $2 \times 2$, max. & $2$ & $1$ & & $\times \nicefrac{1}{8}$ \\ \midrule
        \texttt{conv4\_1} - $3 \times 3$, $512$ ch. & $1$ & $1$ & \checkmark & $\times \nicefrac{1}{8}$ \\
        \texttt{conv4\_2} - $3 \times 3$, $512$ ch. & $1$ & $1$ & \checkmark & $\times \nicefrac{1}{8}$ \\
        \textbf{\texttt{conv4\_3}} - $3 \times 3$, $512$ ch. & $1$ & $1$ & & $\times \nicefrac{1}{8}$\\
        \bottomrule
    \end{tabular}
    }
    \caption{\small {\bf Training architecture.} During training, we use the default VGG16~\cite{Simonyan2014Very} architecture up to \texttt{conv4\_3}, and fine-tune the last layer (\texttt{conv4\_3}).}
    \label{tab:training}
\end{table}

\begin{table}[t]
    \centering
    \scriptsize{
    \begin{tabular}{c@{\hspace{1mm}}c@{\hspace{1mm}}c@{\hspace{1mm}}c@{\hspace{1mm}}c}
        \toprule
        Layer & Stride & Dilation & ReLU & Resolution \\ \midrule
        \texttt{input} ($\sim 1200 \times 1600$) - $3$ ch. & & & & $\times 1$ \\ \midrule
        \texttt{conv1\_1} - $3 \times 3$, $64$ ch. & $1$ & $1$ & \checkmark & $\times 1$ \\
        \texttt{conv1\_2} - $3 \times 3$, $64$ ch. & $1$ & $1$ & \checkmark & $\times 1$ \\
        \texttt{pool1} - $2 \times 2$, max. & $2$ & $1$ & & $\times \nicefrac{1}{2}$ \\ \midrule
        \texttt{conv2\_1} - $3 \times 3$, $128$ ch. & $1$ & $1$ & \checkmark & $\times \nicefrac{1}{2}$ \\
        \texttt{conv2\_2} - $3 \times 3$, $128$ ch. & $1$ & $1$ & \checkmark & $\times \nicefrac{1}{2}$ \\
        \texttt{pool2} - $2 \times 2$, max. & $2$ & $1$ & & $\times \nicefrac{1}{4}$ \\ \midrule
        \texttt{conv3\_1} - $3 \times 3$, $256$ ch. & $1$ & $1$ & \checkmark & $\times \nicefrac{1}{4}$ \\
        \texttt{conv3\_2} - $3 \times 3$, $256$ ch. & $1$ & $1$ & \checkmark & $\times \nicefrac{1}{4}$ \\
        \texttt{conv3\_3} - $3 \times 3$, $256$ ch.& $1$ & $1$ & \checkmark & $\times \nicefrac{1}{4}$ \\
        \texttt{pool3} - $2 \times 2$, avg. & $1$ & $1$ & & $\times \nicefrac{1}{4}$ \\ \midrule
        \texttt{conv4\_1} - $3 \times 3$, $512$ ch. & $1$ & $2$ & \checkmark & $\times \nicefrac{1}{4}$ \\
        \texttt{conv4\_2} - $3 \times 3$, $512$ ch. & $1$ & $2$ & \checkmark & $\times \nicefrac{1}{4}$ \\
        \texttt{conv4\_3} - $3 \times 3$, $512$ ch. & $1$ & $2$ & \footnotemark[1] & $\times \nicefrac{1}{4}$ \\
        \bottomrule
    \end{tabular}
    }
    \caption{\small {\bf Testing architecture.} At test time, we slightly modify the training architecture: the last pooling layer \texttt{pool3} is replaced by an average pooling with a stride of $1$ and the following convolutional layers are dilated by a factor of $2$. This maintains the same receptive field but offers higher resolution feature maps.}
    \label{tab:testing}
\end{table}

\footnotetext[1]{We noticed that ReLU has a significant negative impact on the off-the-shelf descriptors, but not on the fine-tuned ones. Thus, we report results without ReLU for the off-the-shelf model and with ReLU for the fine-tuned one.}

\section{Details of the training loss}
\label{sec:loss}

\begin{figure}
    \centering
    \includegraphics[width=0.9\columnwidth]{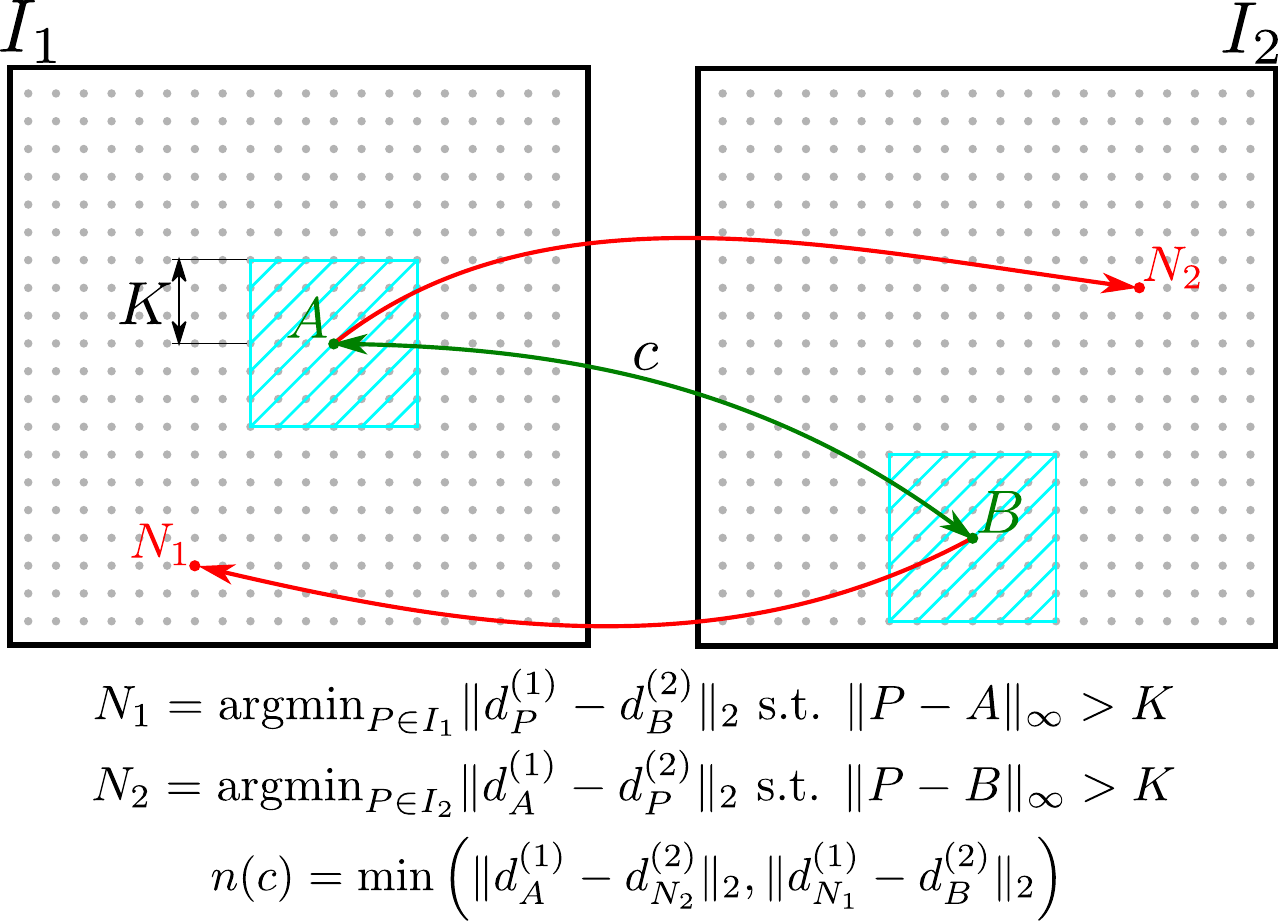}
    \caption{\small \textbf{In-image-pair negative mining procedure.} For each correspondence $c: A \leftrightarrow B$, the negative sample is chosen between the hardest negative of $A$ in $I_2$ ($N_2$) or of $B$ in $I_1$ ($N_1$). Since adjacent pixels at feature map level have overlapping receptive fields in the input image, the negative descriptor is chosen to be at least $K$ pixels away from the ground-truth correspondence.}
    \label{fig:loss}
\end{figure}

This section gives more insight into the loss $\mathcal{L}$ that we used for fine-tuning the \texttt{conv4\_3} layer of the VGG16 network. 
In particular, in Figure~\ref{fig:loss} we explain in more detail the in-image-pair negative mining expressed in Equations (10) and (11) of the paper.

The parameter $K$ controls the size of the neighbourhood from where negative samples are \emph{not} selected. For a value of $K = 0$, all feature map pixels apart from the considered correspondence $c: A \leftrightarrow B$ are considered as possible negatives. In this case, a value of the margin loss $m(c)$ lower than $M$ ($p(c) < n(c)$) signifies that $A$ and $B$ would be matched using mutual nearest neighbors. This is due to the symmetric negative selection. However, in practice, this is too restrictive since adjacent pixels have a significant overlap in their receptive field so the descriptors can be very close. Since the receptive field at the \texttt{conv4\_3} level is around $65\times65$ pixels at the input resolution, we choose a value of $K = 4$ at the feature map level, which enforces that potential negatives have less than $50\%$ spatial overlap.

Another parameter of the training loss is the margin $M$. Since the descriptors are L2 normalized, the squared distance between two descriptors is guaranteed to be lower than $4$. We have settled for $M = 1$ as in previous work~\cite{Mishchuk2017Working}. It is worth noting that, due to the the negative mining scheme, this margin is rarely reached, \ie, the detection scores continue to be optimized.

\begin{figure*}[p]
    \centering
    \begin{tabular}{c@{\hspace{1cm}}c@{\hspace{1cm}}c}
        Image & Off-the-shelf & Trained \\
        \includegraphics[height=0.15\textheight]{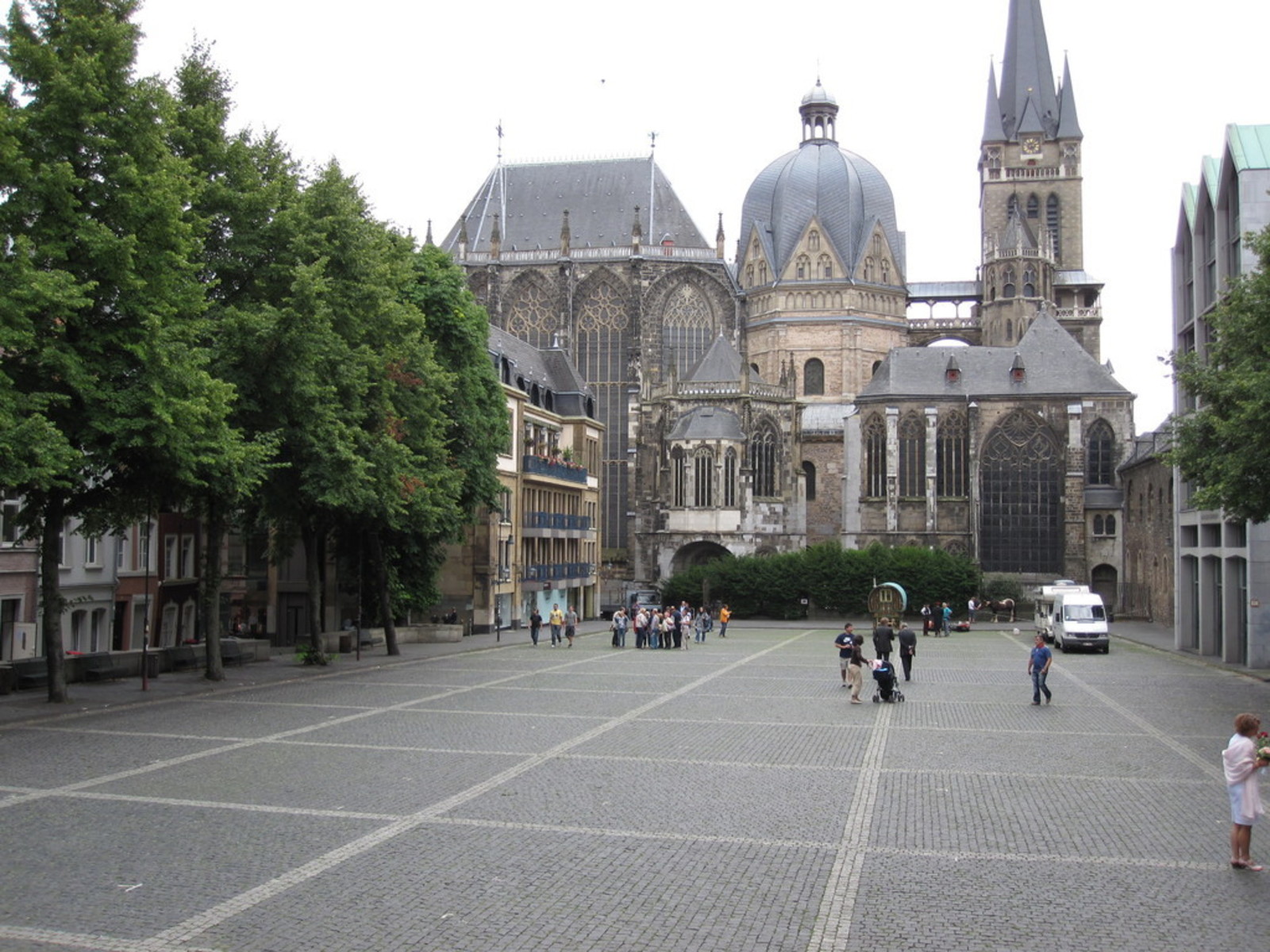} & \includegraphics[height=0.15\textheight]{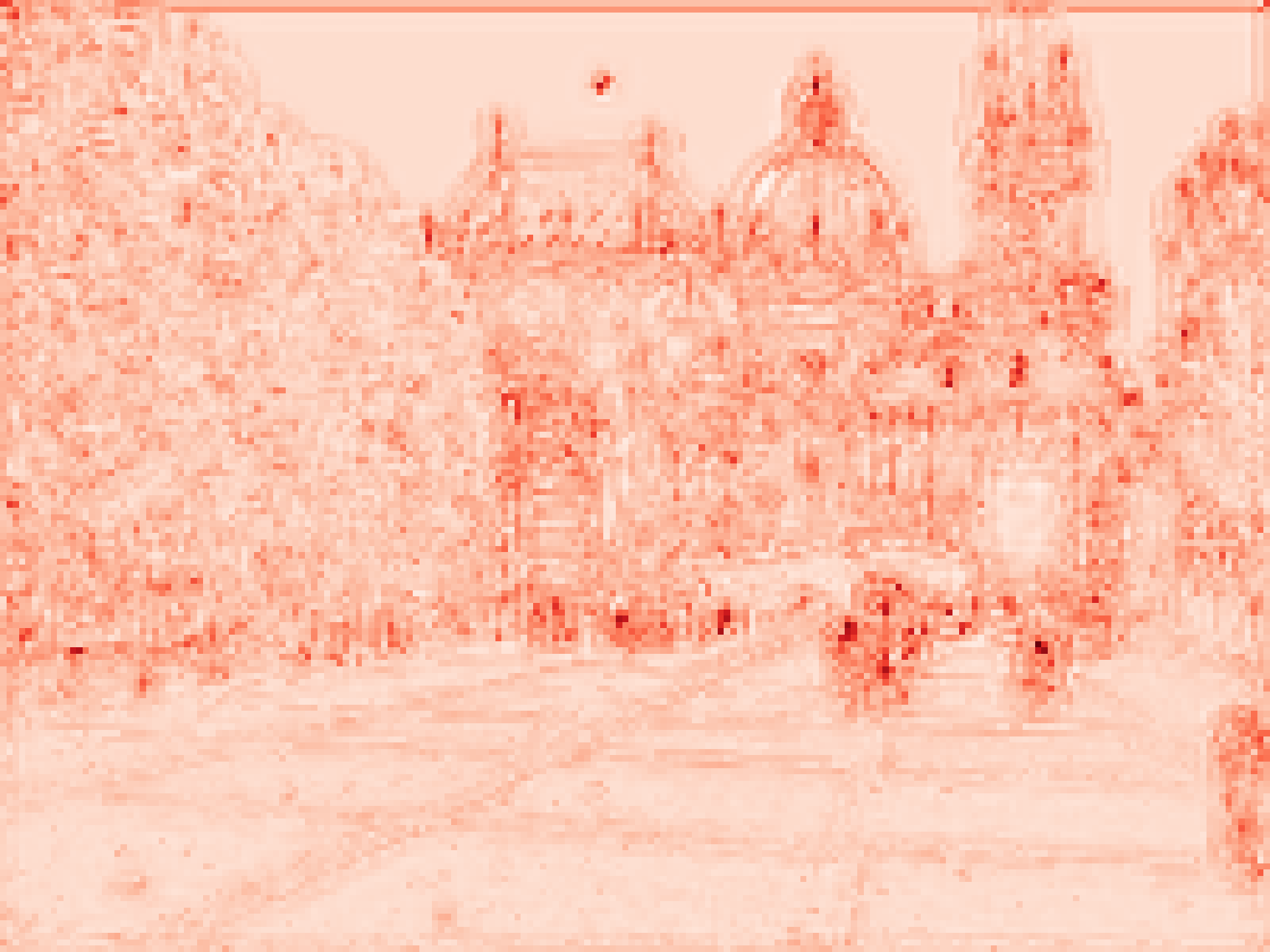} & \includegraphics[height=0.15\textheight]{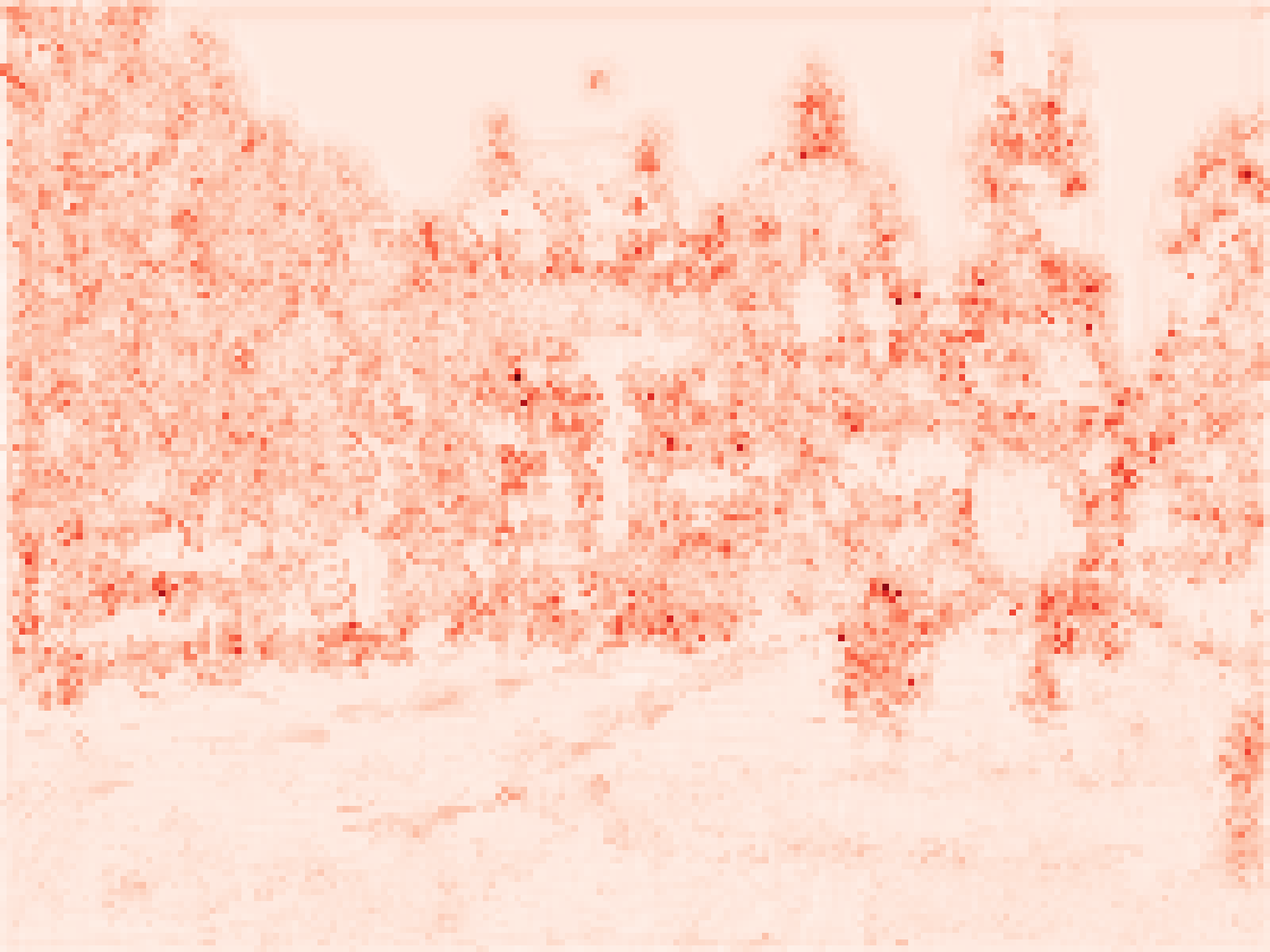} \\
        \includegraphics[height=0.15\textheight]{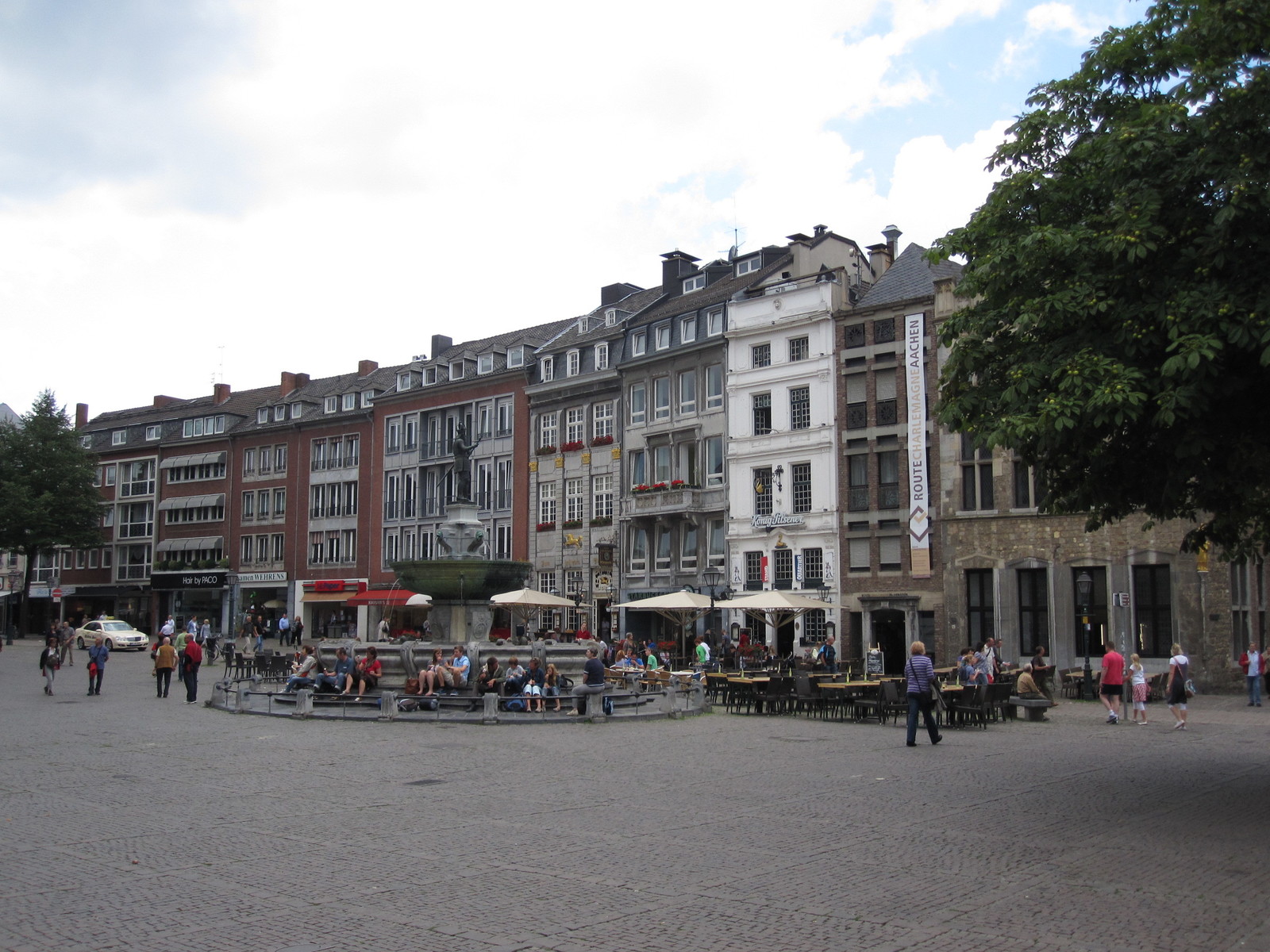} & \includegraphics[height=0.15\textheight]{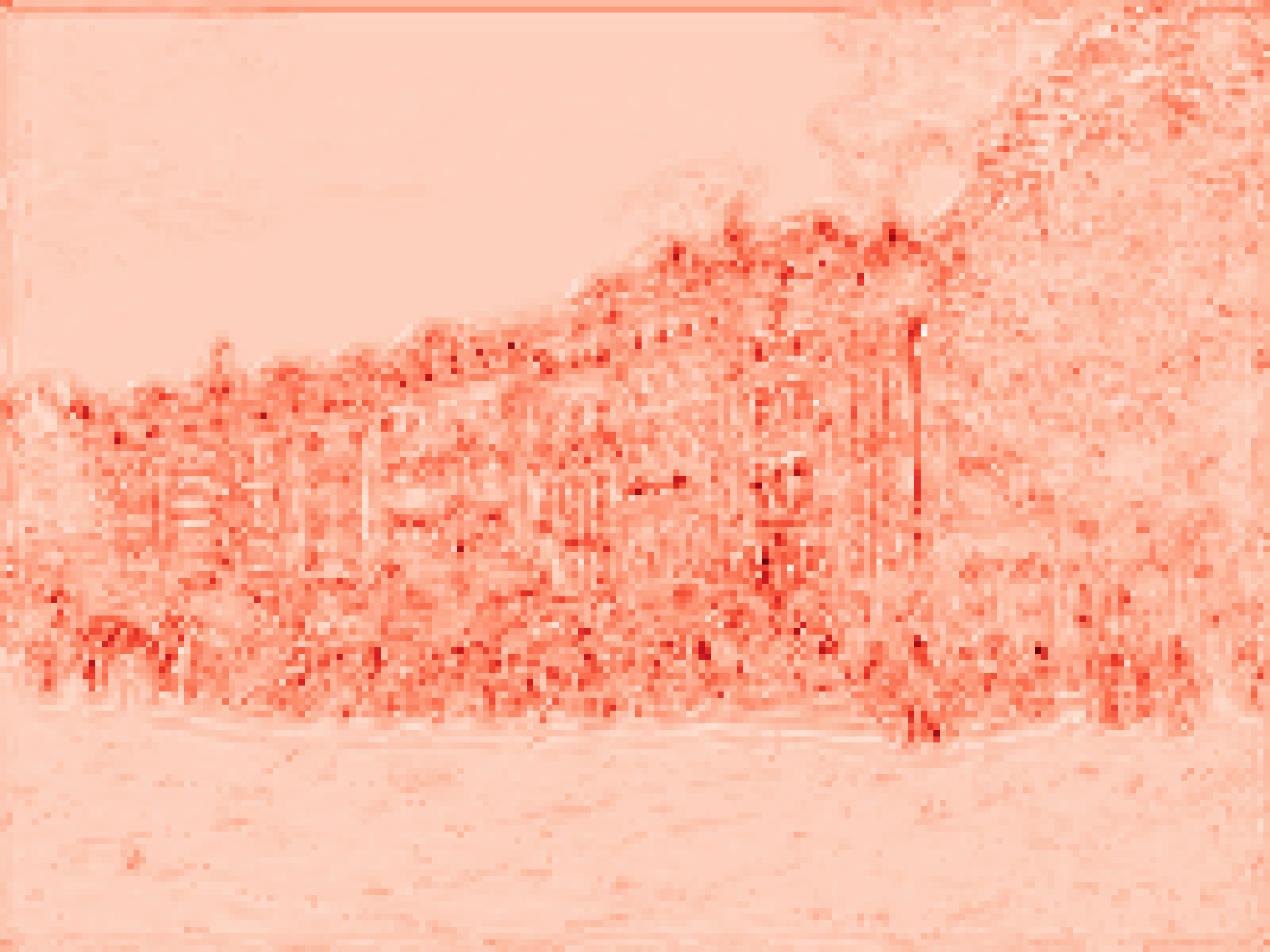} & \includegraphics[height=0.15\textheight]{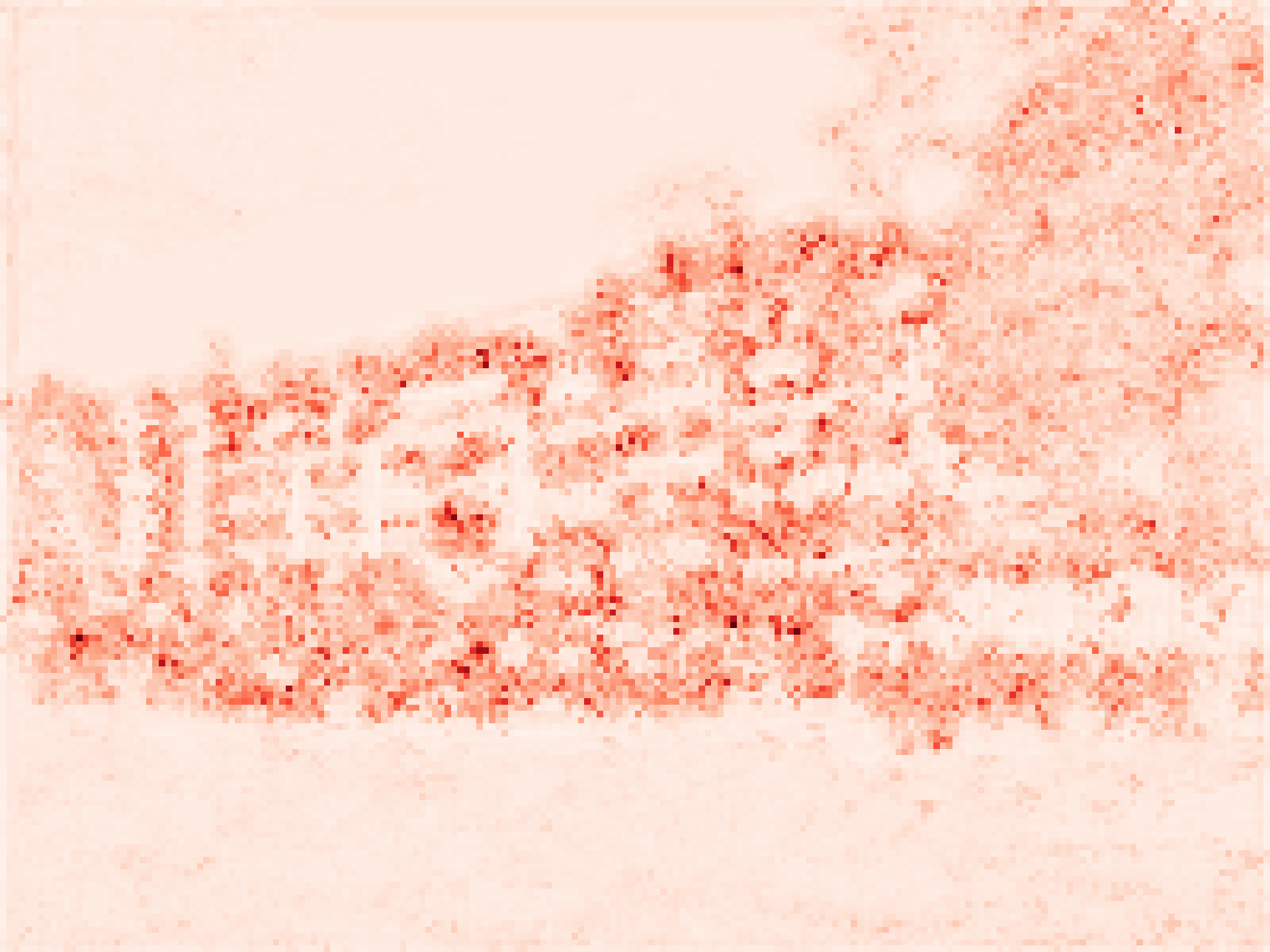} \\
        \includegraphics[height=0.20\textheight]{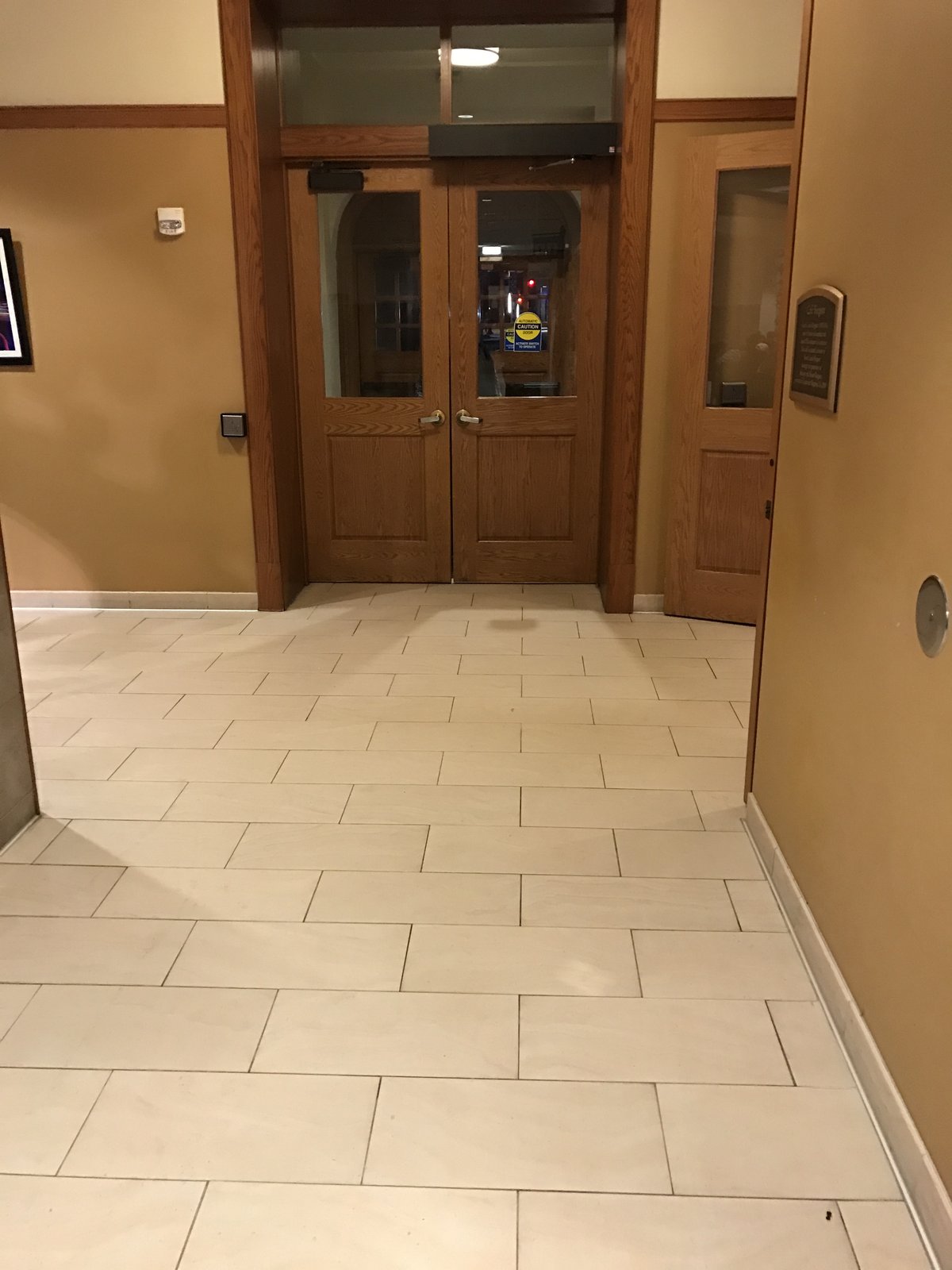} & \includegraphics[height=0.20\textheight]{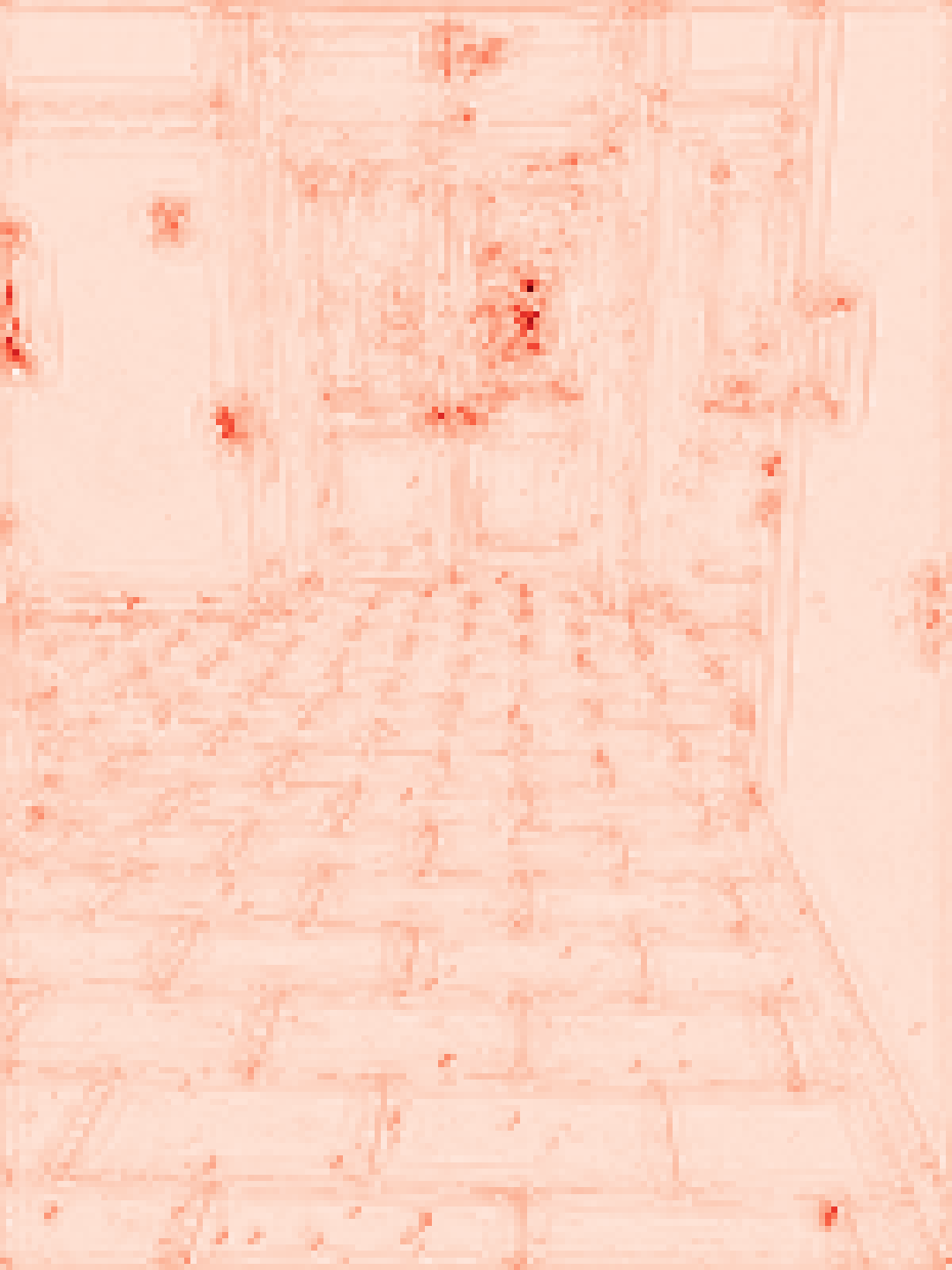} & \includegraphics[height=0.20\textheight]{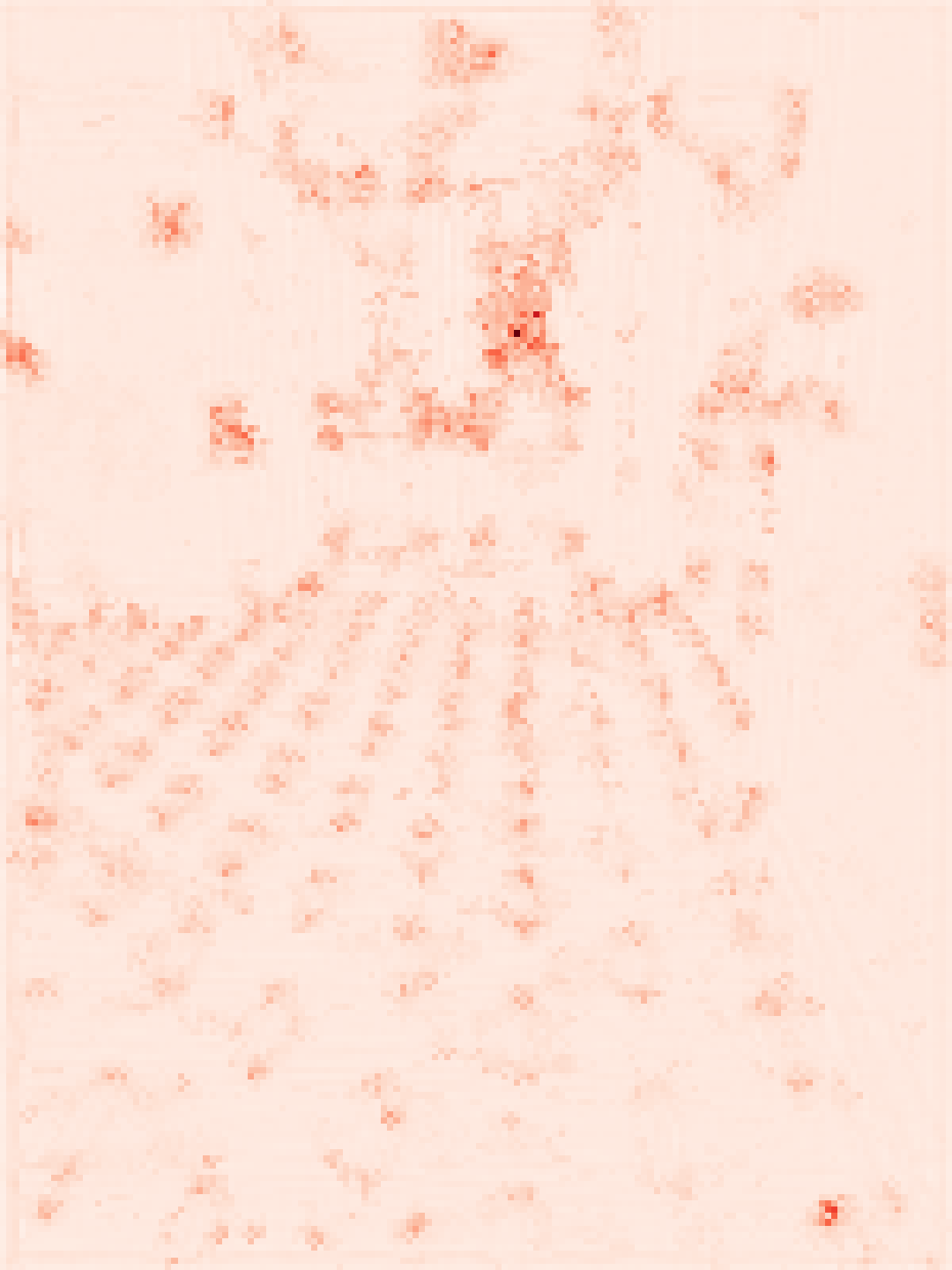} \\
        \includegraphics[height=0.20\textheight]{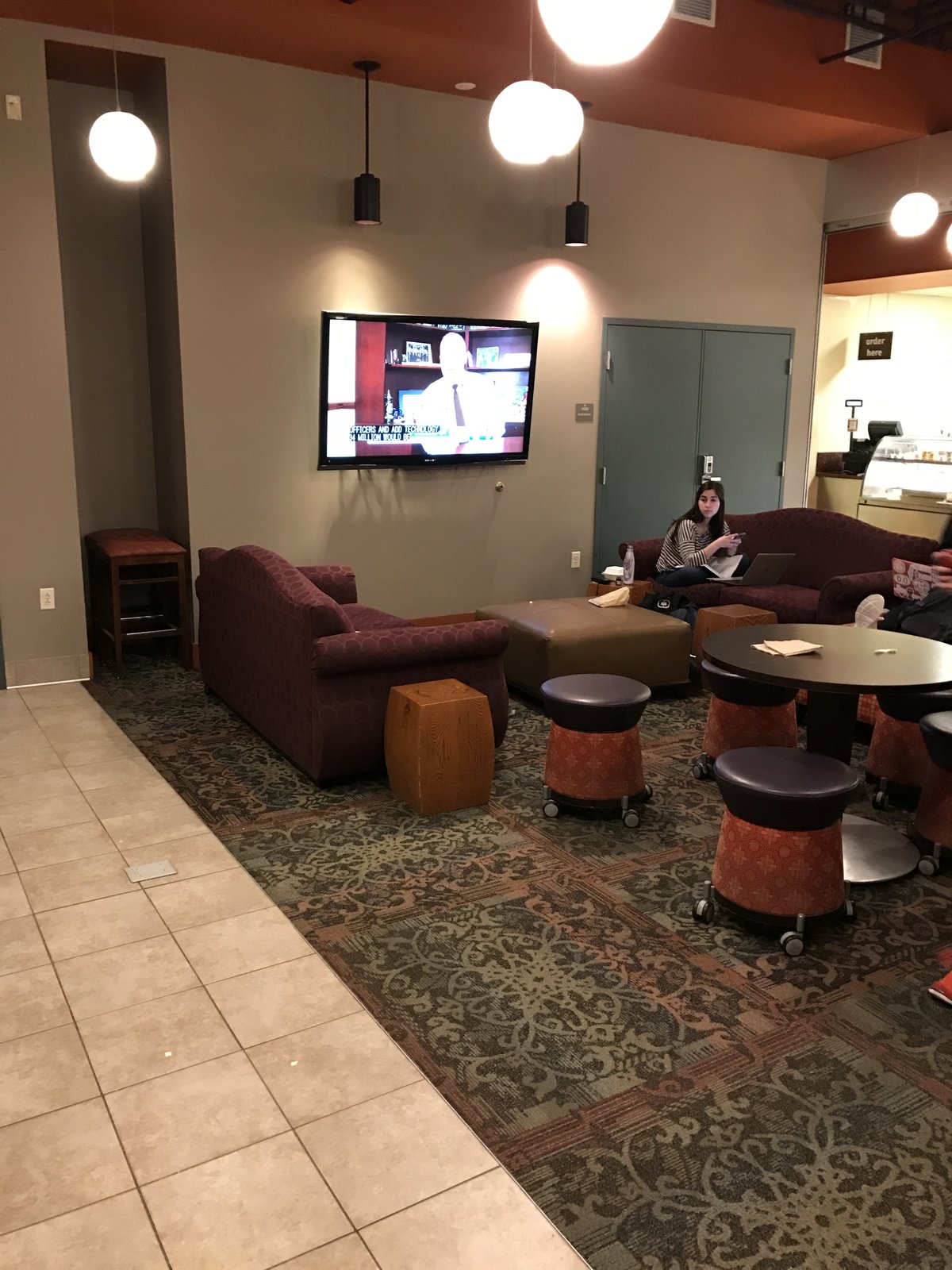} & \includegraphics[height=0.20\textheight]{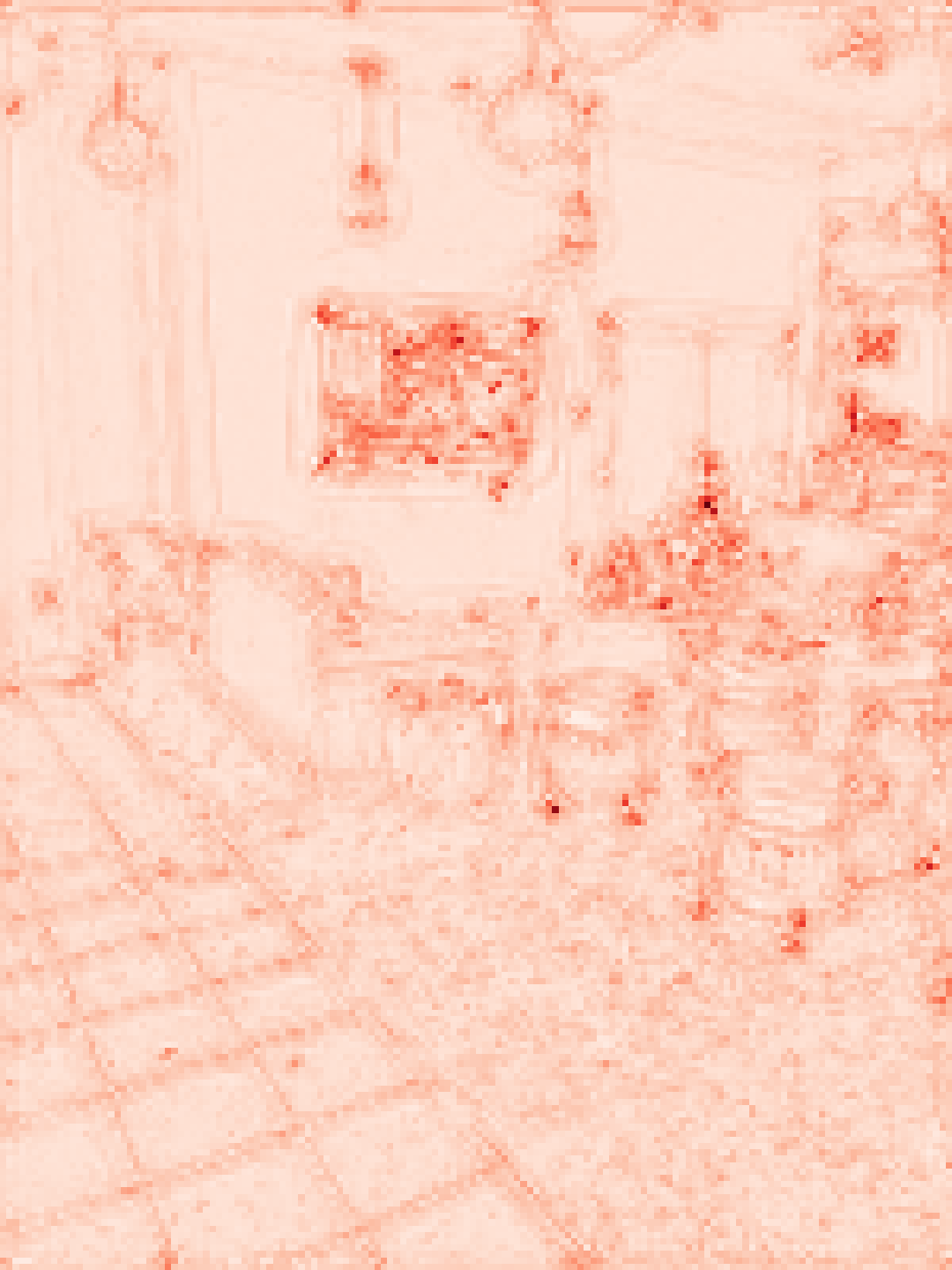} & \includegraphics[height=0.20\textheight]{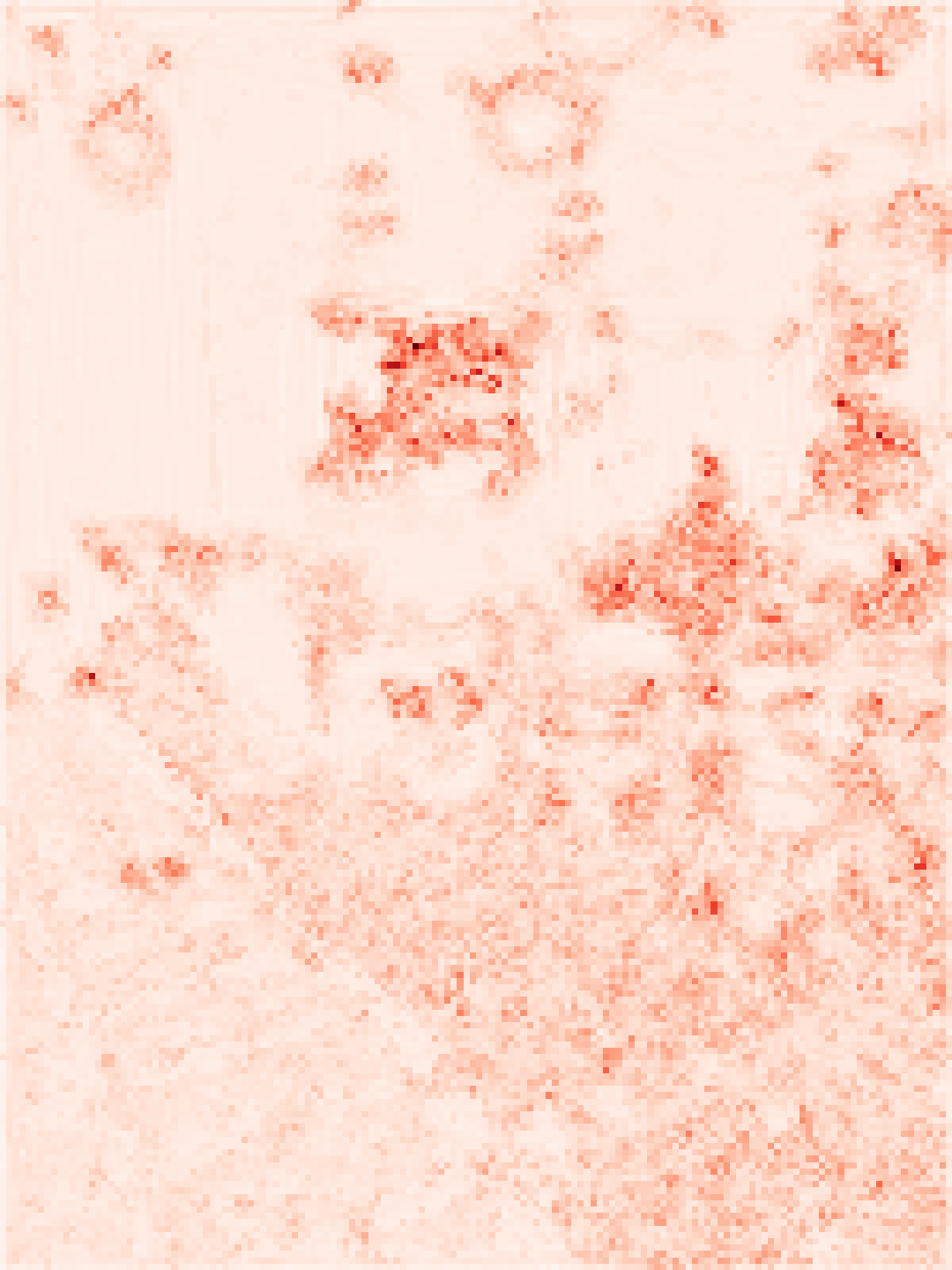} \\
    \end{tabular}
    \caption{\small {\bf Soft detection scores for different scenes before and after fine-tuning.} White represents low soft-detection scores while red signifies higher ones. The training lowers the soft-detection scores on repetitive structures (\eg ground, floor, walls) while it enhances the score on more distinctive points. This shown by the increased contrast of the trained soft-detection maps with respect to their off-the-shelf counterparts.}
    \label{tab:soft-det}
\end{figure*}

Figure~\ref{tab:soft-det} shows the soft detection scores before and after fine-tuning. As expected, some salient points have increased scores, while repetitive structures are weighted down. Even though most of our training data is from outdoors scenes, these observations seem to translate well to indoors images too.

\section{Qualitative examples}
\label{sec:qualitative}
\begin{figure}[p]
    \centering
    % viewpoint
    \includegraphics[width=\columnwidth]{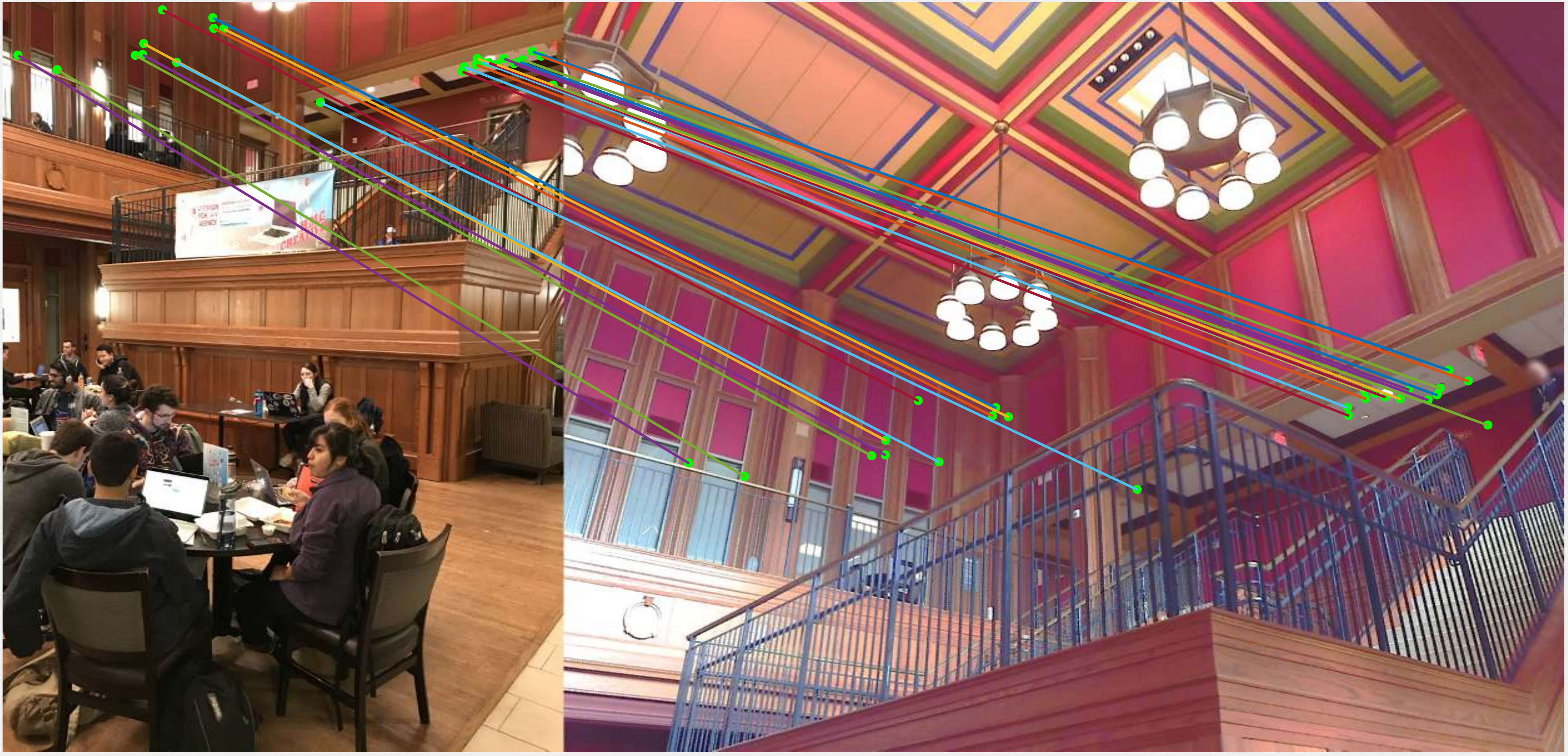}
    % object matching
    \includegraphics[width=\columnwidth]{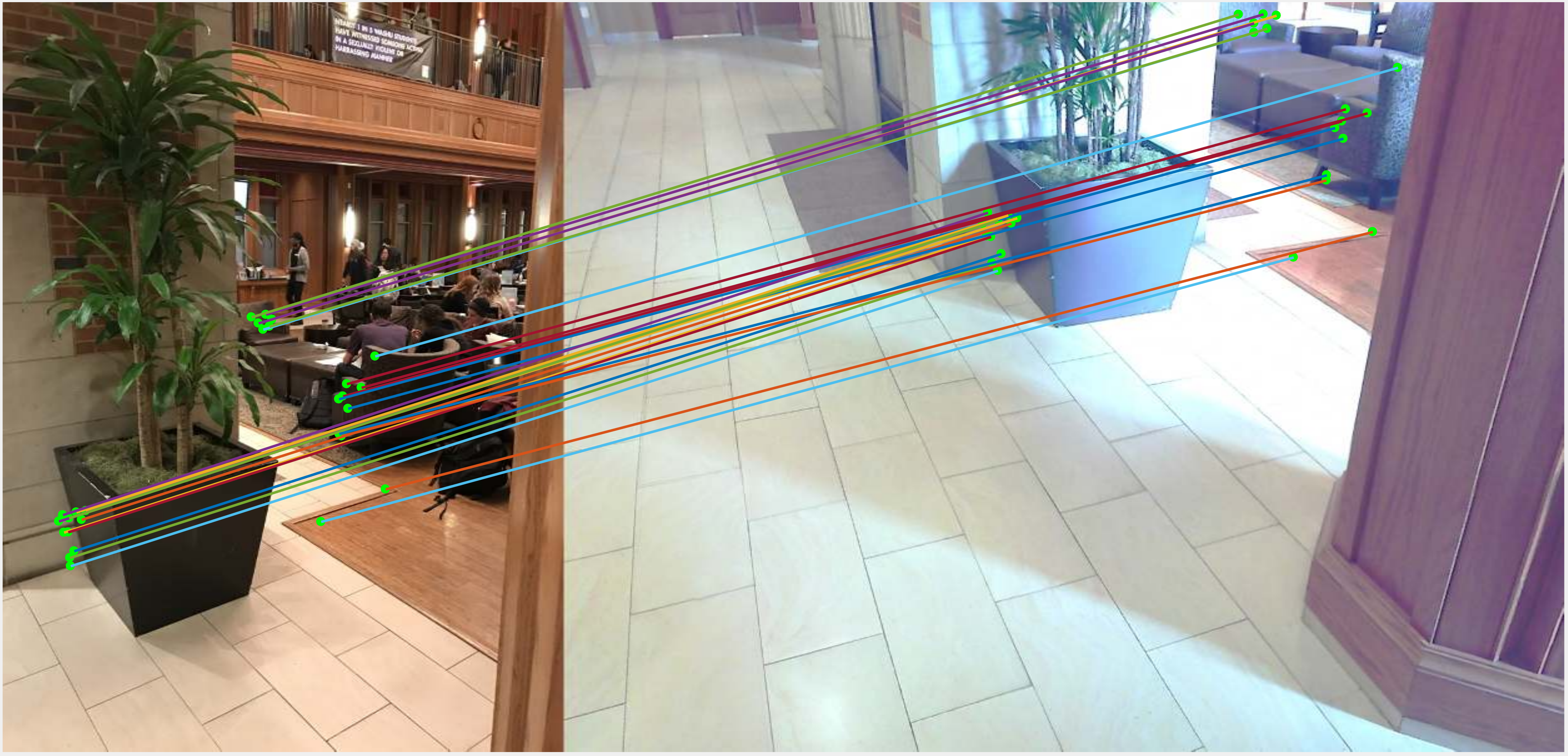}
    \includegraphics[width=\columnwidth]{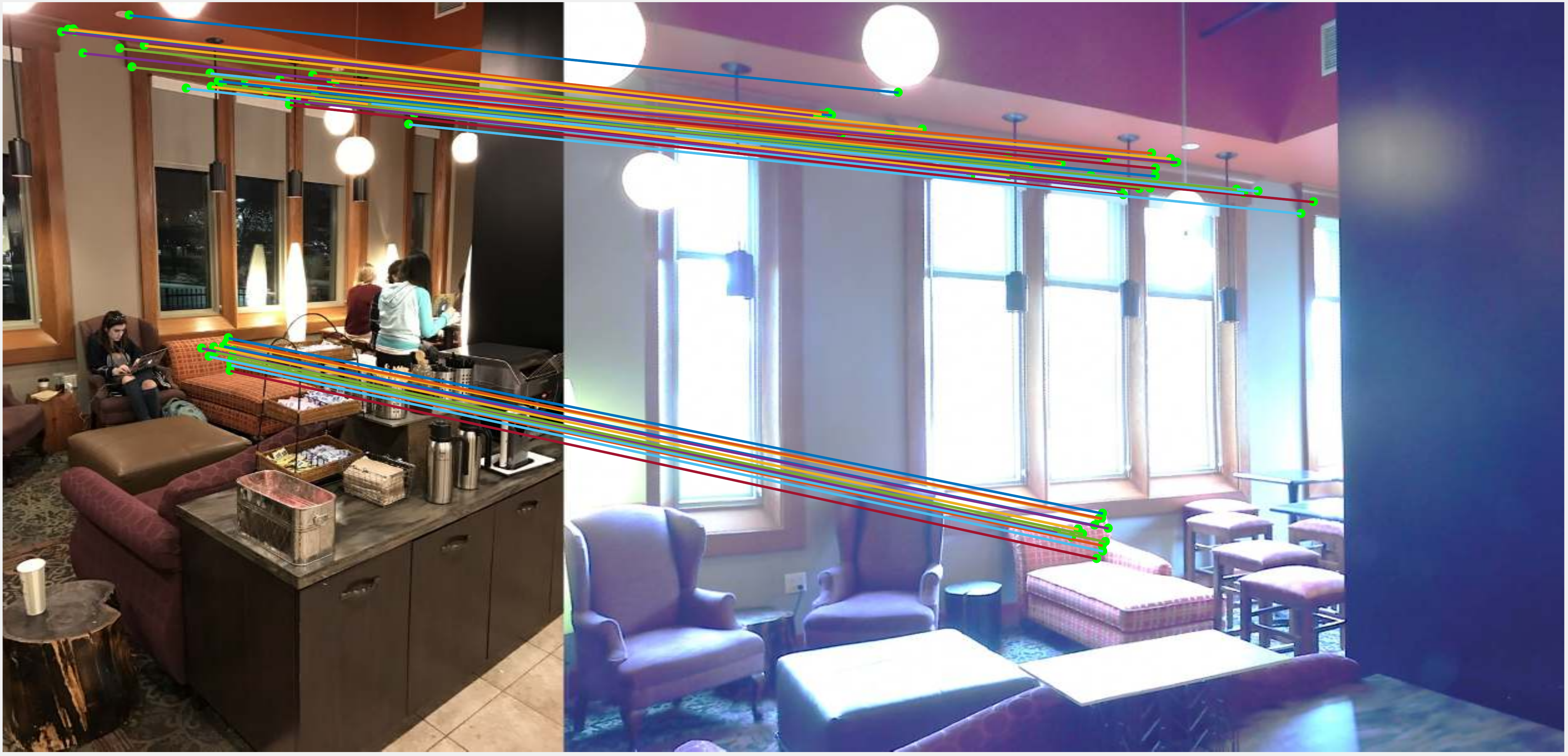}
    \includegraphics[width=\columnwidth]{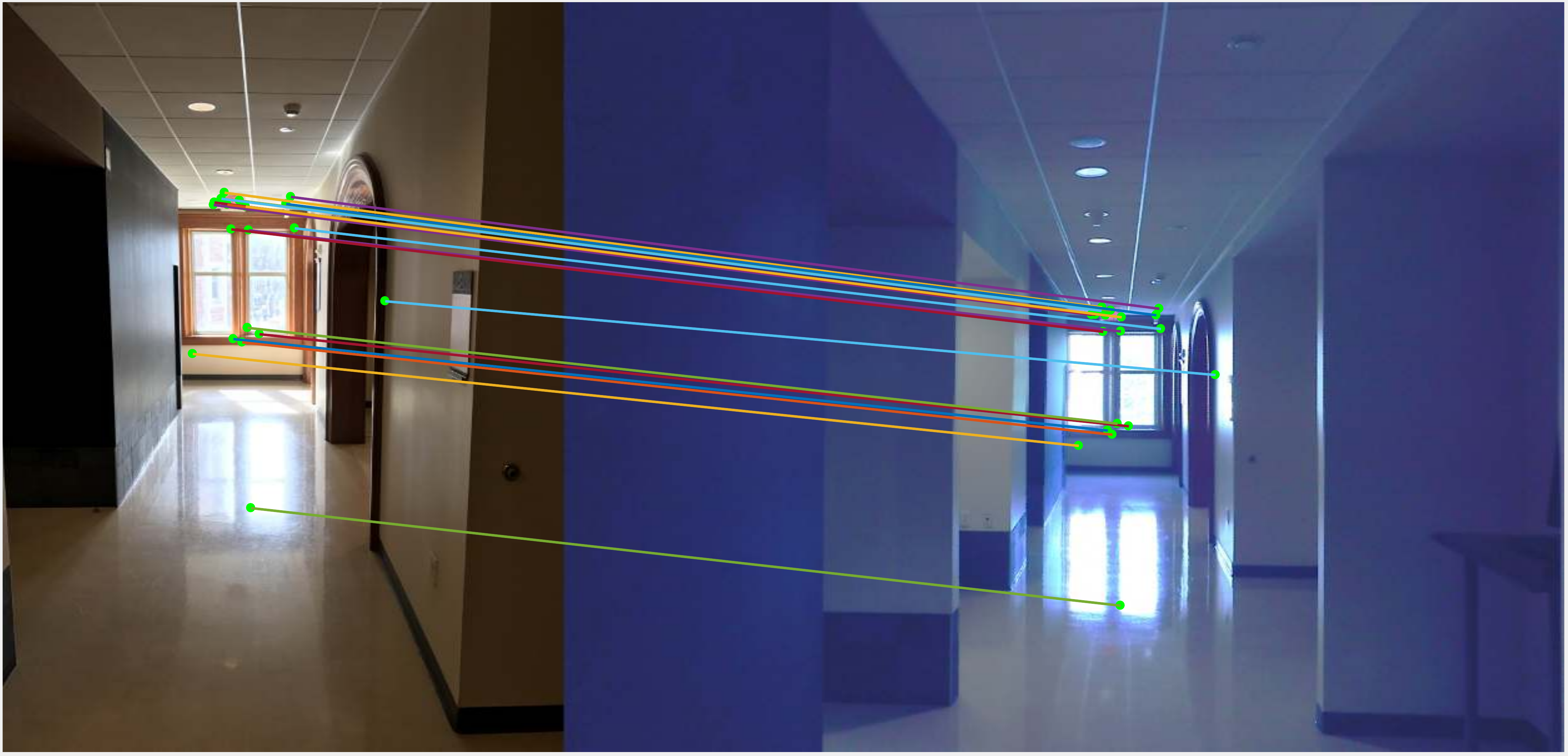}
    % background
    \includegraphics[width=\columnwidth]{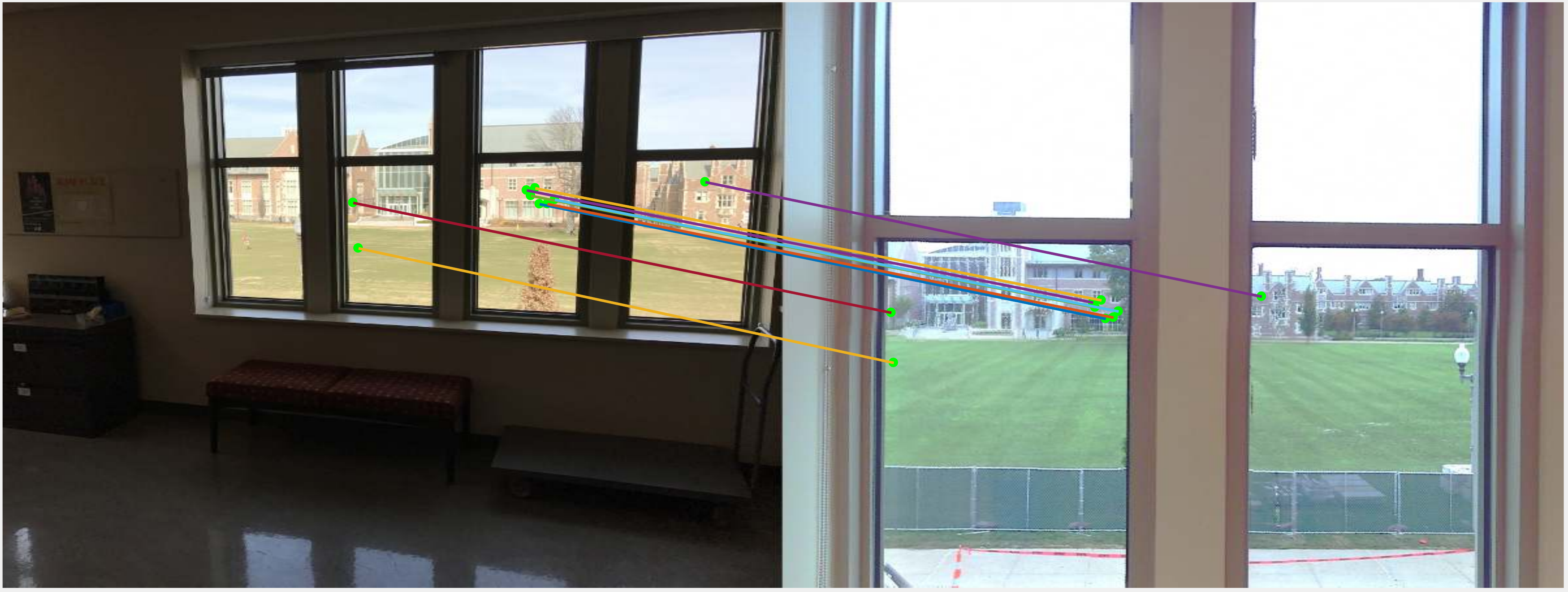}
    % others
    \caption{\small \textbf{Examples of correctly matched image pairs from the InLoc~\cite{Taira2018InLoc} dataset.} Our features are robust to significant changes in viewpoint as it can be seen in the first example. In textureless areas, our features act as an object matcher - correspondences are found between the furniture of different scenes. Sometimes, matches are even found across windows on nearby buildings.}
    \label{fig:inloc-good}
\end{figure}

\begin{figure}[p]
    \centering
    % similar object from a different scene
    \includegraphics[width=\columnwidth]{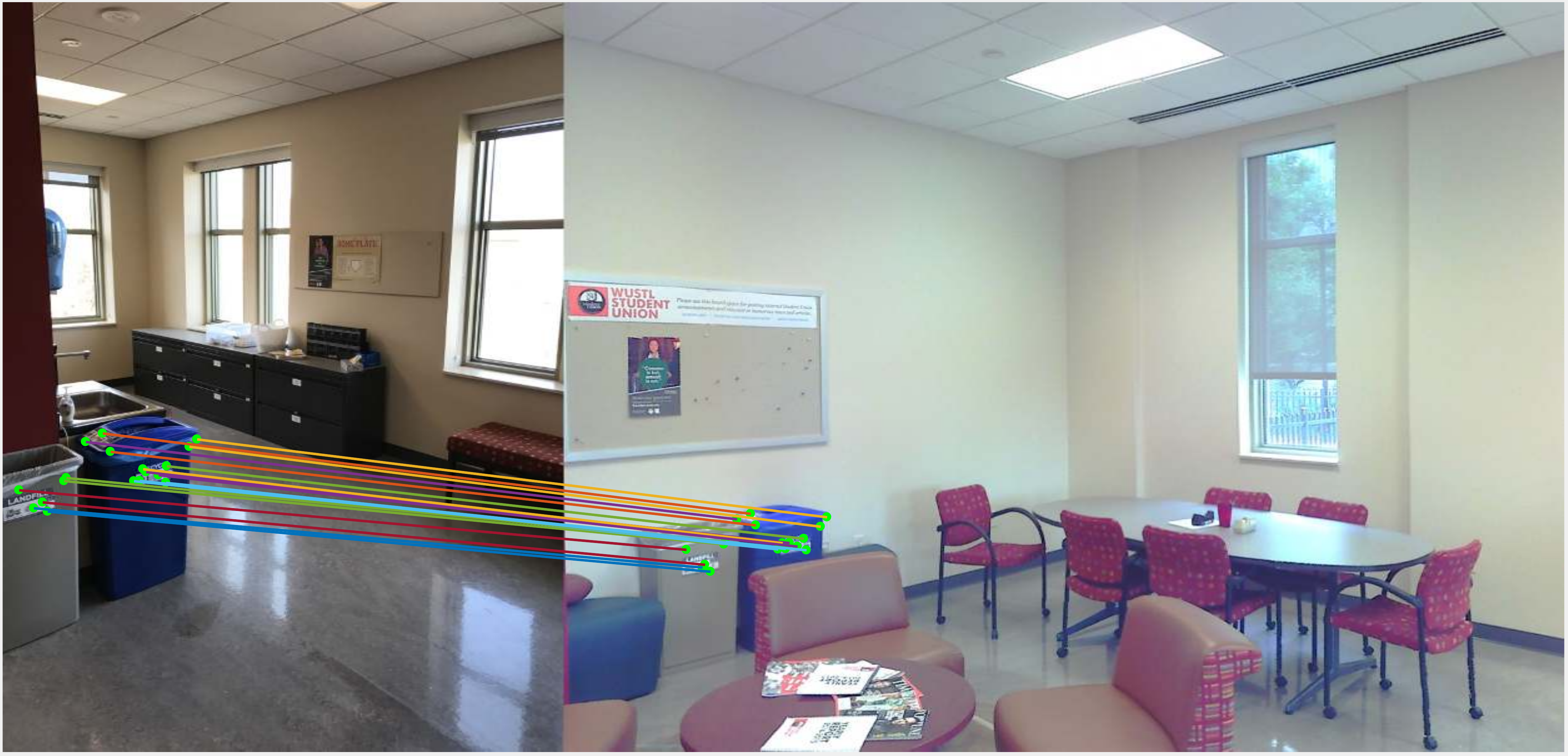}
    \includegraphics[width=\columnwidth]{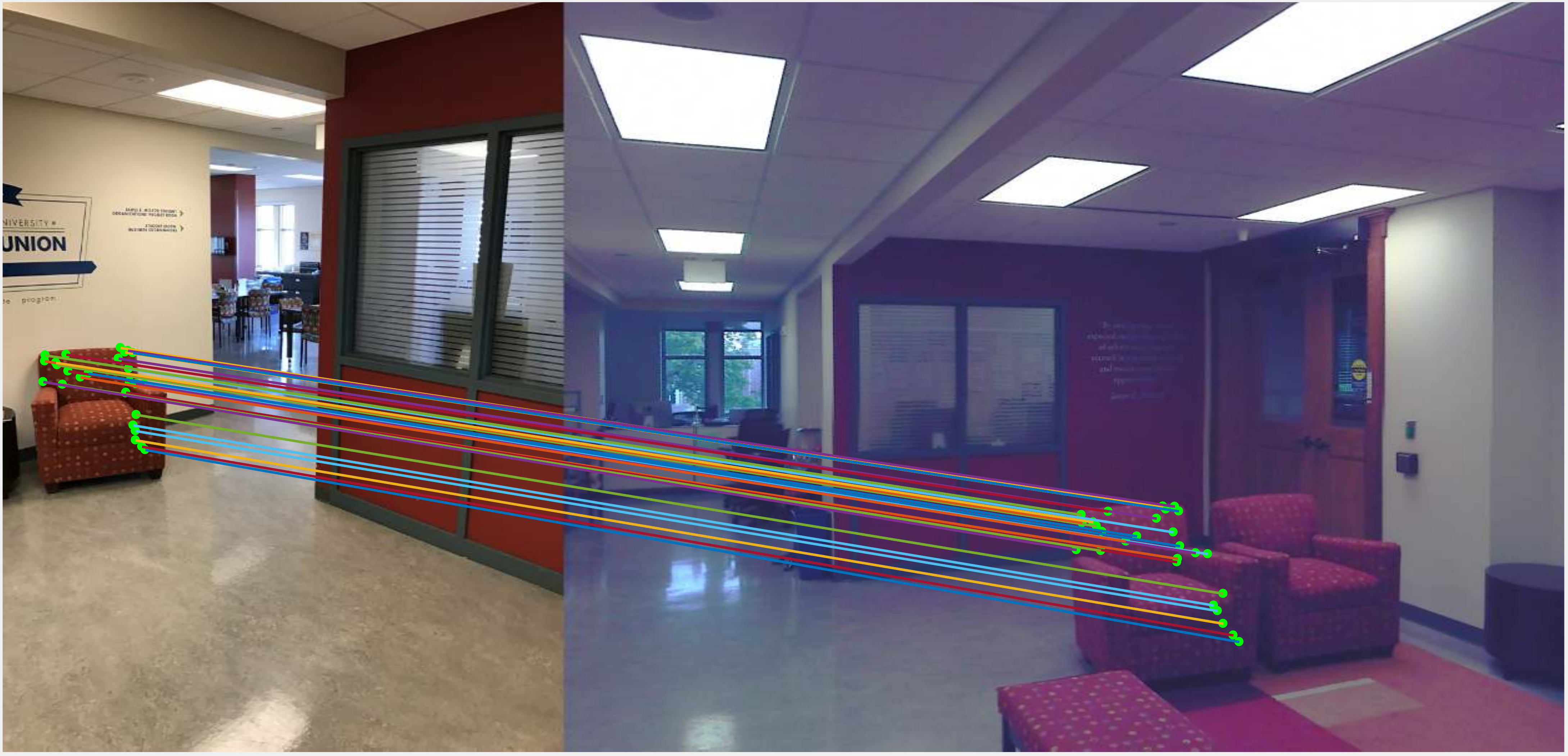}
    \includegraphics[width=\columnwidth]{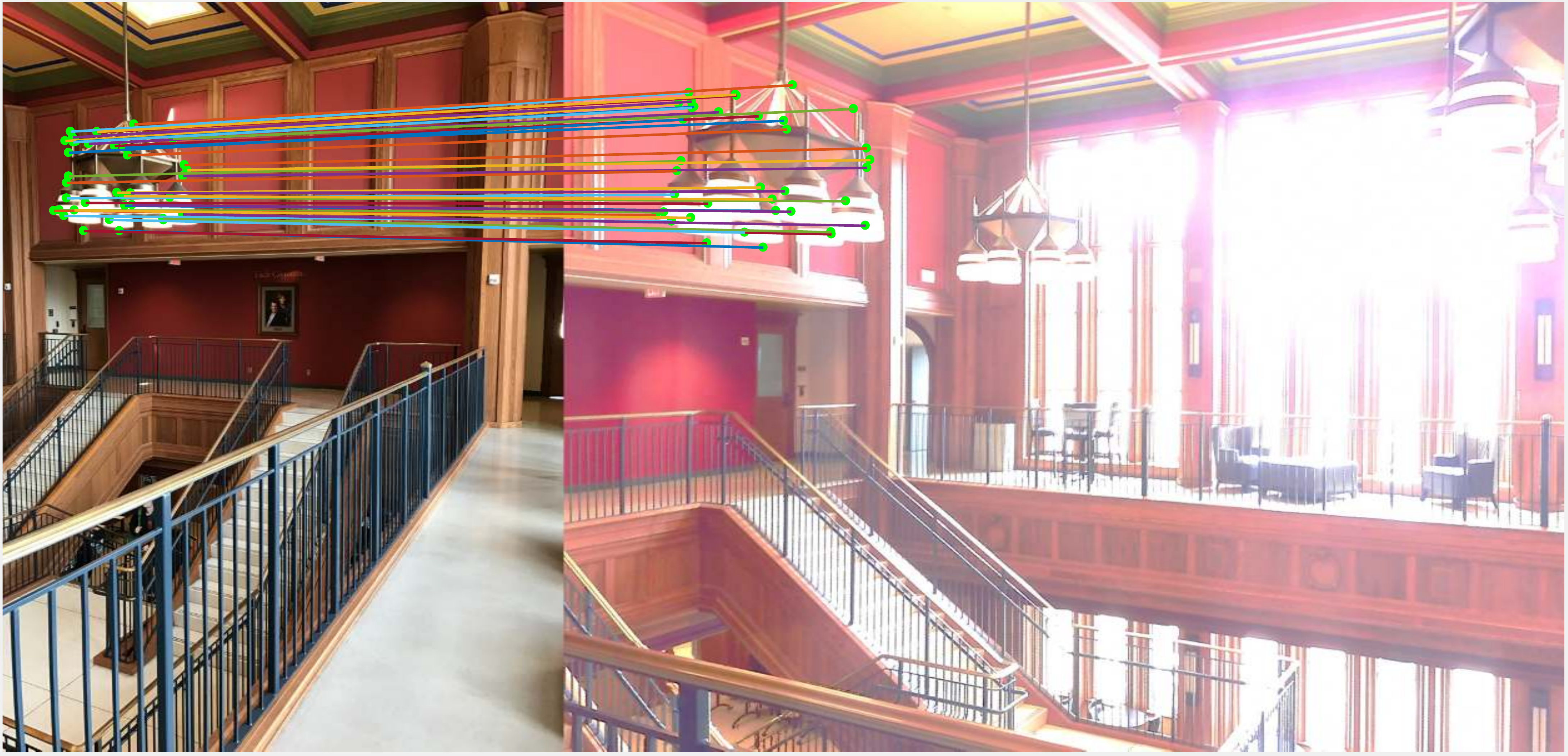}
    % repeated patterns
    \includegraphics[width=\columnwidth]{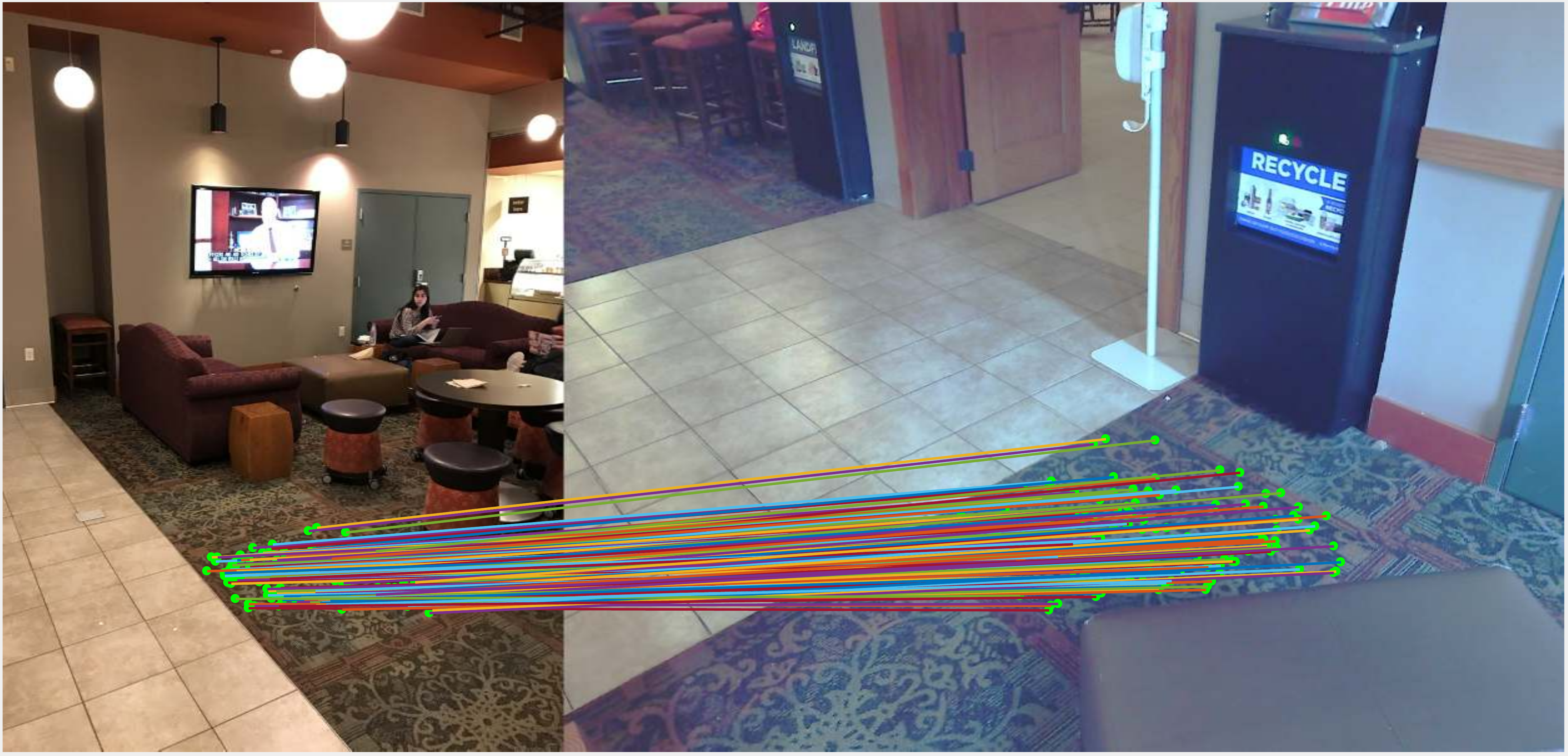}
    \includegraphics[width=\columnwidth]{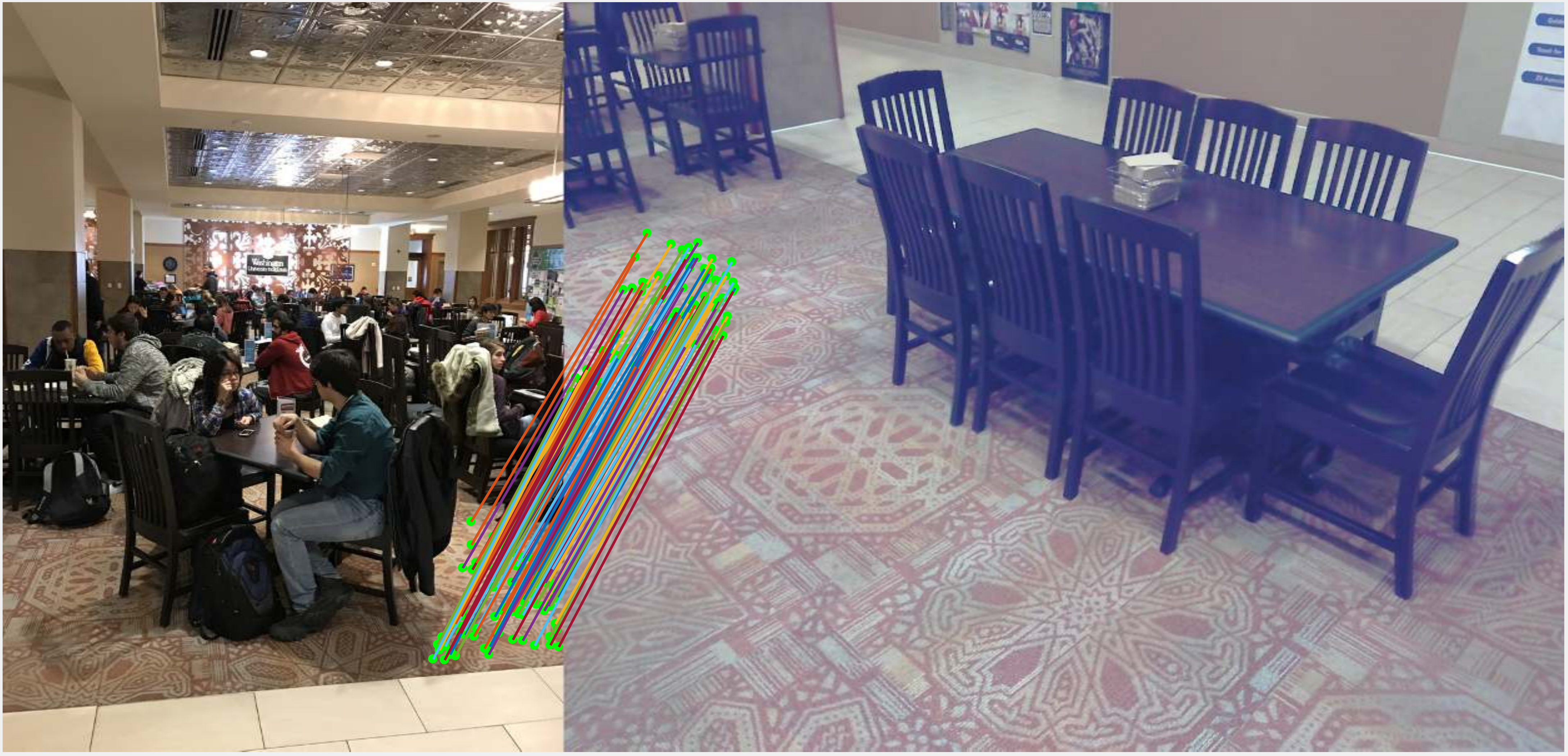}
    \caption{\small \textbf{Failure cases from the InLoc~\cite{Taira2018InLoc} dataset.} Even though they are visually correct, the matches sometimes put in correspondence identical objects from different scenes. Another typical error case is due to repeated patterns (e.g. on carpets) which yield a significant number of inliers.}
    \label{fig:inloc-bad}
\end{figure}

\begin{figure*}[p]
    \centering
    \includegraphics[width=0.725\textwidth]{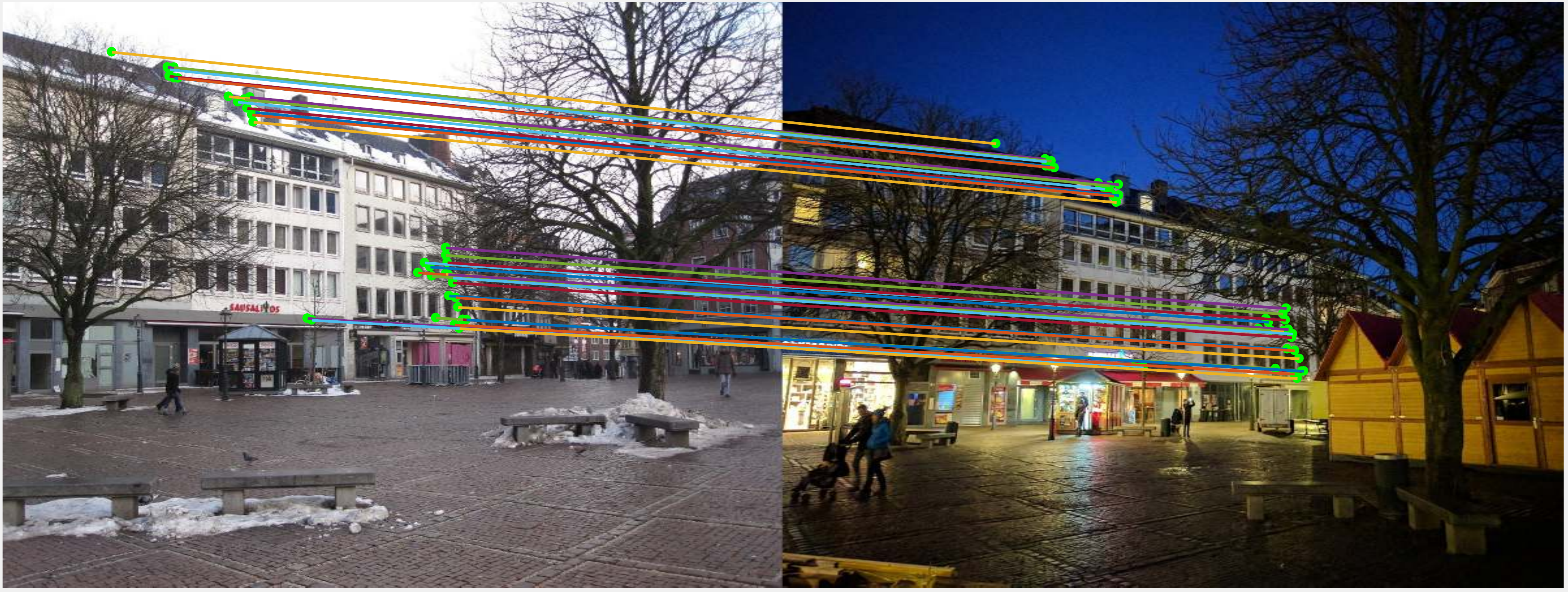}
    \includegraphics[width=0.725\textwidth]{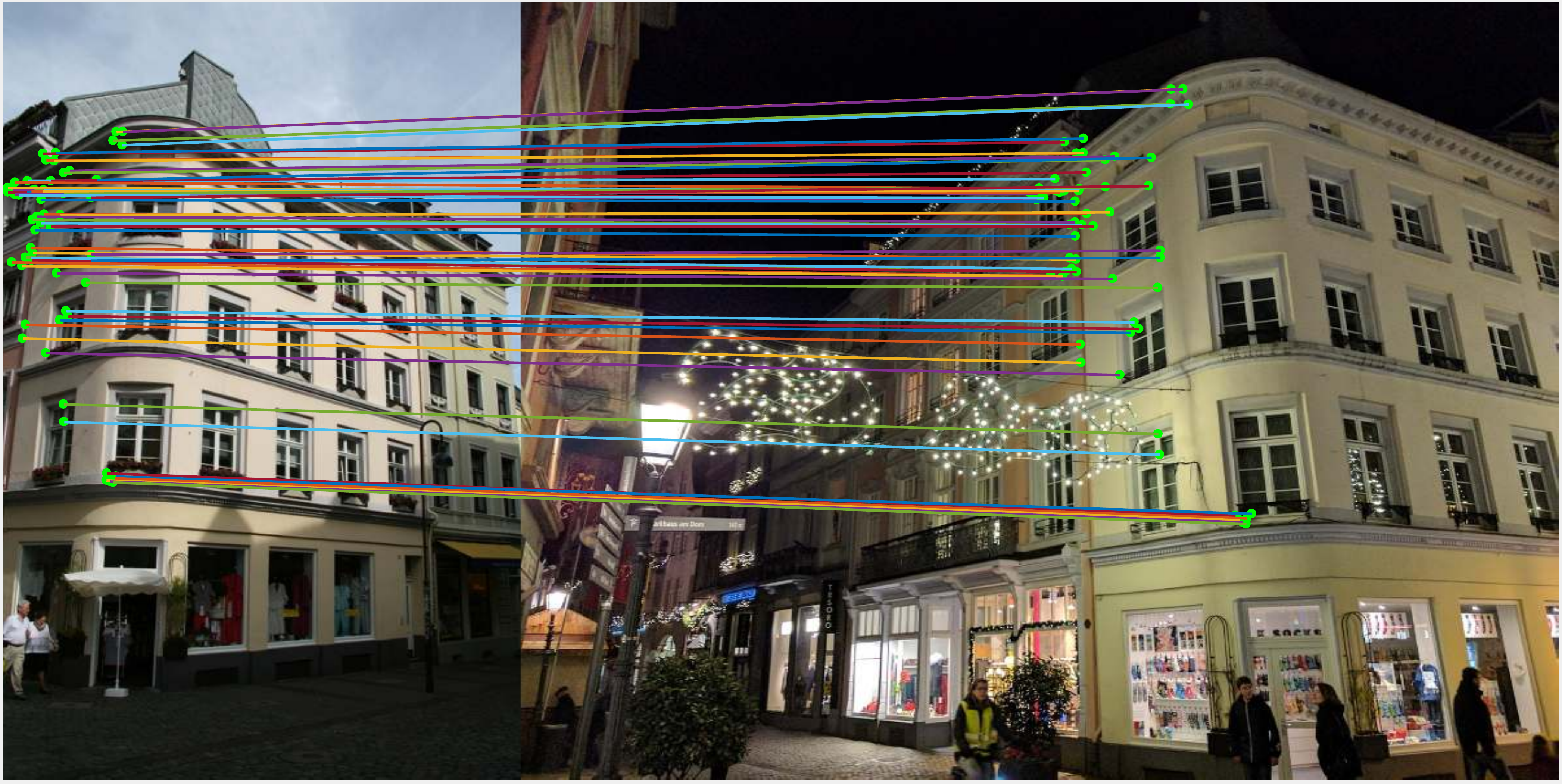}
    \includegraphics[width=0.725\textwidth]{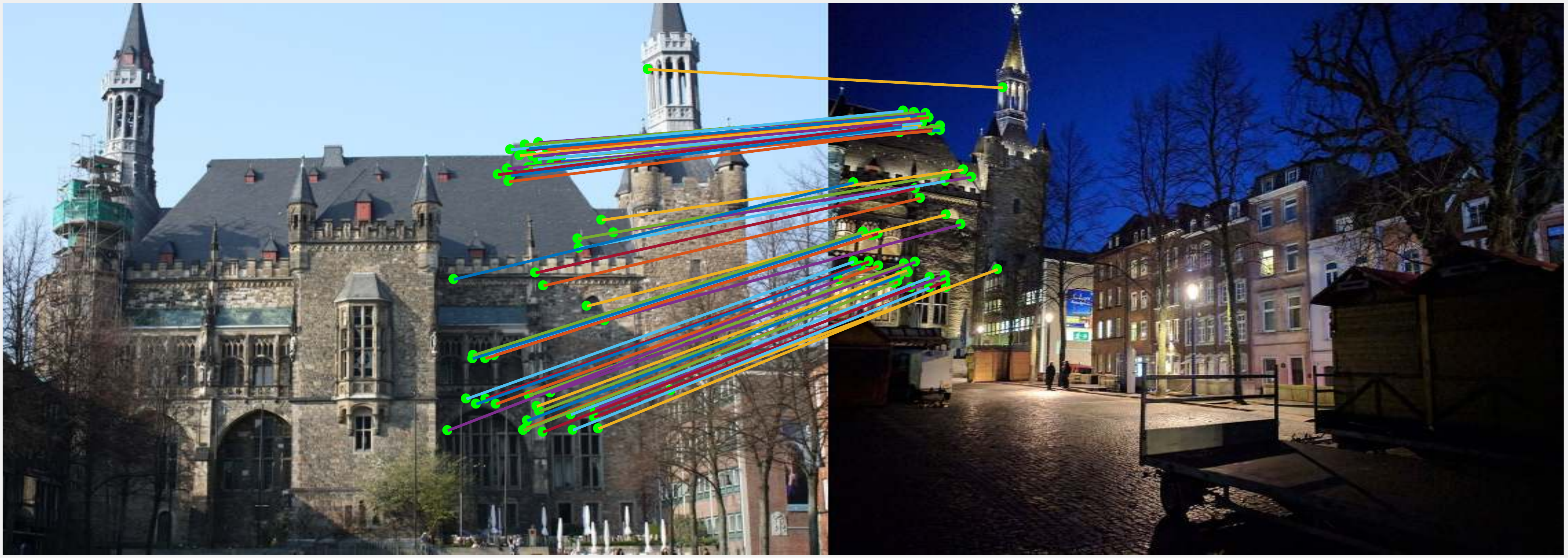}
    \includegraphics[width=0.725\textwidth]{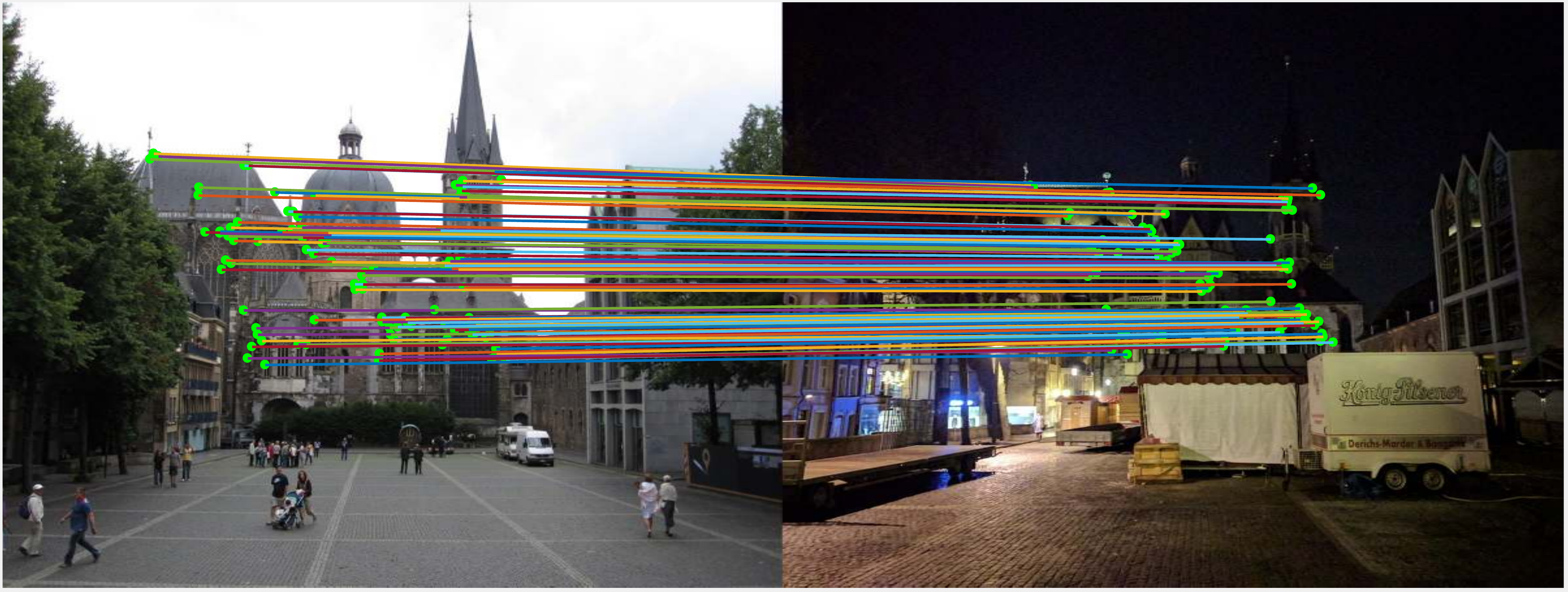}
    \caption{\small \textbf{Examples of correctly matched image pairs from the Aachen Day-Night~\cite{Sattler2017Benchmarking, Sattler2012Image} dataset.} Our features consistently provide a significant number of good matches between images with strong illumination changes. The first two image pairs come from scenes where no other method was able to register the night-time image. For the last two, DELF~\cite{Noh2017Largescale} was the only other method that succeeded.}
    \label{fig:aachen-qualitative}
\end{figure*}

Figures~\ref{fig:inloc-good} and~\ref{fig:inloc-bad} show examples from the InLoc~\cite{Taira2018InLoc} dataset: firstly, we show a few good matches in challenging conditions (significant viewpoint changes and textureless areas) and then we illustrate the main failure modes of D2 features on indoors scenes (repeated objects / patterns). Figure~\ref{fig:aachen-qualitative} shows some example matches on the difficult scenes from the Aachen Day-Night~\cite{Sattler2017Benchmarking,Sattler2012Image} camera localization challenge.

{\small
\bibliographystyle{ieee}
\bibliography{shortstrings,paper}

\begin{thebibliography}{10}\itemsep=-1pt

\bibitem{DTU}
Henrik Aan{\ae}s, Anders~Lindbjerg Dahl, and Kim~Steenstrup Pedersen.
\newblock Interesting interest points.
\newblock {\em IJCV}, 97(1):18--35, 2012.

\bibitem{adelson1984pyramid}
Edward~H. Adelson, Charles~H. Anderson, James~R. Bergen, Peter~J. Burt, and
  Joan~M. Ogden.
\newblock Pyramid methods in image processing.
\newblock {\em RCA engineer}, 29(6):33--41, 1984.

\bibitem{Arandjelovic2016NetVLAD}
Relja Arandjelovic, Petr Gronat, Akihiko Torii, Tomas Pajdla, and Josef Sivic.
\newblock {NetVLAD}: {CNN} architecture for weakly supervised place
  recognition.
\newblock In {\em Proc. CVPR}, 2016.

\bibitem{Arandjelovic2012Three}
Relja Arandjelovic and Andrew Zisserman.
\newblock Three things everyone should know to improve object retrieval.
\newblock In {\em Proc. CVPR}, 2012.

\bibitem{HPATCHES}
Vassileios Balntas, Karel Lenc, Andrea Vedaldi, and Krystian Mikolajczyk.
\newblock {HPatches}: A benchmark and evaluation of handcrafted and learned
  local descriptors.
\newblock In {\em Proc. CVPR}, 2017.

\bibitem{Balntas2016Learning}
Vassileios Balntas, Edgar Riba, Daniel Ponsa, and Krystian Mikolajczyk.
\newblock Learning local feature descriptors with triplets and shallow
  convolutional neural networks.
\newblock In {\em Proc. BMVC.}, 2016.

\bibitem{Bay2006ECCV}
Herbert Bay, Tinne Tuytelaars, and Luc Van~Gool.
\newblock {SURF: Speeded Up Robust Features}.
\newblock In {\em Proc. ECCV}, 2006.

\bibitem{Brown2011PAMI}
Matthew Brown, Gang Hua, and Simon Winder.
\newblock {Discriminative Learning of Local Image Descriptors}.
\newblock {\em IEEE PAMI}, 33(1):43--57, 2011.

\bibitem{calonder2010brief}
Michael Calonder, Vincent Lepetit, Christoph Strecha, and Pascal Fua.
\newblock {BRIEF}: Binary robust independent elementary features.
\newblock In {\em Proc. ECCV}, 2010.

\bibitem{Choy2016NIPS}
Christopher~B. Choy, JunYoung Gwak, Silvio Savarese, and Manmohan Chandraker.
\newblock {Universal Correspondence Network}.
\newblock In {\em NIPS}, 2016.

\bibitem{HAN}
Kai Cordes, Bodo Rosenhahn, and J{\"o}rn Ostermann.
\newblock Increasing the accuracy of feature evaluation benchmarks using
  differential evolution.
\newblock In {\em IEEE Symposium on Differential Evolution (SDE)}, 2011.

\bibitem{Deng2009ImageNet}
Jia Deng, Wei Dong, Richard Socher, Li-Jia Li, Kai Li, and Li Fei-Fei.
\newblock {ImageNet}: A large-scale hierarchical image database.
\newblock In {\em Proc. CVPR}, 2009.

\bibitem{Detone2018CVPRW}
Daniel DeTone, Tomasz Malisiewicz, and Andrew Rabinovich.
\newblock {SuperPoint: Self-Supervised Interest Point Detection and
  Description}.
\newblock In {\em CVPR Workshops}, 2018.

\bibitem{Dong2015CVPR}
Jingming Dong and Stefano Soatto.
\newblock {Domain-size pooling in local descriptors: DSP-SIFT}.
\newblock In {\em Proc. CVPR}, 2015.

\bibitem{Fathy2018ECCV}
Mohammed~E. Fathy, Quoc-Huy Tran, M. Zeeshan~Zia, Paul Vernaza, and Manmohan
  Chandraker.
\newblock {Hierarchical Metric Learning and Matching for 2D and 3D Geometric
  Correspondences}.
\newblock In {\em Proc. ECCV}, 2018.

\bibitem{felzenszwalb2010object}
Pedro~F. Felzenszwalb, Ross~B. Girshick, David McAllester, and Deva Ramanan.
\newblock Object detection with discriminatively trained part-based models.
\newblock {\em IEEE PAMI}, 32(9):1627--1645, 2010.

\bibitem{Furukawa2010PAMI}
Yasutaka Furukawa and Jean Ponce.
\newblock Accurate, dense, and robust multiview stereopsis.
\newblock {\em IEEE PAMI}, 32(8):1362--1376, 2010.

\bibitem{Gordo2017IJCV}
Albert Gordo, Jon Almaz\'{a}n, Jerome Revaud, and Diane Larlus.
\newblock {End-to-End Learning of Deep Visual Representations for Image
  Retrieval}.
\newblock {\em IJCV}, 124(2):237--254, 2017.

\bibitem{Harris1988Combined}
Chris Harris and Mike Stephens.
\newblock A combined corner and edge detector.
\newblock In {\em Proceedings of the Alvey Vision Conference}, 1988.

\bibitem{He2015Deep}
Kaiming He, Xiangyu Zhang, Shaoqing Ren, and Jian Sun.
\newblock Deep residual learning for image recognition.
\newblock In {\em Proc. CVPR}, 2016.

\bibitem{Heinly2015CVPR}
Jared Heinly, Johannes~L. Sch{\"o}nberger, Enrique Dunn, and Jan-Michael Frahm.
\newblock {Reconstructing the World* in Six Days *(As Captured by the Yahoo 100
  Million Image Dataset)}.
\newblock In {\em Proc. CVPR}, 2015.

\bibitem{holschneider1990real}
Matthias Holschneider, Richard Kronland-Martinet, Jean Morlet, and Ph
  Tchamitchian.
\newblock A real-time algorithm for signal analysis with the help of the
  wavelet transform.
\newblock In {\em Wavelets, Time Frequency Methods and Phase Space}, pages
  286--297. 1990.

\bibitem{AMOS}
Nathan Jacobs, Nathaniel Roman, and Robert Pless.
\newblock Consistent temporal variations in many outdoor scenes.
\newblock In {\em Proc. CVPR}, 2007.

\bibitem{kingma2014adam}
Diederik~P. Kingma and Jimmy Ba.
\newblock Adam: A method for stochastic optimization.
\newblock In {\em Proc. ICLR}, 2015.

\bibitem{leutenegger2011brisk}
Stefan Leutenegger, Margarita Chli, and Roland Siegwart.
\newblock {BRISK: Binary robust invariant scalable keypoints}.
\newblock In {\em Proc. ICCV}, 2011.

\bibitem{Li2012ECCV}
Yunpeng Li, Noah Snavely, Dan Huttenlocher, and Pascal Fua.
\newblock {Worldwide Pose Estimation using 3{D} Point Clouds}.
\newblock In {\em Proc. ECCV}, 2012.

\bibitem{Li2018MegaDepth}
Zhengqi Li and Noah Snavely.
\newblock {MegaDepth}: Learning single-view depth prediction from internet
  photos.
\newblock In {\em Proc. CVPR}, 2018.

\bibitem{lindeberg1994scale}
Tony Lindeberg.
\newblock Scale-space theory: A basic tool for analyzing structures at
  different scales.
\newblock {\em Journal of applied statistics}, 21(1-2):225--270, 1994.

\bibitem{liu2016ssd}
Wei Liu, Dragomir Anguelov, Dumitru Erhan, Christian Szegedy, Scott Reed,
  Cheng-Yang Fu, and Alexander~C. Berg.
\newblock {SSD}: Single shot multibox detector.
\newblock In {\em Proc. ECCV}, 2016.

\bibitem{Lowe2004Distinctive}
David~G. Lowe.
\newblock Distinctive image features from scale-invariant keypoints.
\newblock {\em IJCV}, 60(2):91--110, 2004.

\bibitem{Luo2018GeoDesc}
Zixin Luo, Tianwei Shen, Lei Zhou, Siyu Zhu, Runze Zhang, Yao Yao, Tian Fang,
  and Long Quan.
\newblock {GeoDesc}: Learning local descriptors by integrating geometry
  constraints.
\newblock In {\em Proc. ECCV}, 2018.

\bibitem{Mikolajczyk2004Scale}
Krystian Mikolajczyk and Cordelia Schmid.
\newblock Scale \& affine invariant interest point detectors.
\newblock {\em IJCV}, 60(1):63--86, 2004.

\bibitem{Mikolajczyk2005PAMI}
Krystian Mikolajczyk and Cordelia Schmid.
\newblock A performance evaluation of local descriptors.
\newblock {\em IEEE PAMI}, 27(10):1615--1630, 2005.

\bibitem{Mikolajczyk2005IJCV}
Krystian Mikolajczyk, Tinne Tuytelaars, Cordelia Schmid, Andrew Zisserman, Jiri
  Matas, Frederik Schaffalitzky, Timor Kadir, and Luc Van~Gool.
\newblock A comparison of affine region detectors.
\newblock {\em IJCV}, 65(1):43--72, 2005.

\bibitem{Mishchuk2017Working}
Anastasiya Mishchuk, Dmytro Mishkin, Filip Radenovic, and Jiri Matas.
\newblock Working hard to know your neighbor's margins: Local descriptor
  learning loss.
\newblock In {\em NIPS}, 2017.

\bibitem{Mishkin2018ECCV}
Dmytro Mishkin, Filip Radenovi{\'{c}}, and Ji{\v{r}}i Matas.
\newblock {Repeatability Is Not Enough: Learning Discriminative Affine Regions
  via Discriminability}.
\newblock In {\em Proc. ECCV}, 2018.

\bibitem{Muja2014PAMI}
Marius Muja and David~G. Lowe.
\newblock {Scalable Nearest Neighbor Algorithms for High Dimensional Data}.
\newblock {\em IEEE PAMI}, 36:2227--2240, 2014.

\bibitem{Noh2017Largescale}
Hyeonwoo Noh, Andre Araujo, Jack Sim, Tobias Weyand, and Bohyung Han.
\newblock Largescale image retrieval with attentive deep local features.
\newblock In {\em Proc. ICCV}, 2017.

\bibitem{Ono2019LFNet}
Yuki Ono, Eduard Trulls, Pascal Fua, and Kwang~Moo Yi.
\newblock {LF-Net}: Learning local features from images.
\newblock In {\em Advances in Neural Information Processing Systems}, 2019.

\bibitem{philbin2007object}
James Philbin, Ondrej Chum, Michael Isard, Josef Sivic, and Andrew Zisserman.
\newblock Object retrieval with large vocabularies and fast spatial matching.
\newblock In {\em Proc. CVPR}, 2007.

\bibitem{Radenovic2018PAMI}
Filip Radenovi{\'c}, Giorgos Tolias, and Ondrej Chum.
\newblock {Fine-tuning CNN Image Retrieval with No Human Annotation}.
\newblock {\em IEEE PAMI}, 2018.

\bibitem{yolov1}
Joseph Redmon, Santosh Divvala, Ross Girshick, and Ali Farhadi.
\newblock {You Only Look Once}: Unified, real-time object detection.
\newblock In {\em Proc. CVPR}, 2016.

\bibitem{fasterrcnn}
Shaoqing Ren, Kaiming He, Ross Girshick, and Jian Sun.
\newblock Faster {R-CNN}: Towards real-time object detection with region
  proposal networks.
\newblock In {\em NIPS}, 2015.

\bibitem{Rocco2017Convolutional}
Ignacio Rocco, Relja Arandjelovic, and Josef Sivic.
\newblock Convolutional neural network architecture for geometric matching.
\newblock In {\em Proc. CVPR}, 2017.

\bibitem{rublee2011orb}
Ethan Rublee, Vincent Rabaud, Kurt Konolige, and Gary Bradski.
\newblock {ORB: An efficient alternative to SIFT or SURF}.
\newblock In {\em Proc. ICCV}, 2011.

\bibitem{Sattler2017Benchmarking}
Torsten Sattler, Will Maddern, Carl Toft, Akihiko Torii, Lars Hammarstrand,
  Erik Stenborg, Daniel Safari, Masatoshi Okutomi, Marc Pollefeys, Josef Sivic,
  Fredrik Kahl, and Tomas Pajdla.
\newblock Benchmarking {6DoF} outdoor visual localization in changing
  conditions.
\newblock In {\em Proc. CVPR}, 2018.

\bibitem{Sattler2017CVPR}
Torsten Sattler, Akihiko Torii, Josef Sivic, Marc Pollefeys, Hajime Taira,
  Masatoshi Okutomi, and Tomas Pajdla.
\newblock {Are Large-Scale 3D Models Really Necessary for Accurate Visual
  Localization?}
\newblock In {\em Proc. CVPR}, 2017.

\bibitem{Sattler2012Image}
Torsten Sattler, Tobias Weyand, Bastian Leibe, and Leif Kobbelt.
\newblock Image retrieval for image-based localization revisited.
\newblock In {\em Proc. BMVC.}, 2012.

\bibitem{Savinov2017NIPS}
Nikolay Savinov, Lubor Ladicky, and Marc Pollefeys.
\newblock Matching neural paths: transfer from recognition to correspondence
  search.
\newblock In {\em NIPS}, 2017.

\bibitem{Savinov2017CVPR}
Nikolay Savinov, Akihito Seki, Lubor Ladicky, Torsten Sattler, and Marc
  Pollefeys.
\newblock {Quad-networks: unsupervised learning to rank for interest point
  detection}.
\newblock In {\em Proc. CVPR}, 2017.

\bibitem{schoenberger2016sfm}
Johannes~Lutz Sch\"{o}nberger and Jan-Michael Frahm.
\newblock Structure-from-motion revisited.
\newblock In {\em Proc. CVPR}, 2016.

\bibitem{Schonberger2017Comparative}
Johannes~L. Sch\"onberger, Hans Hardmeier, Torsten Sattler, and Marc Pollefeys.
\newblock Comparative evaluation of hand-crafted and learned local features.
\newblock In {\em Proc. CVPR}, 2017.

\bibitem{Schoenberger2018CVPR}
Johannes~L. Sch{\"o}nberger, Marc Pollefeys, Andreas Geiger, and Torsten
  Sattler.
\newblock {Semantic Visual Localization}.
\newblock In {\em Proc. CVPR}, 2018.

\bibitem{schoenberger2016mvs}
Johannes~Lutz Sch\"{o}nberger, Enliang Zheng, Marc Pollefeys, and Jan-Michael
  Frahm.
\newblock Pixelwise view selection for unstructured multi-view stereo.
\newblock In {\em Proc. ECCV}, 2016.

\bibitem{SimoSerra2015ICCV}
Edgar Simo-Serra, Eduard Trulls, Luis Ferraz, Iasonas Kokkinos, Pascal Fua, and
  Francesc Moreno-Noguer.
\newblock {Discriminative Learning of Deep Convolutional Feature Point
  Descriptors}.
\newblock In {\em Proc. ICCV}, 2015.

\bibitem{Simonyan2014PAMI}
Karen Simonyan, Andrea Vedaldi, and Andrew Zisserman.
\newblock {Learning Local Feature Descriptors Using Convex Optimisation}.
\newblock {\em IEEE PAMI}, 36(8):1573--1585, 2014.

\bibitem{Simonyan2014Very}
Karen Simonyan and Andrew Zisserman.
\newblock Very deep convolutional networks for large-scale image recognition.
\newblock In {\em Proc. ICLR}, 2015.

\bibitem{Svarm2017PAMI}
Linus Sv{\"a}rm, Olof Enqvist, Fredrik Kahl, and Magnus Oskarsson.
\newblock {City-Scale Localization for Cameras with Known Vertical Direction}.
\newblock {\em IEEE PAMI}, 39(7):1455--1461, 2017.

\bibitem{Taira2018InLoc}
Hajime Taira, Masatoshi Okutomi, Torsten Sattler, Mircea Cimpoi, Marc
  Pollefeys, Josef Sivic, Tomas Pajdla, and Akihiko Torii.
\newblock {InLoc}: Indoor visual localization with dense matching and view
  synthesis.
\newblock In {\em Proc. CVPR}, 2018.

\bibitem{Tola2010}
Engin Tola, Vincent Lepetit, and Pascal Fua.
\newblock {DAISY: An Efficient Dense Descriptor Applied to Wide-Baseline
  Stereo}.
\newblock {\em IEEE PAMI}, 32(5):815--830, 2010.

\bibitem{Tolias2016ICLR}
Giorgos Tolias, Ronan Sicre, and Herv{\'e} J{\'e}gou.
\newblock Particular object retrieval with integral max-pooling of cnn
  activations.
\newblock In {\em Proc. ICLR}, 2016.

\bibitem{Torii2015CVPR}
Akihiko Torii, Relja Arandjelovic, Josef Sivic, Masatoshi Okutomi, and Tomas
  Pajdla.
\newblock {24/7 Place Recognition by View Synthesis}.
\newblock In {\em Proc. CVPR}, 2015.

\bibitem{PHOS}
Vassilios Vonikakis, Dimitris Chrysostomou, Rigas Kouskouridas, and Antonios
  Gasteratos.
\newblock Improving the robustness in feature detection by local contrast
  enhancement.
\newblock In {\em IEEE International Conference on Imaging Systems and
  Techniques (IST)}, 2012.

\bibitem{wilson2014robust}
Kyle Wilson and Noah Snavely.
\newblock Robust global translations with 1dsfm.
\newblock In {\em Proc. ECCV}, 2014.

\bibitem{Yi2016LIFT}
Kwang~Moo Yi, Eduard Trulls, Vincent Lepetit, and Pascal Fua.
\newblock {LIFT}: Learned invariant feature transform.
\newblock In {\em Proc. ECCV}, 2016.

\bibitem{Yu2016ICLR}
Fisher Yu and Vladlen Koltun.
\newblock {Multi-Scale Context Aggregation by Dilated Convolutions}.
\newblock In {\em ICLR}, 2016.

\bibitem{ASIFT}
Guoshen Yu and Jean-Michel Morel.
\newblock {ASIFT: An algorithm for fully affine invariant comparison}.
\newblock {\em Image Processing On Line}, 1:11--38, 2011.

\bibitem{Zamir2016ECCV}
Amir~R. Zamir, Tilman Wekel, Pulkit Agrawal, Colin Wei, Jitendra Malik, and
  Silvio Savarese.
\newblock {Generic 3D Representation via Pose Estimation and Matching}.
\newblock In {\em Proc. ECCV}, 2016.

\bibitem{Zeiler2014Visualizing}
Matthew~D. Zeiler and Rob Fergus.
\newblock Visualizing and understanding convolutional networks.
\newblock In {\em Proc. ECCV}, 2014.

\bibitem{Zhang2018CVPR}
Linguang Zhang and Szymon Rusinkiewicz.
\newblock {Learning to Detect Features in Texture Images}.
\newblock In {\em Proc. CVPR}, 2018.

\bibitem{Zhou2016ECCVW}
Hao Zhou, Torsten Sattler, and David~W. Jacobs.
\newblock {Evaluating Local Features for Day-Night Matching}.
\newblock In {\em ECCV Workshops}, 2016.

\end{thebibliography}
}

\end{document}